\documentclass{article}

\usepackage[final]{neurips_2023}

\usepackage[varqu,varl,var0,scaled=0.97]{inconsolata}
\usepackage[utf8]{inputenc} %
\usepackage[T1]{fontenc}    %
\usepackage{hyperref}       %
\hypersetup{
    colorlinks=true,
    linkcolor=blue,
    filecolor=blue,
    citecolor=Sepia,
    urlcolor=magenta,
}
\usepackage{url}            %
\usepackage{booktabs}       %
\usepackage{amsfonts}       %
\usepackage{nicefrac}       %
\usepackage{microtype}      %
\usepackage{xcolor}         %

\definecolor{Sepia}{RGB}{112, 66, 20}

\usepackage{amsmath}
\usepackage{amssymb}
\usepackage{mathtools}
\usepackage{amsthm}

\theoremstyle{plain}
\newtheorem{theorem}{Theorem}[section]

\theoremstyle{definition}
\newtheorem{definition}[theorem]{Definition}

\theoremstyle{remark}

\usepackage{url}
\usepackage{graphicx}
\usepackage{xspace}
\usepackage{tabularx}
\usepackage{array}
\usepackage{makecell}
\usepackage{overpic}
\usepackage{enumitem}
\usepackage{sidecap}

\usepackage{soul}
\newcommand{\hlc}[2][yellow]{{%
    \colorlet{foo}{#1}%
    \sethlcolor{foo}\hl{#2}}%
}

\usepackage{colortbl}
\definecolor{TableHL}{rgb}{0.93,0.96,0.98}

\usepackage[draft,inline,nomargin]{fixme}
\fxsetup{theme=color,mode=multiuser}
\FXRegisterAuthor{cz}{acz}{\color{red}chiyuan}  %
\FXRegisterAuthor{nc}{anc}{\color{RoyalBlue}nicholas}  %
\FXRegisterAuthor{kl}{akl}{\color{Plum}katherine}  %
\FXRegisterAuthor{mj}{amj}{\color{TealBlue}matthew}  %
\FXRegisterAuthor{dei}{adi}{\color{RawSienna}daphne}  %
\FXRegisterAuthor{ft}{aft}{\color{RubineRed}florian}  %
\FXRegisterAuthor{an}{aan}{\color{ForestGreen}andrew}  %

\newcommand{\mem}{\operatorname{\mathsf{mem}}}
\newcommand{\infl}{\operatorname{\mathsf{infl}}}
\newcommand{\realnews}{RealNews\xspace}
\newcommand{\wikien}{Wiki40B:en\xspace}

\newcommand{\cfour}{C4\xspace}

\newcommand{\inmodels}{\textsf{\small IN models}\xspace}
\newcommand{\outmodels}{\textsf{\small OUT models}\xspace}

\newcommand{\eglink}[1]{\href{#1}{\textsf{link}} $\rhd$\xspace}
\newcommand{\egomit}{\hlc[gray!40]{[...]}\xspace}
\newcommand{\egcomment}[1]{\hlc[blue!20]{#1}}

\title{Counterfactual Memorization\\ in Neural Language Models}

\author{
  Chiyuan Zhang \\
  Google Research \\
  \texttt{chiyuan@google.com}
 \And
 Daphne Ippolito \\
 Carnegie Mellon University\\
 \texttt{daphnei@cmu.edu} 
 \And
 Katherine Lee \\
 Google DeepMind \\
 \texttt{katherinelee@google.com}
 \AND
 Matthew Jagielski \\
 Google DeepMind \\
 \texttt{jagielski@google.com}
 \And
 Florian Tram{\`e}r \\
 ETH Z{\"u}rich \\
 \texttt{florian.tramer@inf.ethz.ch}
 \And
 Nicholas Carlini \\
 Google DeepMind \\
 \texttt{ncarlini@google.com}
}

\begin{document}

\maketitle

\begin{abstract}
Modern neural language models that are widely used in various NLP tasks risk memorizing sensitive information from their training data.
Understanding this memorization is important in real world applications and also from a learning-theoretical perspective. An open question in previous studies of language model memorization is how to filter out ``common'' memorization. In fact, most memorization criteria strongly correlate with the number of occurrences in the training set, capturing memorized familiar phrases, public knowledge, templated texts, or other repeated data.
We formulate a notion of \emph{counterfactual memorization} which characterizes how a model's predictions change if a particular document is omitted during training.
We identify and study counterfactually-memorized training examples in standard text datasets.
We estimate the influence of each memorized training example on the validation set and on generated texts, showing how this can provide direct evidence of the source of memorization at test time.
\end{abstract}

\section{Introduction}

Modern neural language models (LMs) have achieved impressive results in generating high quality text~\citep[e.g.][]{NEURIPS2020_1457c0d6,zhang2022opt,chowdhery2022palm,openai2023gpt} and have led to breakthroughs in many downstream natural language processing tasks~\citep{devlin-etal-2019-bert,t52020,Bommasani2021-rp}. The paradigm of taking a single large-scale pre-trained model and fine-tuning it for many tasks motivates the study of these models' ability to \emph{generalize} by avoiding \emph{memorizing} their training data. Moreover, memorization of sensitive user information or copyrighted materials in the training data~\citep{carlini2020extracting,vyas2023provable,lee2023language} leads to practical concerns in real world applications.

Previous work on memorization in neural language models demonstrated the ability to extract memorized training data, including sensitive data such as phone numbers and usernames~\citep{carlini2020extracting, ziegler2021copilot,carlini2019secret, henderson2017ethical, thakkar2020understanding, thomas2020investigating}. 
One issue with these extraction attacks is that they primarily identify ``common'' and frequently occurring strings in the training set.
For example, as shown in the analysis of \citet{2021dedup}, near-duplicate training examples, which are very common in standard text corpora, account for a large majority of the memorized content.
To filter out such commonly occurring strings from all memorized texts, previous work applied various heuristic rules to distinguish frequently-occurring sequences from memorization of isolated pieces of information.

In this paper, we propose a principled causal perspective to disentangle memorization of common vs rare data, by directly tying a model's predictions to the presence or absence of individual training examples.
We define \emph{counterfactual memorization} as a measure of the change in a model's prediction when a particular example is excluded from the training set.
Counterfactual memorization accounts for the commonality of an example
as removing one instance of a text that is common across multiple documents will have a minor effect on the model's prediction on that text. The mathematical formulation of counterfactual memorization extends a prior definition of label memorization in classification models~\citep{feldman2020does} to the context of neural language modeling.

Formally, a training document $x$ is considerded counterfactually memorized, when the language model predicts $x$ accurately \emph{if and only if} the model was trained on $x$.
This allows us to construct a %
procedure to quantitatively measure
the memorization of isolated pieces of text, whose sole presence in the training dataset have a large effect on the model's predictions.

Following \citet{feldman2020neural}, we further extend this definition to \emph{counterfactual influence}, which measures the influence of a memorized training text sample on another text example. Counterfactual influence allows us to trace the source of information for a model's predictions, by locating the training example(s) which significantly contributed to it. With these tools, we study memorization across several standard text datasets. Our main contributions are as follows:
\begin{enumerate}[itemsep=0.1em,topsep=0.1em,leftmargin=10mm]
    \item We define counterfactual memorization in neural LMs which gives us a principled perspective to distinguish memorization of ``rare'' and ``common'' information in neural LMs (Section~\ref{sec:methods}).
    \item We estimate counterfactual memorization on several standard text datasets, and confirm that rare memorized examples exist in all of them.
    We study common patterns across memorized text and the memorization profiles of individual internet domains.
    (Section~\ref{sec:memorization}).
    \item We identify an inverse correlation between number of duplicates and counterfactual memorization as compared with previous definitions of memorization (Section~\ref{sec:dup}).
    \item We extend the definition of counterfactual memorization to counterfactual \emph{influence}, and study the impact of memorized examples on the test-time prediction of the validation set examples and generated examples (Section~\ref{sec:influence}).
\end{enumerate}

\section{Related Work}

Previous work analyzed the memorization of large language models on sensitive information (e.g. phone numbers) in the training data~\citep{carlini2020extracting, ziegler2021copilot} or synthetically injected ``canaries''~\citep{carlini2019secret, henderson2017ethical, thakkar2020understanding, thomas2020investigating}. However, not all the memorized texts are equally interesting --- as confirmed in a later study~\citep{2021dedup}, near-duplicated training examples are very common in standard text corpus, and those commonly occurring phrases contribute significantly to memorized texts. In order to distinguish ``common'' memorization of common phrases or public knowledge from ``rare'' memorization of private, rare information, various heuristics were adopted in previous investigations. Our paper proposed a principled perspective towards this problem. Our intuition comes from psychologies studies that categorize human (declarative) memory into \emph{episodic} memory~\citep{tulving1983elements} of specific contents of individual events, and \emph{semantic} memory~\citep{squire1992memory} about general knowledge like grammars and factual information. We would like the models to obtain semantic memory but avoid episodic memory. The capture the latter, we proposed a notion of counterfactual memorization. The mathematical formulation of counterfactual memorization is borrowed from a notion of label memorization in \citet{feldman2020does} and adapted to the context of neural LMs in this paper. This formulation has been studied empirically in the context of computer vision in \citet{feldman2020neural}. In a follow up work, \citet{ilyas2022datamodels} showed that it is possible to fit a \emph{datamodel} to predict the outcome of training a model on a specific training subset and evaluating on a specific input. However, this procedure requires training a massive number of models (e.g. 300,000 for CIFAR-10) on random subsets of the training data, thus is computationally infeasible for the scale of language models considered here.

The general idea of measuring model behavior on held-out training data is common in machine learning. In cross validation, held-out data is used to estimate the test performance for model selection; in learning theory, leave-one-out stability was shown to be deeply connected to generalization~\citep[e.g.][]{mukherjee2006learning}; in differential privacy, the worst case performance difference of models trained on two ``neighboring'' datasets (identical except a single example being held-out or replaced) quantifies the privacy guarantee of a learning algorithm~\citep{dwork2014algorithmic,nasr2021adversary, jagielski2020auditing}. Most previous work aimed for an overall measurement, while our paper focused on characterizing the behaviors of individual examples.

We estimated a \emph{counterfactual influence} to study how a memorized training example impact the model prediction at test time. Influence functions have been used in statistics to assess robust estimators since \citet{hampel1974influence}. Previous papers adopted it to analyze neural network predictions~\citep{Koh2017UnderstandingBP,Koh2019OnTA}. However, the estimation was found to be computational expensive and fragile~\citep{Basu2021InfluenceFI}. \citet{Pruthi2020EstimatingTD} tracks the gradient updates during training to estimate the influence from a training example; \citet{feldman2020does,feldman2020neural} use aggregated statistics from multiple models independently trained on heldout data subsets to estimate the influence. Further extensions were shown to work well on detecting mislabeled data in classification problems~\citep{wang2022data} and characterizing hallucinations in Neural Machine Translation~\citep{raunak-etal-2021-curious}.  Alternative methods also looked at simple data statistics (e.g. co-occurrence counts) without model re-training to infer the causal effects on language models' predictions~\citep{elazar2022measuring}. In this paper, we adapt the approach from \citet{feldman2020does}, and formulate counterfactual influence directly with subset sampling, as oppose to leave-one-out influence. We also extend the estimation to assess the influence on generated examples.

Counterfactual is an important notion in statistical causality~\citep{pearl2000models,rubin2005causal,pearl2009causal,imbens2015causal} useful for studying causal probabilistic inference under alternative conditions. Such counterfactuals may or may not be directly testable (e.g. a counterfactual treatment in medical studies). In this paper, we directly measure the counterfactual influence of a training example by comparing the behavior of the model trained with and without that example.

\section{Counterfactual Memorization}
\label{sec:methods}
To quantify memorization of rare details of a specific training document,
we define the following notion of \emph{counterfactual memorization}. The mathematical formulation is borrowed from \citet{feldman2020does}, where it was originally proposed to quantify label memorization in multi-class classification problems. We extend it to the context of unsupervised neural language modeling.

\begin{definition}[Counterfactual Memorization]
Given a training algorithm $A$ that maps a training dataset $D$ to a trained model $f$, and a measure $M(f,x)$ of the performance of $f$ on a specific example $x$, the counterfactual memorization of a training example $x$ in $D$ is given by
\begin{equation}
    \mem(x) \triangleq
    \underbrace{\mathbb{E}_{S\subset D,x\in S}[M(A(S), x)]}_{\text{performance on $x$ when trained with $x$}} - 
    \underbrace{\mathbb{E}_{S\subset D,x\not\in S}[M(A(S), x)]}_{\text{performance on $x$ when \textbf{not} trained with $x$}},
    \label{eq:mem}
\end{equation}
where $S$ and $S'$ are subsets of training examples sampled from $D$. The expectation is taken with respect to the random sampling of $S$ and $S'$, as well as the randomness in the training algorithm $A$.
\end{definition}

That is, our memorization definition compares the difference between two expected performance measures on a given example $x$.
On one side, we compute the expected performance of a model when trained on datasets that \emph{contain} the example $x$, 
and, on the other side, we compute the expected performance of a model when trained on datasets that do \emph{not} contain the example $x$.
Throughout this paper we use per-token accuracy as the measure $M$.
In other words, we ask the model to predict the next token based on the groundtruth context (preceding tokens), measure the 0-1 loss of the argmax token prediction, and then average it across all predicted tokens. 

The expectations in Equation~\eqref{eq:mem} can be empirically estimated via sampling. Specifically, we train $m$ different models on independently sampled subsets $S_1,\ldots,S_m$ of equal size $|S_i|=r|D|$ for a fixed $r\in(0,1)$. We then divide these models into two groups: the first group contains all models trained on subsets $S$ where $x\in S$; and the second group are all models trained on subsets $S$ where $x\not\in S$. We take the average performance on $x$ in the two groups separately and compute the difference between the two:
\begin{equation}
    \widehat{\mem}(x) \triangleq
    \operatorname*{mean}_{i:x\in S_i}[M(A(S_i),x)] -
     \operatorname*{mean}_{i:x\not\in S_i}[M(A(S_i),x)].
    \label{def:sampling}
\end{equation}
This difference quantifies how the presence or absence of the example $x$ in a model's training set affect the model's performance on $x$.
If there is a large difference between including an example in the training set versus not including it, then we consider this example \emph{counterfactually memorized}.

For each $x$, we refer to models trained with $x$ in the training set ($\{A(S_i):x\in S_i\}$) as \inmodels and the models $x$ was not trained on ($\{A(S_i):x\not\in S_i\}$) as \outmodels. Note we do not need to \emph{retrain} a model for each example $x$. Instead, we train $m$ models once on random subsets of $D$, and compute the estimation (Equation \ref{def:sampling}) for \emph{all} examples using the same set of $m$ models. \citet{ilyas2022datamodels} recently showed that it may also be possible to directly \emph{predict} these scores using a regression model, yet this approach is  computationally prohibitive for large language models.

\section{Analyzing Counterfactual Memorization}
\label{sec:memorization}

We estimate and analyze counterfactual memorization of training examples in three standard text datasets: \realnews~\citep{zellers2019defending}, \cfour~\citep{JMLR:v21:20-074} and \wikien~\citep{49029}.
Unless otherwise specified, we use Transformer-based language models~\citep{vaswani2017attention} equivalent to (decoder only) T5-base~\citep{t52020} with $\sim$112M parameters.
To save computation and enable more direct comparisons across datasets, we truncate the training set for each datasets by taking the first $2^{21}$ documents.
To estimate counterfactual memorization, we train 400 models for each dataset, each on a random $25\%$ subset of the training examples.
In practice, we use a hash-based filtering mechanism to efficiently approximate random subset sampling (details in Appendix~\ref{sec:subsampling}), as the data loading APIs for large text corpora generally support only sequential visits to examples with limited shuffling and subsampling capability within a window.

We train each model for 60 epochs\footnote{Modern language models are usually trained for fewer epochs if the training set is massive.
Since we have a smaller subsampled training set, we train the models for more epochs to allow the models to fit the training data sufficiently to study memorization effects.} using the Adam optimizer~\citep{kingma2014adam} with learning rate 0.1 and weight decay $10^{-5}$.
For \cfour/\realnews/\wikien, respectively, our models converge to an average per-token accuracy of 44.21\%/47.59\%/66.35\% on the subsampled training set, and 27.90\%/31.09\%/49.55\% on the validation set. On average, the models start to overfit at around epoch 5, as indicated by the signal that the validation accuracy starting to decrease.

\subsection{Distribution of Memorization}
Table~\ref{tab:realnews-egs} shows examples from the \realnews training set sampled at various memorization levels.
Examples with the highest memorization are generally unconventional text such as all-capital letters, structured formats (i.e., tables or bullet list), and multilingual texts.
After those artificial examples, examples with intermediate-to-high memorization are most often news reports of specific events.
One of our main goals is to be able to separate memorization of such examples containing details of specific events from memorization of common facts or highly duplicated template texts.
Indeed, templated documents with many near-duplicate copies in the training data generally have low counterfactual memorization.
\cfour and \wikien have similar trends.
Interestingly, though Wikipedia articles are less likely to be auto-generated from templates than the web in general, we do observe repetitive patterns in low-scoring documents, such as ``\_START\_ARTICLE\_ \texttt{<place name>}, Virginia \_START\_PARAGRAPH\_ \texttt{<place name>} is an unincorporated community in \texttt{<county name>}, in the U.S. state of Virginia.''

\begin{table}[]
    \caption{\small Examples of \realnews training set sampled at high, intermediate and low memorization. The URL of each document is included at the beginning of each example. \egomit indicate omitted text for brevity. In the last block, two near-duplicate examples are shown; the \hlc{yellow highlights} in the last block indicate differences.}
    \centering\tiny
    \begin{tabularx}{\linewidth}{@{}llX@{}}
        \toprule
        Index & $\mem$ & Text \\\midrule
        2090855 & 0.6546 & \eglink{http://www.jta.org/1937/12/29/archive/american-jewish-congress-plans-drive-on-job-discrimination} THE AMERICAN JEWISH CONGRESS ANNOUNCED TODAY THE PUBLICATION OF A REPORT ON JEWISH NON-EMPLOYMENT AS A RESULT OF ECONOMIC DISCRIMINATION, \egomit THEREAFTER ONE OF THE DEPARTMENTS OF A.T.\& T. ” ALMOST UNPRECEDENTEDLY ” ENGAGED A JEWISH APPLICANT.
        \\
        2085736 & 0.5755 & \eglink{https://fox59.com/2016/01/28/recipe-chinese-pork-vegetable-soup-with-wonton-noodles/} x RECIPE: Chinese Pork \& Vegetable Soup with Wonton Noodles Chinese Pork \& Vegetable Soup with Wonton Noodles 1 pork tenderloin (about 1-1 1/4 pound size), cooked and cut into 1/2-inch cubes* 5 cups lower-sodium chicken broth 1 cup water** 1/4 cup \egomit Makes 4 servings (about 1 1/2 cups each) Recipe by PorkBeInspired.com with adaptations by culinary dietitian \& nutritionist Kim Galeaz, RDN CD
        \\
        1680600 & 0.5807 & \eglink{https://www.sbs.com.au/language/english/what-is-acknowledgement-of-country_1} Language English \hlc[gray!40]{[... Arabic text ...]} acknowledgement of country \hlc[gray!40]{[... Arabic text ...]} I would like to acknowledge that this meeting is being held on the traditional lands of the (appropriate group) people, and pay my respect to elders both past and present.” \egomit 
        \\\midrule
        2074805 & 0.2835 & \eglink{http://www.christianpost.com/news/texas-students-suspension-for-anti-gay-remark-reversed-94734/} A Texas honors student punished for saying that homosexuality was wrong has had his suspension rescinded \egomit Western Hills High made the correct decision in reversing their course of action. "The decision to rescind the suspension is the correct one. The suspension was wrong and improper," said Staver. "I applaud the student for standing up. We stood with him to resist an unjust suspension and we are pleased that suspension has been reversed." \egomit  Liberty Counsel will continue the right to exercise freedom of conscience and religion," said Staver. "These instances are increasing and will continue to increase unless Christians and people who love liberty stand up and resist this intolerance."
        \\\midrule
        449808 & 0.0361 & \eglink{https://www.thestreet.com/story/12179505/1/first-week-of-dlr-february-2014-options-trading.html} Investors in \hlc{Digital Realty Trust, Inc. ( DLR)} saw new options \hlc{begin trading this week}, for the \hlc{February 2014} expiration. At Stock Options Channel, our YieldBoost formula has looked up and down the \hlc{DLR} options chain for the new \hlc{February 2014} contracts and identified one put and one call contract of particular interest. The put contract at the \$\hlc{45.00} strike price has a current bid of \$\hlc{1.00}. \egomit
        \\
        1157311 & 0.0356 & \eglink{https://www.thestreet.com/story/12425372/1/anf-april-4th-options-begin-trading.html} Investors in \hlc{Abercrombie \& Fitch Co. (ANF)} saw new options \hlc{become available today}, for the \hlc{April 4th} expiration. At Stock Options Channel, our YieldBoost formula has looked up and down the \hlc{ANF} options chain for the new \hlc{April 4th} contracts and identified one put and one call contract of particular interest. The put contract at the \$\hlc{34.00} strike price has a current bid of \$\hlc{1.97}. \egomit
        \\\bottomrule
    \end{tabularx}
    \label{tab:realnews-egs}
\end{table}

\begin{figure}
    \centering
    \begin{overpic}[width=.32\linewidth]{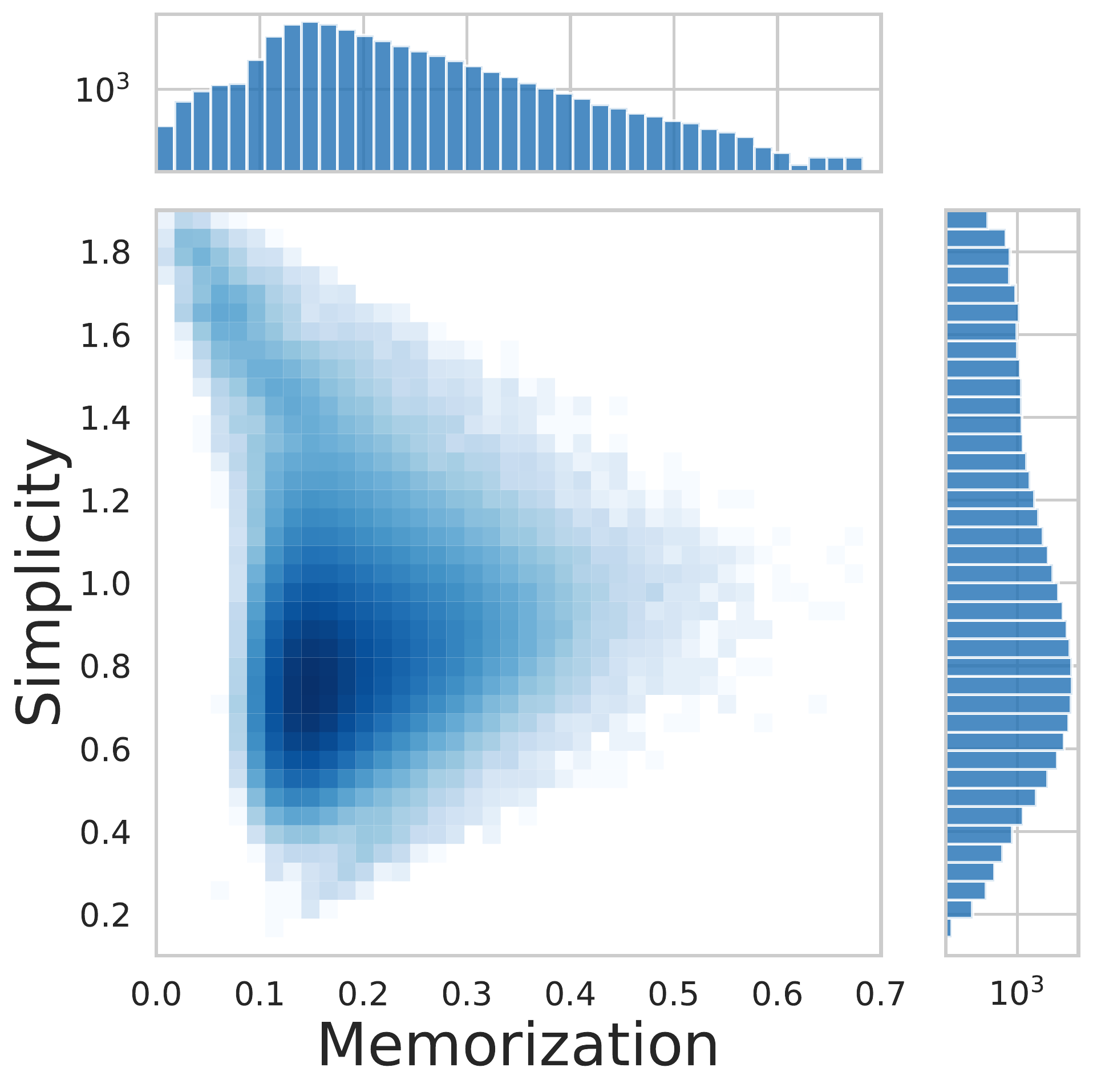}
        \put(30,20){\scriptsize\fbox{\realnews}}
    \end{overpic}\hspace{1pt}
    \begin{overpic}[width=.32\linewidth]{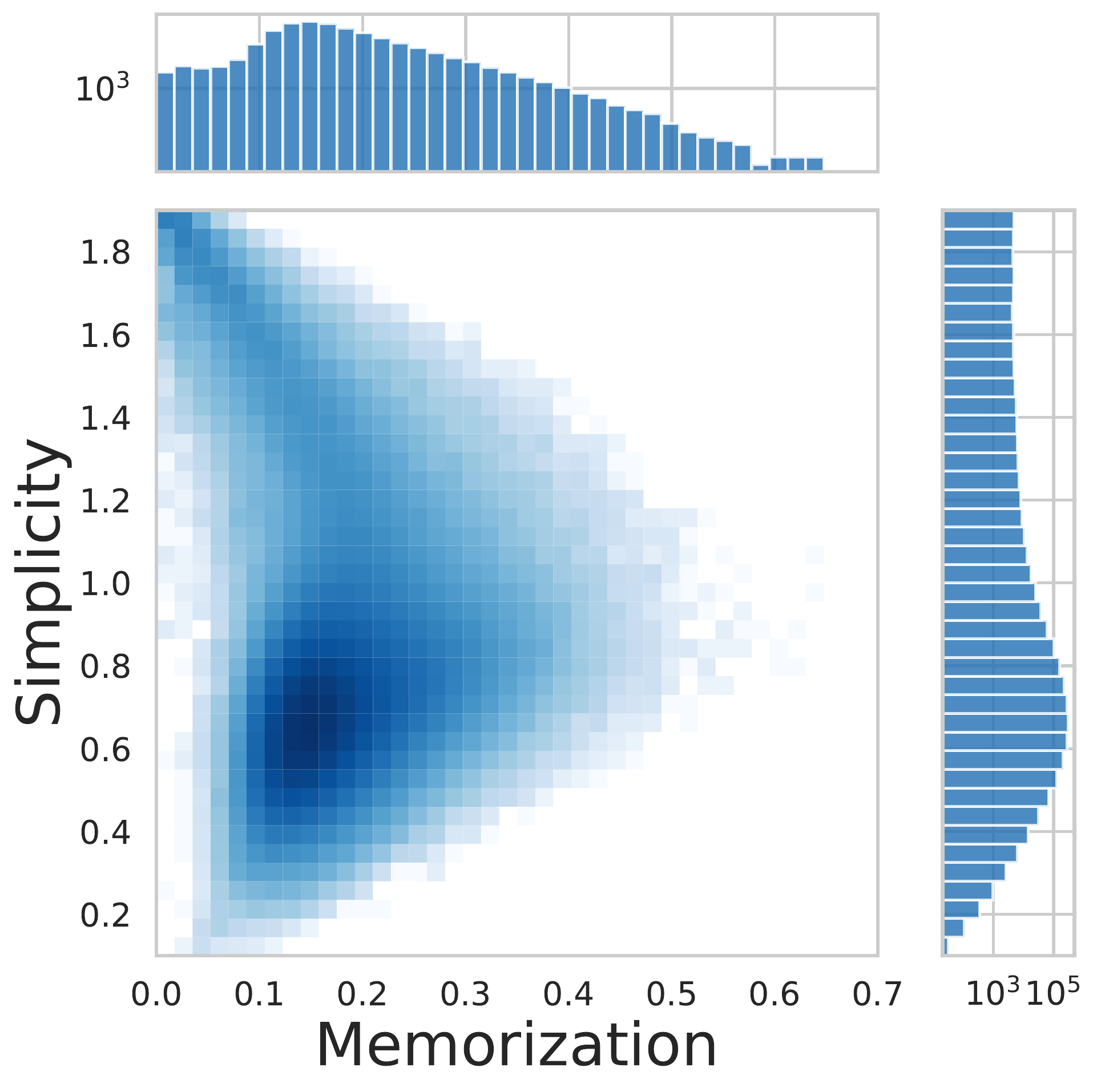}
        \put(55,20){\scriptsize\fbox{\cfour}}
    \end{overpic}\hspace{1pt}
    \begin{overpic}[width=.32\linewidth]{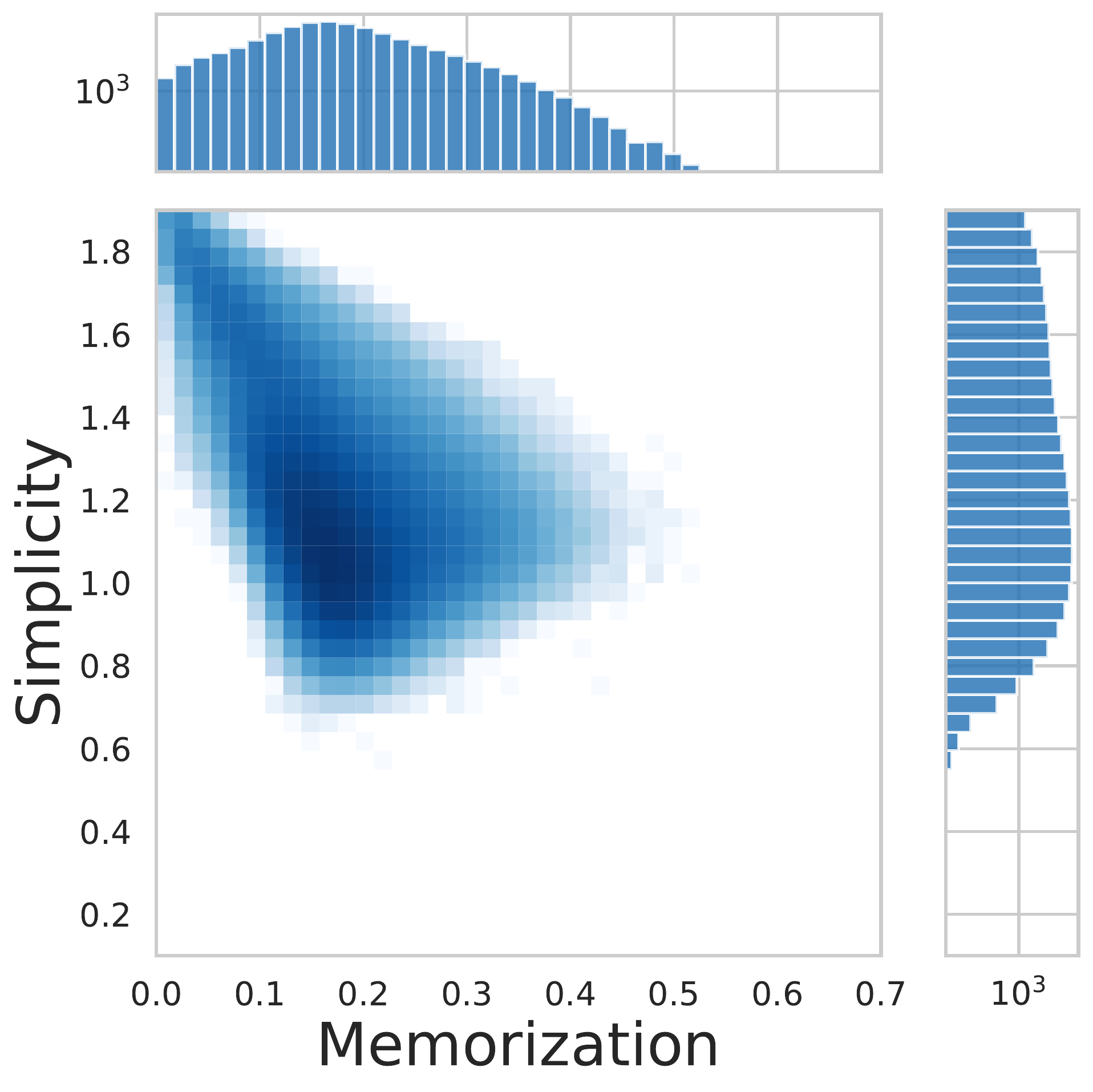}
        \put(25,20){\scriptsize\fbox{\wikien}}
    \end{overpic}
    \vskip-3pt
    \caption{\small The joint distribution of counterfactual memorization (X axis) and simplicity (Y axis), where simplicity is measured as the overall accuracy for an example across all models.
    (Histograms are in log-scale).}
    \label{fig:mem-sim}
\end{figure}

To visualize the distribution of memorization, we plot 2D histograms in Figure~\ref{fig:mem-sim}, where the x-axis shows the difference of \textsf{IN}-accuracy and \textsf{OUT}-accuracy (i.e. the counterfactual memorization), and the y-axis shows the sum of the two, which we term ``simplicity''.
A simple example is one that is scored highly regardless of whether a model saw it during training.
The histograms are plotted in log scale to better visualize the exponential decay in the tail for high memorization and simplicity levels.

From the 2D density plots, we find that easy examples tend to have low memorization. However, there is no simple linear correlation. Peak memorization occurs for examples of intermediate simplicity. For the hardest examples, the memorization scores are low, because even the \textsf{IN}-models could not learn them well. Many hard examples consist of ill formatted text or contained foreign languages. As a result, in \wikien, which contains higher quality texts, the lower bound of the histogram is higher than the other two datasets (Figure~\ref{fig:mem-sim}). Interestingly, the choice of data has a relatively minor effect on memorization: the shape of the memorization histogram is generally consistent across the three datasets; the range of memorization values is only slightly compressed for \wikien.

Figure~\ref{fig:mem-sim} shows the overall distribution of memorization for each training datasets. To obtain a more granular view,
we can also analyze the distributions for texts sourced from individual web domains in \realnews and \cfour, to see whether different data sources display different memorization profiles.
Web domains such as news portals, blogs, and forums differ both stylistically and in how much they reference or even copy from other websites.
Additionally, some domains are represented much more frequently than others in the datasets we studied.
This could lead to considerably different memorization profiles for examples from different domains.

\begin{figure}
    \centering
    
    \begin{minipage}[b]{.69\linewidth}
    \begin{overpic}[width=0.49\linewidth]{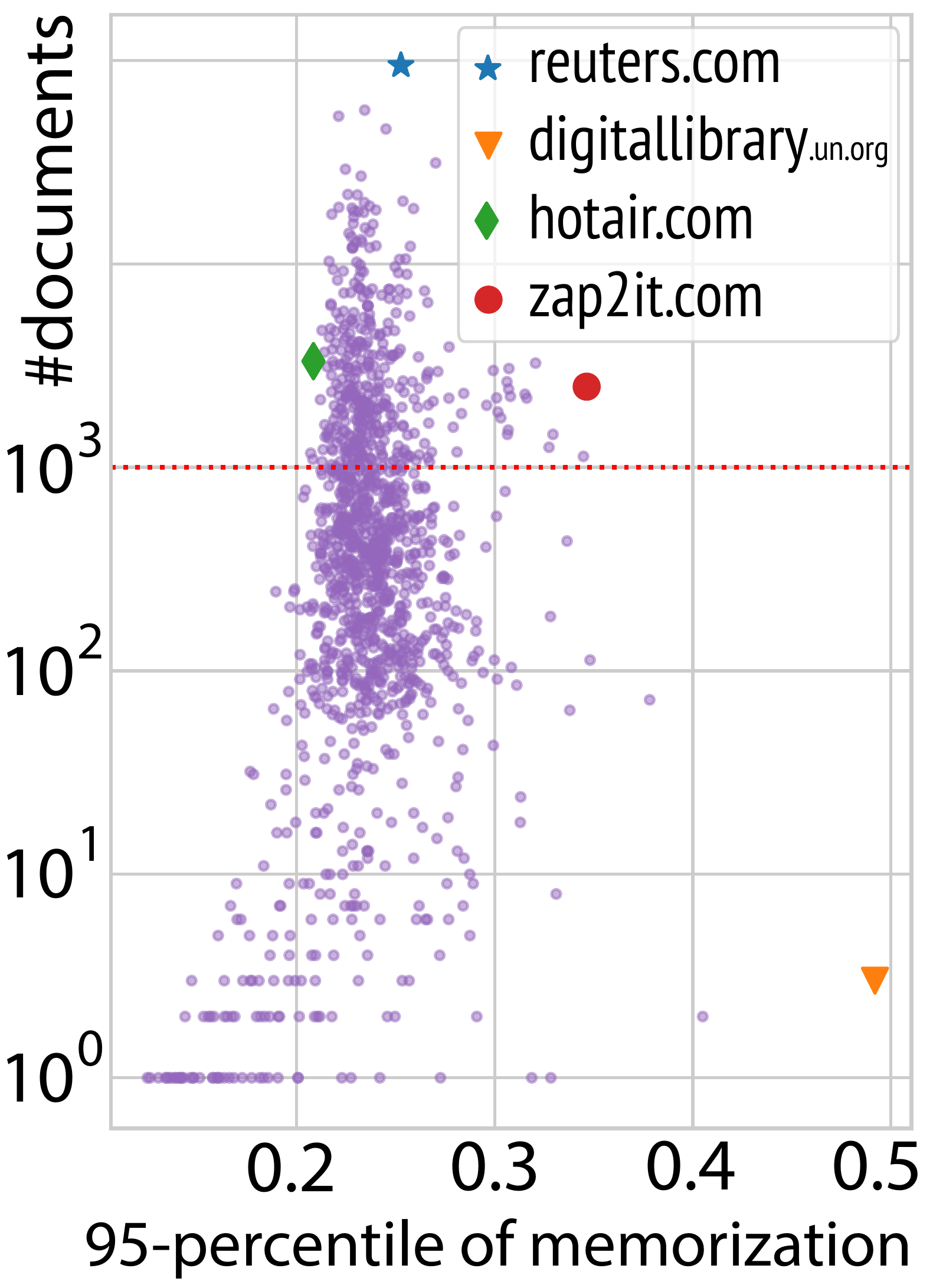}
        \put(40,-5){\scriptsize(a)}
    \end{overpic}
    \begin{overpic}[width=0.49\linewidth]{./figs/largetext/url_scatter_mem-accuracy-c4-v2.pdf}
        \put(40,-5){\scriptsize(b)}
    \end{overpic}
    \end{minipage}
    \begin{minipage}[b]{.29\linewidth}
    \begin{overpic}[width=\linewidth]{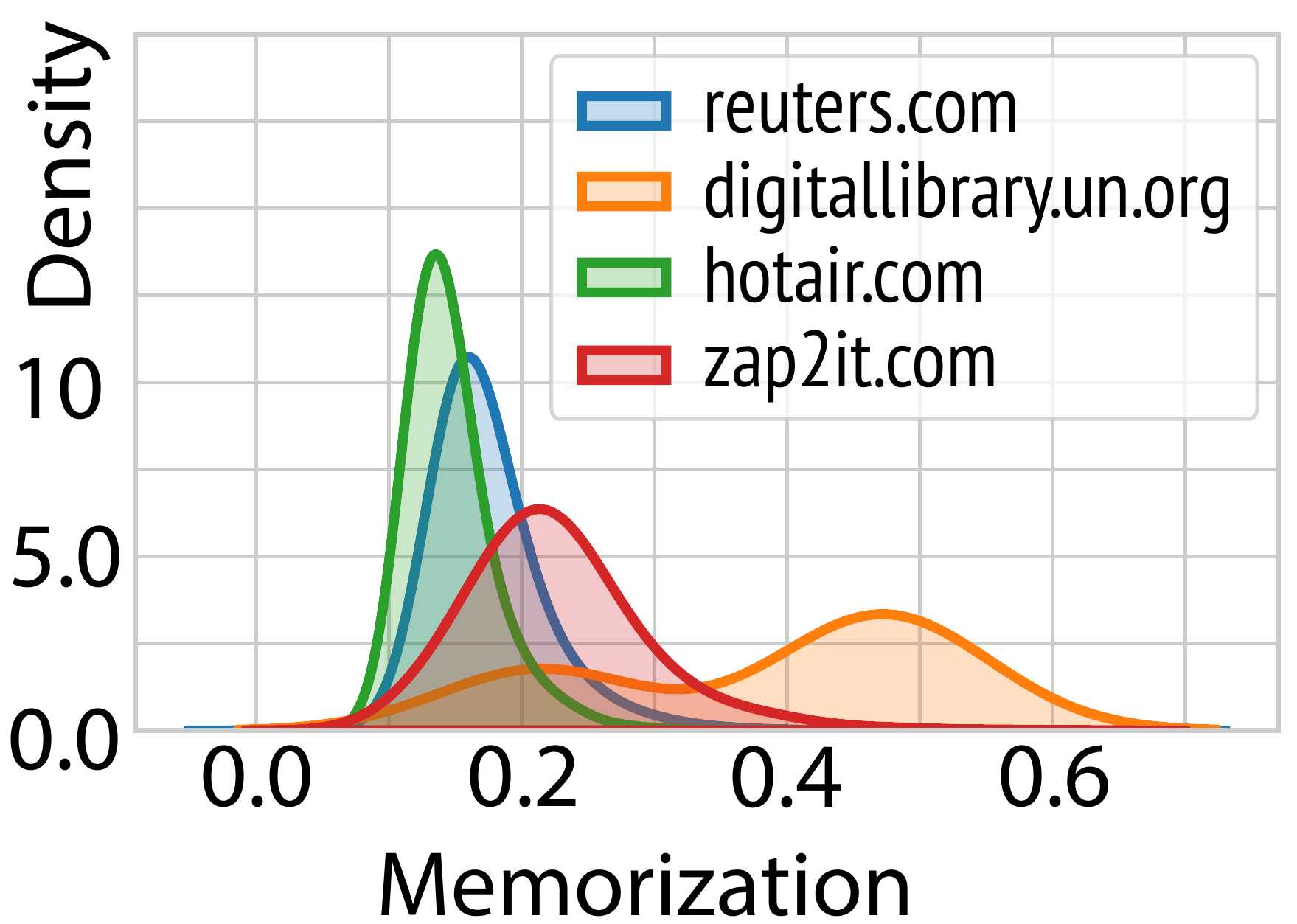}
        \put(55,-5){\scriptsize(c)}
    \end{overpic}
    \vskip15pt
    \begin{overpic}[width=\linewidth]{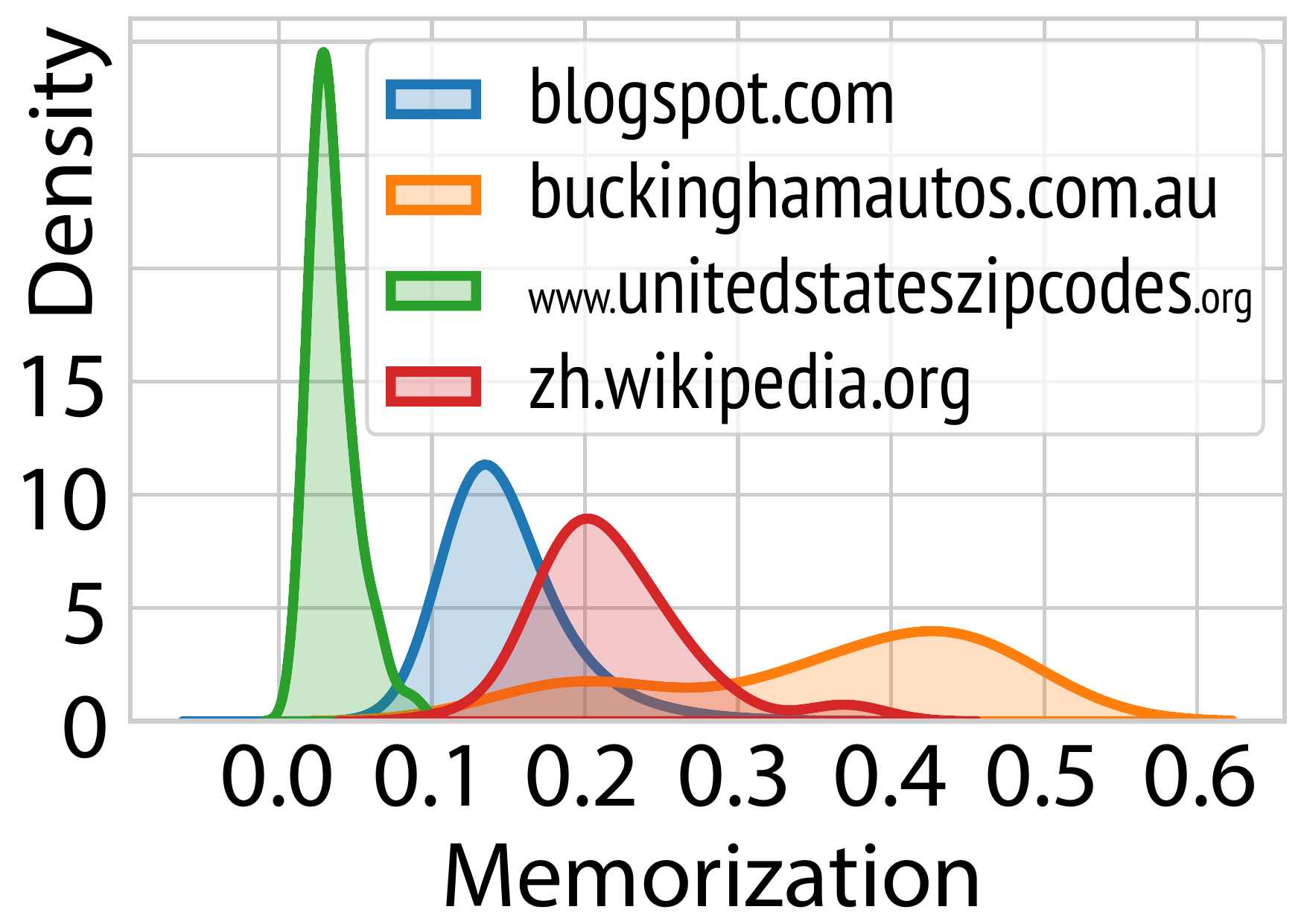}
        \put(55,-5){\scriptsize(d)}
    \end{overpic}
    \end{minipage}
    \caption{\small For each web domain, we plot the 95-percentile of memorization against the number of examples from that domain in \textbf{(a)} \realnews and \textbf{(b)} \cfour. The red dotted line indicates a threshold of a minimum of 1000 articles for \realnews and 50 articles for \cfour. The memorization distributions of a few representative domains are shown for \textbf{(c)} \realnews and \textbf{(d)} \cfour.}
    \label{fig:realnews-url}
\end{figure}

To investigate these effects, we visualize the 95th percentile memorization score in each web domain against the number of examples in that domain for \realnews (Figure~\ref{fig:realnews-url}a) and \cfour (Figure~\ref{fig:realnews-url}b).
\cfour contains many more domain names than \realnews since the latter is collected only from news websites. For both datasets, the domains with a large number of crawled documents show a smaller variance in the 95-percentile values, while ``smaller'' domains depict a wide range of variety in memorization profiles. 
The memorization profiles of a few representative domains are visualized in Figures~\ref{fig:realnews-url}c and \ref{fig:realnews-url}d. The domains we selected for visualization are: the largest domain (blue), the domain with highest 95 percentile memorization (orange), and two domains that have more than 1000 and 50 articles in \realnews and \cfour respectively (green and red). 

In \realnews (Figure~\ref{fig:realnews-url}c), \texttt{reuters.com} contains the largest number of documents but low memorization scores on average. The domain \texttt{digitallibrary.un.org}, the United Nations Digital Library, has high memorization scores potentially because it contains many multilingual documents. We have observed that less frequently occurring tokens, like those in foreign languages or ALL-CAPITAL words tend to cause high memorization.
Similarly, flattened structured data (e.g. tabular texts) also deviates significantly from normal English texts and potentially leads to high memorization, as demonstrated by \texttt{zap2it.com}, a website for TV program listings.
On the other hand, \texttt{hotair.com} is a news commentary website that frequently quotes other major news articles.
This may lead to duplicate text in the dataset which we suspect contributes to its overall lower memorization distribution.

The observations are similar on \cfour: \texttt{blogspot.com} contains a large number of documents in the training set with only moderate amounts of memorization; \texttt{zh.wikipedia.org} and \texttt{buckinghamautos.com.au} have high memorization due to foreign (Chinese) or structured (car sales listings) text; and \texttt{www.unitedstateszipcodes.org} has very low memorization scores because common templates are re-used to generate similar pages for individual zip codes.

\subsection{Number of Models Needed}
To evaluate the impact of a single training example, one may wish to train two models that differ only in that single example. In practice, the stochasticity in a single run of common training algorithms (e.g. SGD) produces too low signal-to-noise ratios to be useful for such estimation. Moreover, leave-one-out estimation means a separate pair of models needs to be trained for each training example, which is computationally costly.
Therefore, we formulated our estimation in Section~\ref{sec:methods} by accumulating statistics from $m$ models independently trained on random training subsets. In our experiments, we set $m=400$. To understand how sensitive our results are to $m$, we analyze the rankings produced by distinct sets of models of size $m$. We vary $m$ from 6 to 192, and partition our set of 400 models into up to 10 sets of $m$ models (e.g. for $m=192$, we construct 2 partitions, and for $m=6$, we construct 10). We then compute the Spearman's R between these partitions to measure the agreement between the rankings produced by each partition. If the rankings are very similar (have Spearman's R close to 1), then this number of models is reliably estimating the true ranking of memorization scores. We plot these Spearman's R values in Figure~\ref{fig:spearman-mem-epochs}a. Even at 96 models, this correlation begins to plateau near 1, lending confidence that 400 models is sufficient for reliable estimation of memorization scores. See Appendix~\ref{app:variance} for more analysis on the sensitivity to $m$.

\subsection{Impact of Number of Training Epochs}
\begin{figure*}
    \centering
    \begin{overpic}[width=.22\linewidth]{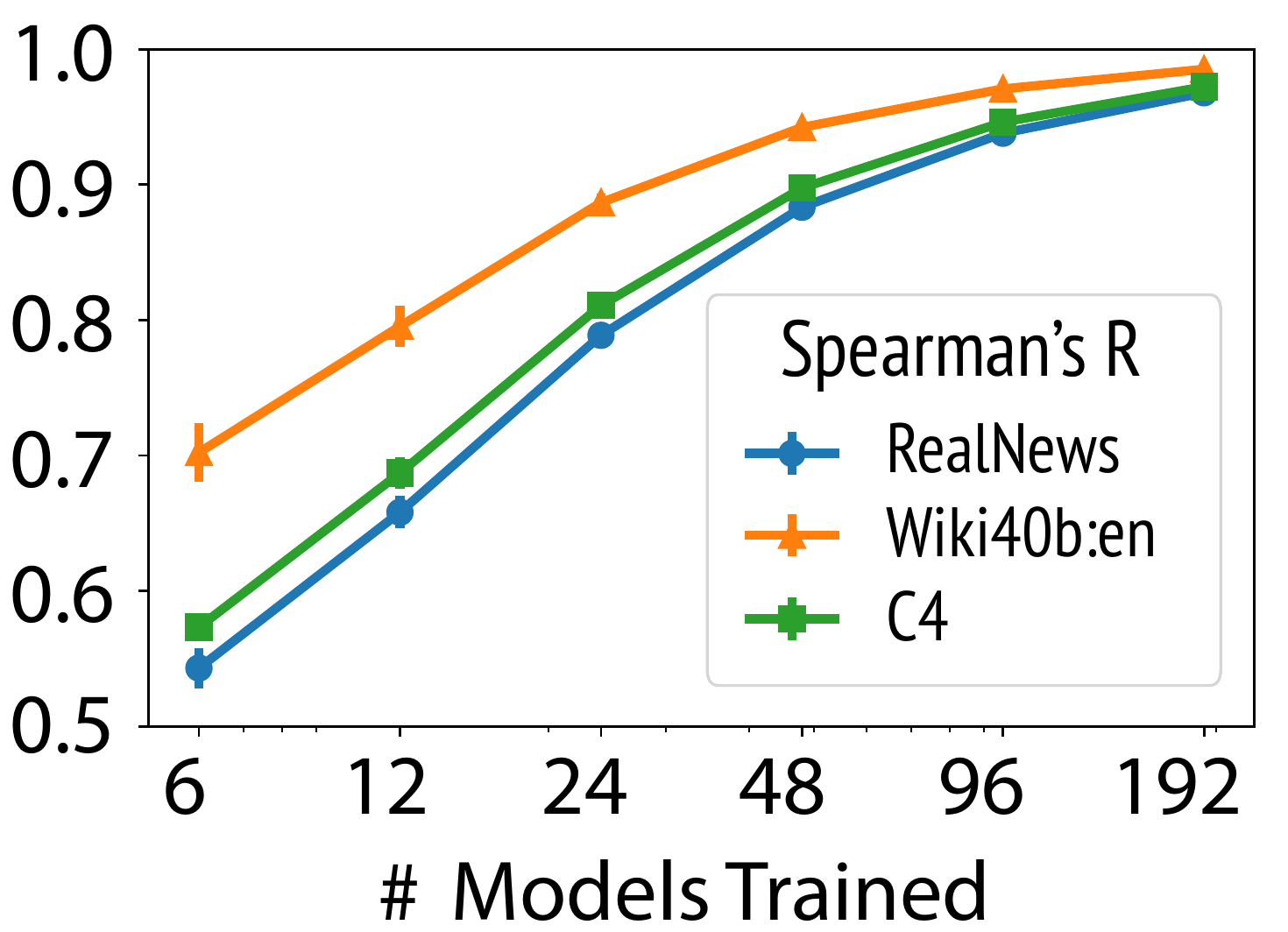}
    \put(5,1){\bfseries\small(a)}
    \end{overpic}
    \begin{overpic}[width=.23\linewidth]{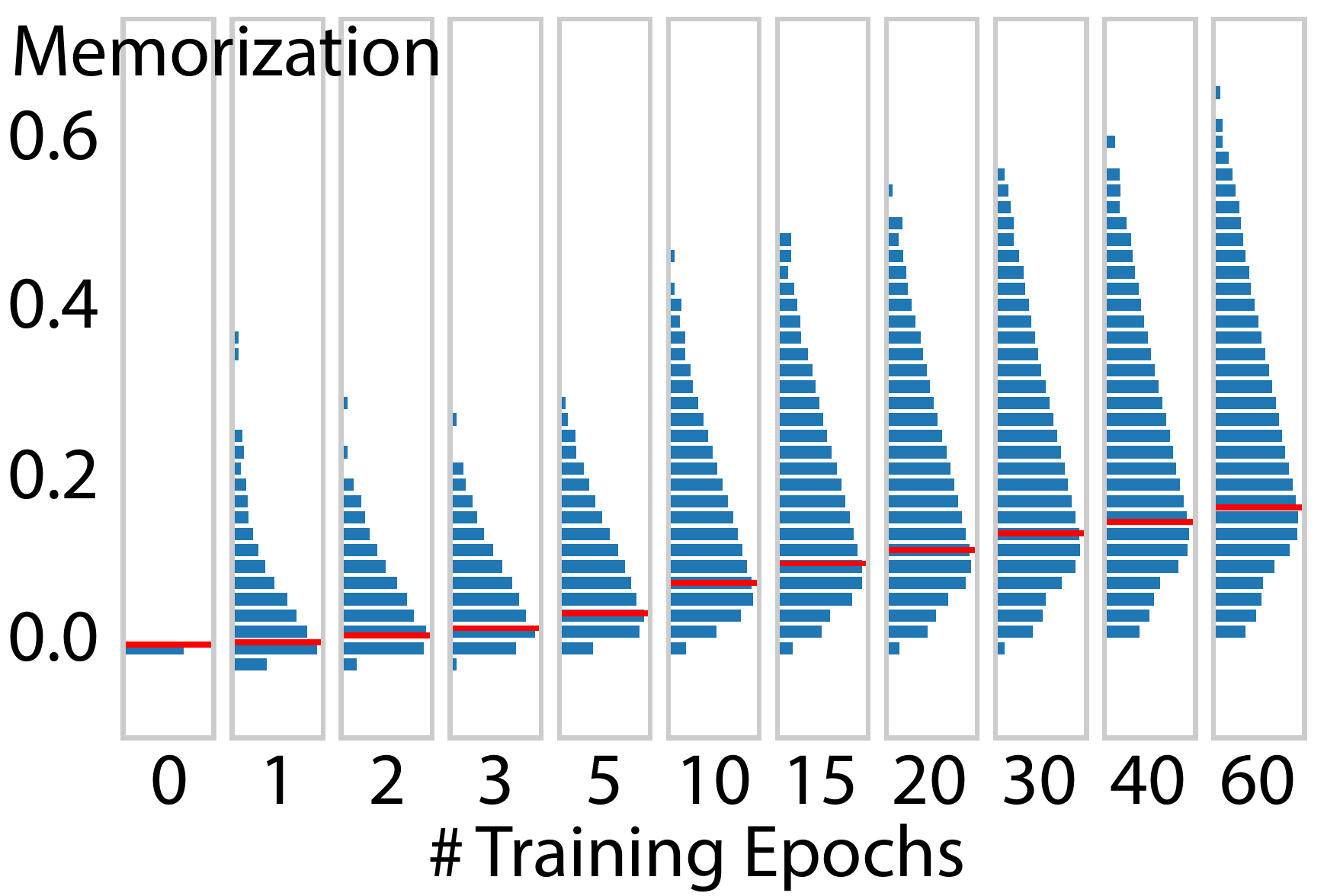}
    \put(1,1){\small\bfseries(b)}
    \end{overpic}
    \begin{overpic}[width=.23\linewidth]{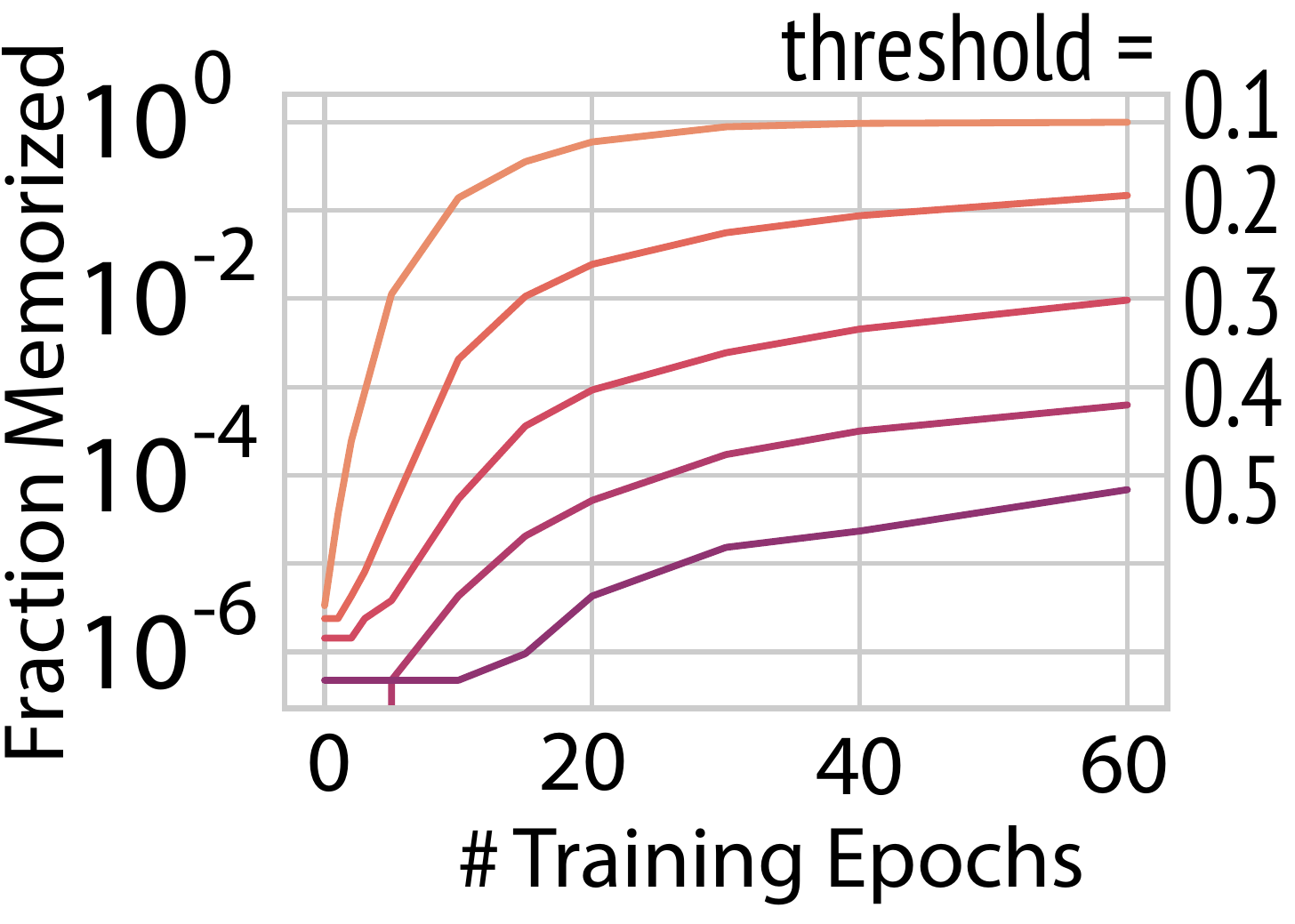}
    \put(5,1){\small\bfseries(c)}
    \end{overpic}
    \begin{overpic}[width=.26\linewidth]{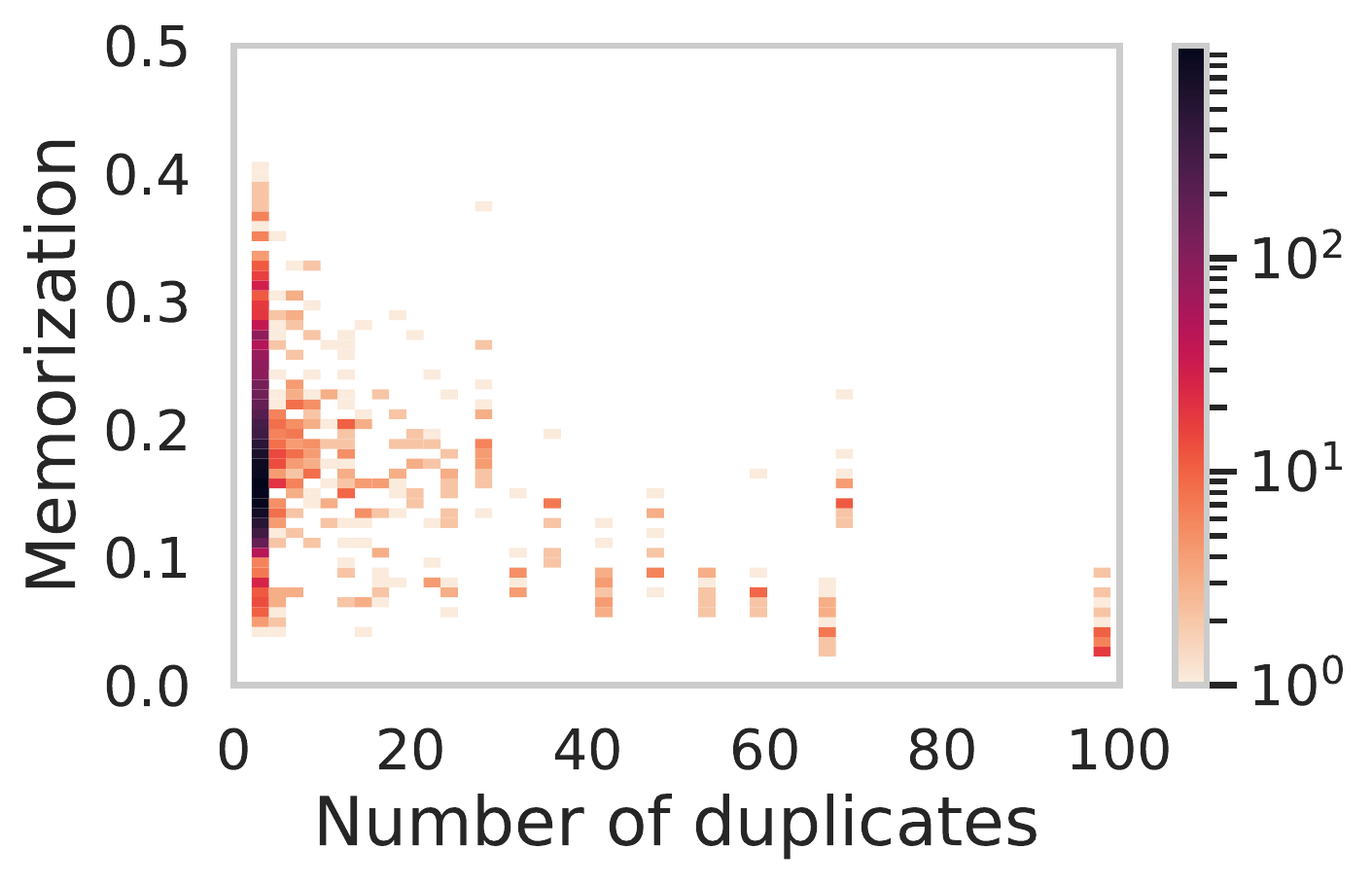}
    \put(5,1){\small\bfseries(d)}
    \end{overpic}
    \caption{\small
    \small
    \textbf{(a)} Spearman's R between memorization rankings from two disjoint sets of $m$ models. The rankings are variable at low numbers of models, but starts to converge at 192 models. All of our other experiments use 400 models. Reported values are averages over up to 10 partitions, with error bars of 10 standard deviations.
    \textbf{(b)} The distribution in memorization of \realnews examples as training progresses.
    \textbf{(c)} The fraction of \realnews{} examples with memorization consistently above the specified threshold as training progresses.
    \textbf{(d)} For \realnews, we plot memorization scores against the number of near-duplicates an example had in the dataset. 
    }
    \label{fig:spearman-mem-epochs}
\end{figure*}

As expected, the overall amount of memorization grows consistently with the number of epochs of training (Figure~\ref{fig:spearman-mem-epochs}b).
This makes sense since training for more epochs increases overfitting.
As training progresses, we also see an increasingly long tail of examples with high memorization scores.
On \realnews{}, about 59\% of examples had consistently increasing memorization scores across all epochs considered.
There were no examples whose memorization decreased in a significant way over training (all observed decreases can be attributed either  to noise or to instability early in training).
Only 0.5\% of examples stayed completely un-memorized with scores which never rose above 0.1, while 85\% of examples had memorization scores which never rose above 0.2.
Figure~\ref{fig:spearman-mem-epochs}c shows the fraction of memorized examples as training progresses, at several thresholds of memorization.
We can see that more training epochs significantly increases memorization.%

\section{Duplicate Text and Memorization}\label{sec:dup}
One of the goals of evaluating counterfactual memorization is to identify examples that have a low number of duplicates yet whose presence versus absence in the training data has a large effect on the model. 
Here, we perform a quantitative study of the (anti-)correlation between duplication and counterfactual memorization compared with the positive correlation between duplication and the ``generation-time memorization'' definitions of memorization used by \cite{2021dedup,carlini2022quantifying, kandpal2022deduplicating}. 

Following the method from \citep{2021dedup}, we first use MinHash~\citep{broder1997resemblance} to identify near-duplicate examples in \realnews train set. We consider a example a duplicate if it has an normalized edit similarity of greater than 0.7 (definition included in Appendix \ref{sec:edit-sim}).
Out of 2.01 million examples, $\sim$38,000 were identified as being a near-duplicate with at least one other example.
Among these frequently-occurring examples, the Pearson correlation between an example's counterfactual memorization score and the number of near-duplicates for that example is -0.39; in other words, memorization does quantitatively decrease when data is repeated more often.

In Figure~\ref{fig:spearman-mem-epochs}d we can see that examples with a large number of near-duplicates have smaller memorization scores. 
Counterfactual memorization primarily differentiates amongst examples with a few number of duplicates. 
This makes sense given that examples with lots of near duplicates would likely have their near duplicates in \textsf{OUT}-models.
This is to be contrasted with ``generation-time memorization'' (discussed in Section \ref{sec:other-defs}) that measures the textual overlap between model generated texts and the training documents.
There, the number of occurrences strongly correlate with the measured memorization~\citep{carlini2020extracting,2021dedup,kandpal2022deduplicating}. 
\textbf{Counterfactual memorization measures a fundamentally different type of memorization from simple textual matching considered in prior work, providing information about how easy or hard a training example is in the context of the rest of the training set.}
In Table~\ref{tab:realnews-egs} we can see this effect qualitatively: sequences with near-duplicates in the training set tend to have low counterfactual memorization (as expected) . 

\section{From Memorization to Influence}
\label{sec:influence}

Counterfactual memorization identifies training examples that contain rare information not conveyed by other examples.
A natural question to ask is whether a model would leak the information in a memorized example during inference.
Previous paper studies membership inference attack~\citep{shokri2017membership, sablayrolles2019whitebox, long2020pragmatic} where an attacker tries to figure out if a particular example exists in the training set. In this paper, we consider standard model evaluation without adversarial attackers, and quantify ``does seeing a particular training example strongly influence the prediction on a validation example?'' Another way of asking this is if a single example in the training set has an large and over-representative impact on the prediction of a validation example. We answer these questions by measuring \emph{counterfactual influence} with a formulation adapted from \citet{feldman2020neural}:

\begin{definition}[Counterfactual Influence]
Given a training algorithm $A$ that maps a training set $D$ to a trained model, and a performance measure $M$, the counterfactual influence of a training example $x\in D$ on another example $x'$ is 
\begin{equation}
    \infl(x\Rightarrow x') \triangleq
    \mathbb{E}_{S\subset D,x\in S}[M(A(S), x')] - 
     \mathbb{E}_{S\subset D,x\not\in S}[M(A(S), x')],
\end{equation}
where $S$ is a subset of training examples sampled from $D$. The expectation is taken with respect to the random sampling of $S$, as well as the randomness in the training algorithm $A$. Here $x'$ can be an example from the validation set or test set, a generated example or a training example. 
\end{definition}
An empirical estimation of the influence can be computed similarly to counterfactual memorization by uniformly sampling $m$ subsets $S_1,\ldots,S_m$ from $D$, where $|S_i| = r|D|$, and calculating 
\begin{equation}
    \widehat{\infl}(x\Rightarrow x') \triangleq
    \operatorname*{mean}_{i:x\in S_i}[M(A(S_i), x')] - 
    \operatorname*{mean}_{i:x\not\in S_i}[M(A(S_i), x')].
    \label{eq:infl-estimate}
\end{equation}

This measures how much a training sample $x$'s presence influences the prediction of a different example $x'$. Note, $\mem(x)=\infl(x\Rightarrow x)$, i.e., counterfactual memorization is self influence.

\begin{figure}
    \centering
    \begin{overpic}[width=.38\linewidth]{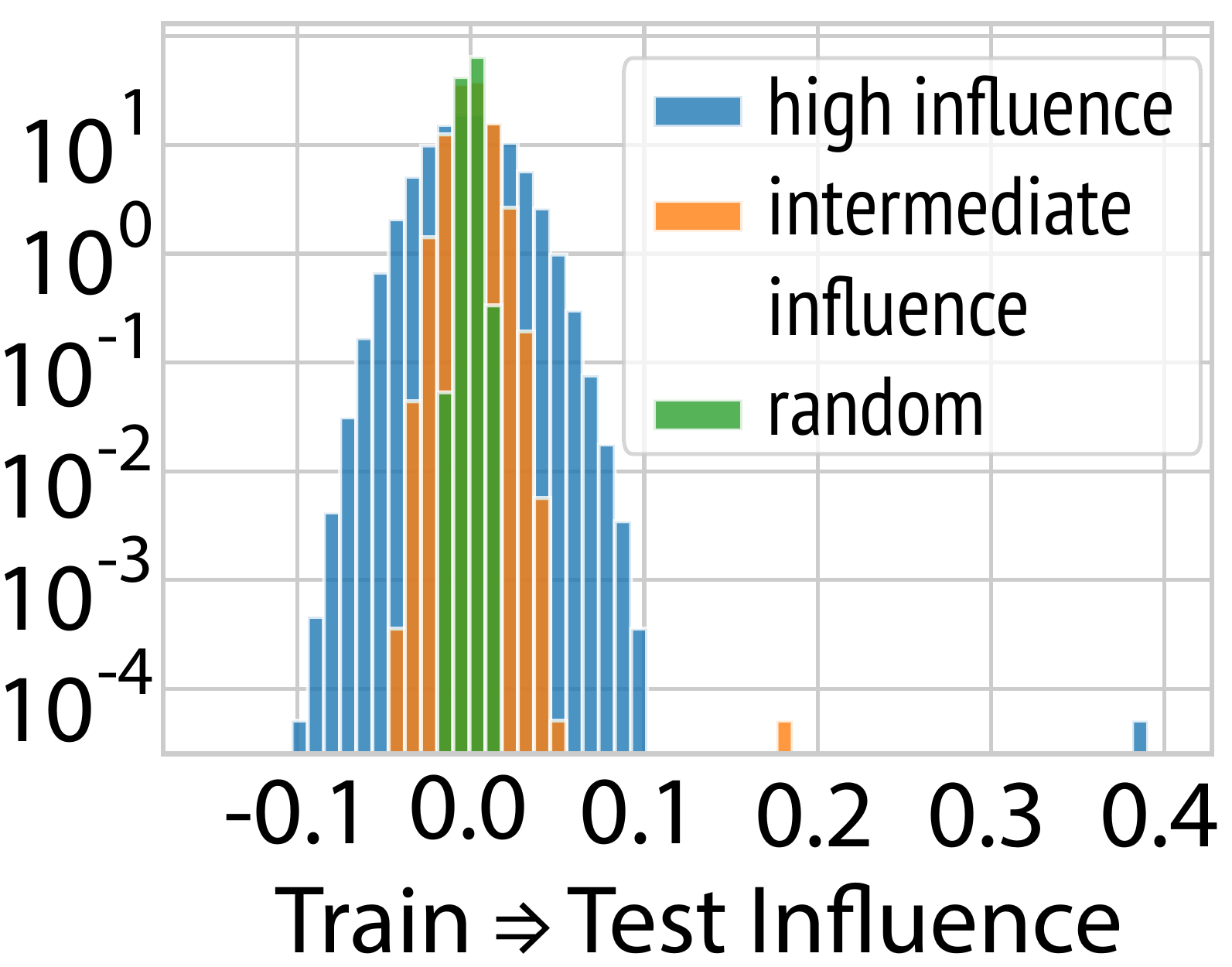}
    \put(30,-7){\small(a) Histogram of influence}
    \end{overpic}
    \begin{overpic}[width=.30\linewidth]{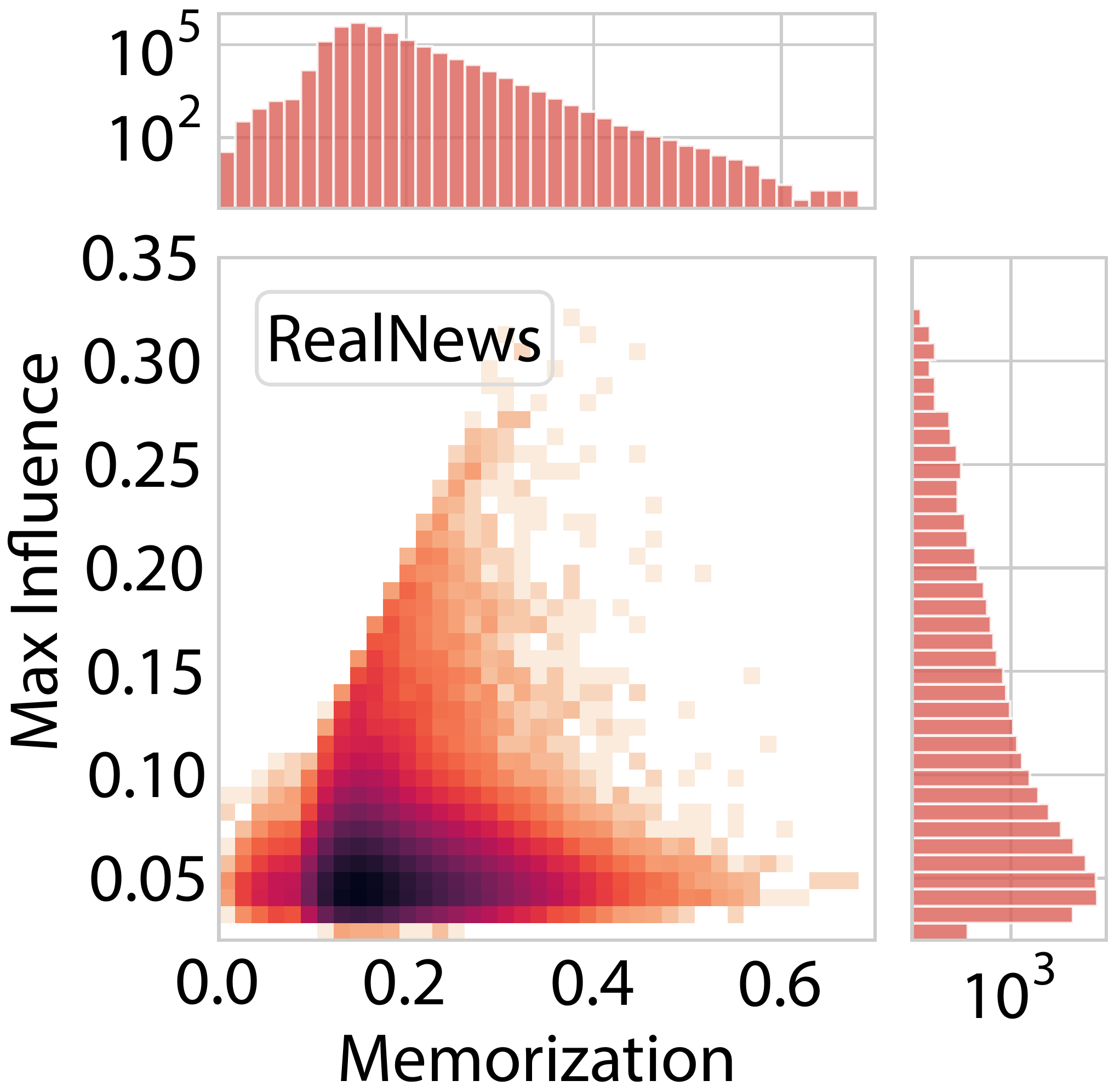}
    \put(20,-7){\small(b) Joint distribution of influence and memorization}
    \end{overpic}
    \includegraphics[width=.30\linewidth]{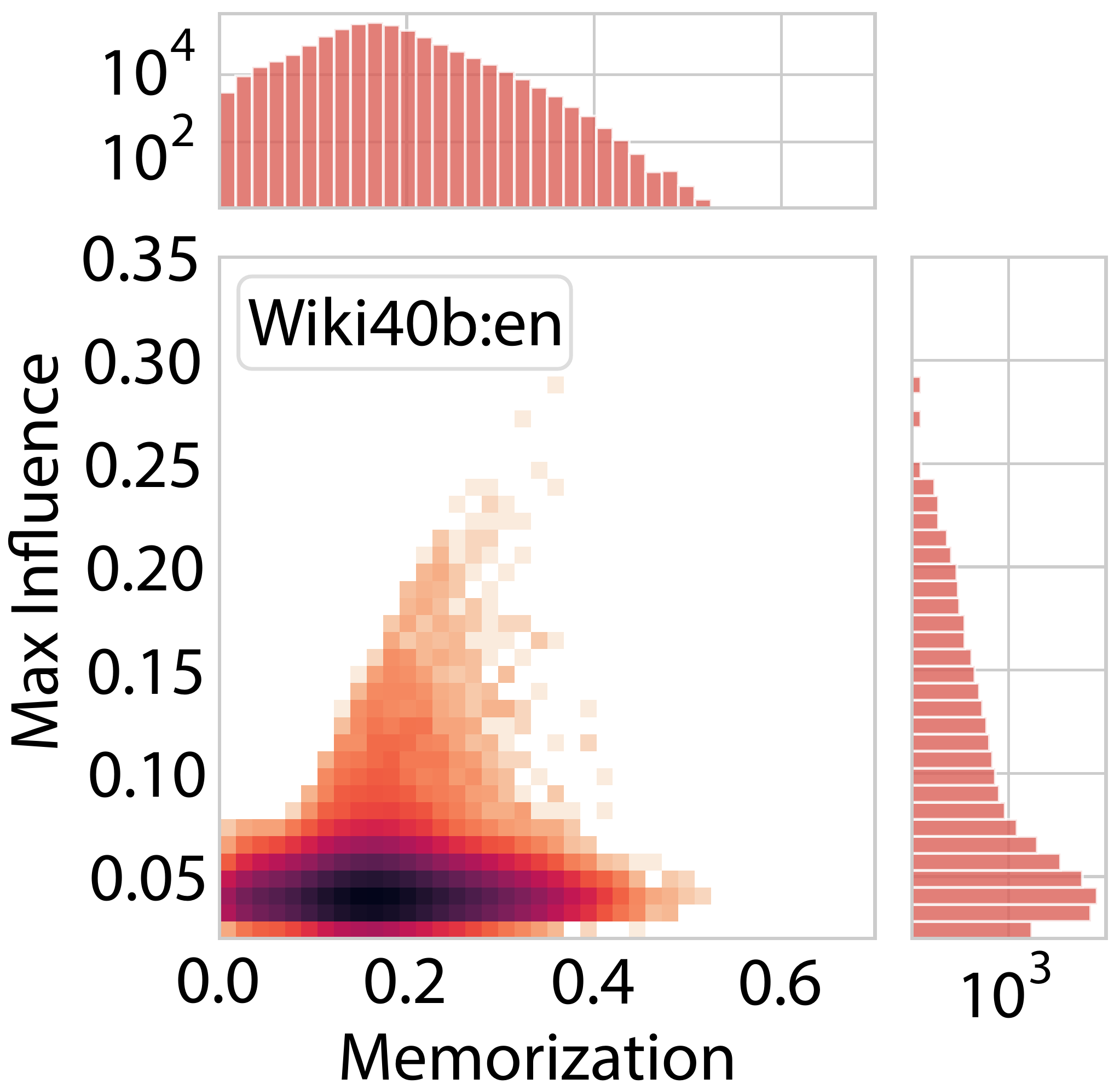}
    \vskip10px
    \caption{\small (a) Histogram of the influence of all the training examples on a specific test example for three different test examples on \realnews. The blue and orange examples have high and intermediate influence from some training examples, as indicated by the outlier values to the right of the each histogram plot. The green one is a random example, where the influence from all individual training examples are close to zero. (b) The joint distribution of the memorization score of each training example and its maximum influence on any validation set example. The histograms are in log scale to better visualize the tail of the distributions. C4 shown in Figure~\ref{fig:mem-infl-full}.}
    \label{fig:mem-infl}
\end{figure}

\textbf{Influence on Examples of the Validation Set}.
With the same models trained for estimating memorization, we can estimate the counterfactual influence on the validation set according to Equation \eqref{eq:infl-estimate}. For each example in the validation set, we can estimate the influence on it from each training example. Figure~\ref{fig:mem-infl}a shows the distribution of influence from all training example on three different examples from the validation set.
The green example was randomly chosen and represents the behavior for most validation examples: it receive close-to-zero influence from all the (individual) training examples.
The blue and orange examples were sampled to have high and intermediate maximum influence. Each of them has one (or a few) strong influencer from the training set, as indicated by the bars to the right of the histogram. They also only receive tiny influence from all the rest of the training examples, though the variance of influence is larger than for the green example.

Intuitively, most training examples will have small influence on validation set examples because the models learn distributional patterns shared across many training examples, and \emph{individual} training examples tend to have insignificant influence here.
However, a training example $x$ with high counterfactual memorization contains rare information that are not shared with other examples.
Therefore, if a validation set example $x'$ contains similar information, $\infl(x\Rightarrow x')$ could be large. Figure~\ref{fig:mem-infl}b shows the relationship between memorization and influence by plotting $\mem(x)$ of each training example $x$ against its maximum influence $\max_{x'}\infl(x\Rightarrow x')$ on $x'$ across the validation set. 

Consistent with our intuition, examples with small memorization scores have small max-influence scores. Larger influence scores on the validation set generally requires larger memorization scores of the training example itself. However, not all training examples with large memorization scores lead to large influence scores. In particular, the max-influences drop significantly for examples with memorization larger than 0.4. One potential reason is that many examples with very high memorization are simply low quality text, so memorization is required in order to learn them, but they do not encode anything interesting that could influence a validation example. On the other hand, even if a memorized example encodes some rare and useful information, the max-influence could still be low because the validation set does not contain a relevant document. This is especially true given that all datasets have considerably smaller validation sets than training sets.

Table~\ref{tab:infl-egs-realnews} shows train-validation example pairs from \realnews sampled at different influence value ranges. We found that the train-validation pairs with the highest influence are almost identical, except some superficial differences,
such as different handling of quotation / em dash marks. As we move to intermediate influence ranges, we commonly found reports on the same events.
Large paragraphs of identical text indicate that one document might be citing the other or both citing from a third party.
At low influence, two types of correlations are commonly observed: 1) templated texts with high similarity---the reason for a low influence is that there are many similar training examples that split the influence; 2) superficially related documents due to a shared prefix such as \emph{ST. CLOUD -- This week in our ``Behind the Scenes'' series on WJON} or a shared substring of some common knowledge like \emph{FSIS, the Centers for Disease Control and Prevention}.
Due to high signal-to-noise ratio, here were no noticeable relationships in the document pairs with influence scores below 0.02. 

\begin{table}[]
    \caption{\small Train-validation example pairs of \realnews sampled at a variety of influence levels.
    \egomit indicate text omitted for brevity. Differences in each document pair are \hlc{highlighted yellow}.}
    \label{tab:infl-egs-realnews}
    \centering\tiny
    \begin{tabularx}{\linewidth}{@{}ccX@{}}
    \toprule
    Index & Estim. & Text \\\midrule
    \makecell[tc]{Validation\\1334662} & \makecell[tc]{$\infl$\\\textbf{0.3780}} & \eglink{http://www.metro.us/news/archaeologists-vs-robbers-in-israel-s-race-to-find-ancient-scrolls/kZgpfb---HhAktviTsXh_U3HpnRICjA} \hfill\egcomment{Identical URL except with the \texttt{http://} protocol instead of \texttt{https://}. The text is identical to the training example.}
    \\\rowcolor{TableHL}
    \makecell[tc]{Train\\2077314} & \makecell[tc]{$\mem$ \\ 0.3780} & \eglink{https://www.metro.us/news/archaeologists-vs-robbers-in-israel-s-race-to-find-ancient-scrolls/kZgpfb---HhAktviTsXh_U3HpnRICjA} By Ari Rabinovitch TZEELIM VALLEY, Israel (Reuters) - The disposable paper face masks offer little protection from the clouds of dust that fill the cliffside cave where Israeli archaeologists are wrapping \egomit insight into how people lived 2,000 to 8,000 years ago. (Editing by Jeffrey Heller/Jeremy Gaunt)
    \\\midrule
    \makecell[tc]{Validation\\838341} & \makecell[tc]{$\infl$\\\textbf{0.1209}} & \eglink{https://www.dailymail.co.uk/wires/ap/article-3670409/Retired-pope-offers-assessment-papacy.html} VATICAN CITY\hlc{ (AP) }— Emeritus Pope Benedict XVI is offering a first-ever papal assessment of his own pontificate in a book that recounts his decision to resign, his surprise at his successor and his attempts to dismantle what he calls the Vatican's "gay lobby." "Benedict XVI: The Final Conversations," is due out in September, the latest \egomit Vatican career by simply spreading gossip that he was gay. \hlc{\_\_\_} \hfill\egcomment{Different websites, but (almost) identical report.}
    \\\rowcolor{TableHL}
    \makecell[tc]{Train\\614881} & \makecell[tc]{$\mem$\\0.1650} & \eglink{https://www.deseretnews.com/article/765687763/Retired-pope-offers-first-ever-assessment-of-his-own-papacy.html} VATICAN CITY\hlc{ }— Emeritus Pope Benedict XVI is offering a first-ever papal assessment of his own pontificate in a book that recounts his decision to resign, his surprise at his successor and his attempts to dismantle what he calls the Vatican's "gay lobby." "Benedict XVI: The Final Conversations," is due out in September, the latest \egomit Vatican career by simply spreading gossip that he was gay. \hlc{Follow Nicole Winfield at www.twitter.com/nwinfield}
    \\\midrule
    \makecell[tc]{Validation\\682107} & \makecell[tc]{$\infl$\\\textbf{0.0673}} & \eglink{https://www.pasadenastarnews.com/2009/09/16/ducks-ushering-in-a-new-era-for-2009-10-season/} ANAHEIM – On a night when Francois Beauchemin had two assists in Toronto\hlc{, }and Chris Pronger blocked six shots in Detroit,\hlc{ }13,869 Ducks\hlc{ }fans \hlc{might have been lost without their }programs \hlc{on Wednesday night}. It’s only the first game of the preseason, but a new era has clearly begun. A group of mostly newcomers in Ducks uniforms beat Phoenix 3-2 in a shootout\hlc{ on Wednesday at Honda Center. A }familiar face made the biggest impact\hlc{, however,} as Bobby Ryan scored two goals\hlc{. }Ryan also scored two goals in the Ducks’ preseason opener last year \egomit \hfill\egcomment{Different websites on the same event with slightly different wordings.}
    \\\rowcolor{TableHL}
    \makecell[tc]{Train\\494435} & \makecell[tc]{$\mem$\\0.1439} & \eglink{https://www.dailybreeze.com/2009/09/17/newcomers-help-lead-ducks-to-exhibition-win/} ANAHEIM – On a night when Francois Beauchemin had two assists in Toronto\hlc{ }and Chris Pronger blocked six shots in Detroit, \hlc{the }13,869 Ducks\hlc{' }fans 
    \hlc{who showed at at Honda Center Wednesday needed }programs \hlc{to identify the players on their favorite team}. It’s only the first game of the preseason, but a new era has clearly begun. A group of mostly newcomers in Ducks uniforms beat Phoenix 3-2 in a shootout\hlc{, although a }familiar face made the biggest impact as Bobby Ryan scored two goals\hlc{ in regulation and another in the shootout. }Ryan also scored two goals in the Ducks’ preseason opener last year \egomit
    \\\midrule
    \makecell[tc]{Validation\\1165799} & \makecell[tc]{$\infl$\\\textbf{0.0360}} & \eglink{https://kdvr.com/2019/03/14/more-than-70000-pounds-of-butterball-turkeys-recalled-because-of-potential-salmonella/} x \hlc{More than 7}0,000 pounds of \hlc{Butterball }turkey \hlc{recalled because of potential} salmonella WASHINGTON — The U.S. Department of Agriculture’s Food Safety and Inspection\hlc{s }service\hlc{s }announced\hlc{ on }Wednesday \egomit They were shipped to nationwide retail and institutional locations\hlc{. RELATED: View the full recall} FSIS, the Centers for Disease Control and Prevention and \egomit \hfill\egcomment{Different websites reporting the same event, one embeded a lot more information than the other.}
    \\\rowcolor{TableHL}
    \makecell[tc]{Train\\1571976} & \makecell[tc]{$\mem$\\0.2094} & \eglink{https://wreg.com/2019/03/14/butterball-recalls-nearly-80000-pounds-of-turkey-after-salmonella-cases/} x \hlc{Butterball recalls nearly 8}0,000 pounds of turkey \hlc{after} salmonella \hlc{cases }WASHINGTON — The U.S. Department of Agriculture’s Food Safety and Inspection\hlc{ }service\hlc{ }announced\hlc{ }Wednesday \egomit The\hlc{ raw ground turke}y \hlc{was produced on July 7, 2018. The following products under recall }were shipped to nationwide retail and institutional locations\hlc{: 48-oz. plastic wrapped tray containing ``BUTTERBALL everyday Fresh Ground Turkey WITH NATURAL FLAVORING (85\% LEAN/15\% FAT)'' with sell or freeze by date of 7/26/18, lot code 8188, and UPC codes 22655-71555 or 22655-71557 represented on the label. 48-oz. plastic wrapped tray containing ``BUTTERBALL everyday Fresh Ground Turkey WITH NATURAL FLAVORING (93\% LEAN/7\% FAT)'' with sell or freeze by date of 7/26/18, lot} \egomit \hlc{labels here.}  FSIS, the Centers for Disease Control and Prevention and \egomit
    \\\bottomrule
    \end{tabularx}
\end{table}

Influence turns out to be an effective tool for analyzing and attributing the model predictions at test time: for predictions that rely on information obtained by (counterfactual) memorization, we can identify exactly which training example provided such information. Our observation of near-duplicated training-validation document pairs is consistent with recent studies that identifies data contamination in large Internet crawled text corpus~\citep{2021dedup,dodge2021documenting}.

\textbf{Influence on Generated Texts}.
The influence estimation is not restricted to the validation set. We can also estimate influence on generated examples.
In this section, we evaluate on the publicly released generations from the Grover models ~\citep{zellers2019defending}  %
trained on \realnews. Specifically, we take the generations from Grover-Mega (p=0.96), a 1.5-billion-parameter model trained on the \realnews dataset. Comparing with the train-validation influence in Figure~\ref{fig:mem-infl}b, the histogram (c.f. Figure~\ref{fig:grover-infl} in Appendix.) decays faster as max-influence grows. Moreover, the value range of max-influence is also twice smaller. The reason that we did not find a lot of highly influenced generated examples are two fold: 1) there are only 24,576 generation in the public release, which is much fewer than the validation examples. As a result, the corresponding example of many memorized training examples do not get sampled in the generations. 
For comparison, previous work~\citep{carlini2020extracting, 2021dedup} generated 100,000+ examples to identify memorization in generation. These approaches also count duplicates in the training set, which counterfactual memorization filters out. 2)
The Grover model was trained on the full \realnews training set, while we have restricted our analysis to the first 2M training examples. There could be potentially more high influence training examples that are missed in our calculation.

\section{Summary and Discussion}
\label{sec:summary}
We studied memorization in neural language models. 
We formulated a notion of \emph{counterfactual memorization} as a tool that can systematically ignore ``common'' memorization such as general knowledge (e.g. ``Paris is a city in France'') and captures memorization of rare, specific information (e.g. description of a specific episode of event) present in the training examples. We conducted experiments on three commonly used text corpus in language modeling and found memorization in all of them. We further analyze the per-domain memorization profiles for Internet-crawled data, and found that different sources could have substantially different memorization profiles. 

Furthermore, we analyzed how memorized training examples could impact the model predictions at test time via \emph{counterfactual influence}. We found that for examples from both the validation set and the model generated texts, the model predictions could be drastically different depending on the presence or absence of a particular training example with high memorization.

\paragraph{Limitations.}
This study mainly focus on English datasets. While we expect the characterization of memorization would be similar when evaluated on corpus of other (natural) languages, new patterns might be observed on multilingual data or more structured domains such as programming languages.

Both the neural language models and training sets used in this work are orders of magnitude smaller than modern standards such as GPT-3~\citep{NEURIPS2020_1457c0d6}, GPT-4~\citep{openai2023gpt} and PaLM-2~\citep{palm2}. Moreover, we only conducted preliminary investigation of the dynamics of counterfactual memorization during training. Although our experiments effectively estimated and detected memorization, we suspect more interesting examples might emerge if larger, more capable models are analyzed. For example, currently when the information from a memorized training example is leaked in the prediction of a strongly influenced test example, it can usually be explained by a high text overlap between the training and test examples. For models with deeper understanding of languages, we suspect that strong influence could be observed even between documents that have no direct text overlap but that encode similar semantic information.

In order to test this, it will be necessary to scale our framework to larger models and datasets. Moreover, it will be necessary to construct datasets where semantically similar but textually different document pairs exist. One potential source to construct such datasets would be \emph{versioned} Wikipedia articles--two versions of the same article with large time span or edit distance may contain semantically similar (but paraphrased) information.
Such a dataset of paraphrased text pairs would be more broadly useful to understand the ability of different models to disentangle text content and form---by measuring the influence of one piece of text on a paraphrased piece of text.

Counterfactual memorization enables us to identify examples that whose presence or absence has a large impact on the model and the model's ability to score and generate other text. 
The privacy risk for this is low since in order to perform this analysis, one would need to already have access to the dataset and the ability to train models.

\paragraph{Acknowledgments.}
The authors would like to thank Samy Bengio, Christopher A. Choquette-Choo, Ethan Dyer, Michael C. Mozer, Behnam Neyshabur, Andrew Nystrom, and Hanie Sedghi for constructive discussions and feedback. The authors would like to thank Andrew Nystrom for assistance with MinHash-based near-duplicate detection.

\bibliographystyle{plainnat}
\bibliography{refs}

\clearpage \appendix

\section{Difference Between Counterfactual and Generation-Time  Memorization}\label{sec:other-defs}
Many definitions of memorization operate at \textit{generation-time}: a sequence of generated text is marked as memorized if a sufficient amount of overlap is found in the training dataset ~\citep{carlini2020extracting}.
When the training data is not available, heuristic-based methods comparing language model perplexities are used to predict whether a generation contains memorized content \citep{carlini2019secret, thakkar2020understanding, thomas2020investigating,carlini2020extracting,Zanella_B_guelin_2020}.
One difficulty with these approaches is that generation-time instances of memorization are strongly correlated with the number of similar or near-duplicate examples in the training set. As observed in \citet{2021dedup}, large clusters of near-duplicated examples do exist in common language datasets, dominating memorization detected in generated text.
Generation-time methods for measuring memorization are forced to design heuristics to avoid simply identifying these uninteresting instances of memorization.

In contrast, the counterfactual memorization we study in this paper handles the issue of near-duplicates automatically without the need for heuristics.
For a training example, $x$, with many near-duplicate copies in the training set, $\mem(x)$ will be small (because other samples $x' \approx x$ will be present in the training dataset whether or not $x$ is).
This does not mean that counterfactual memorization is the opposite of generation-time memorization.
An example, $x$, with high $\mem(x)$ may have a high chance of being generated if a model is appropriately prompted, despite and possibly \textit{because} it is rare, and thus the example is considered memorized by both definitions.
In summary, generation-time memorization measures the chance a model will directly copy from training examples, while counterfactual memorization aims to discover rare information that is memorized.

\section{Average Accuracy of \inmodels vs \outmodels}

\begin{figure}
    \centering
    \begin{overpic}[percent,width=.32\linewidth]{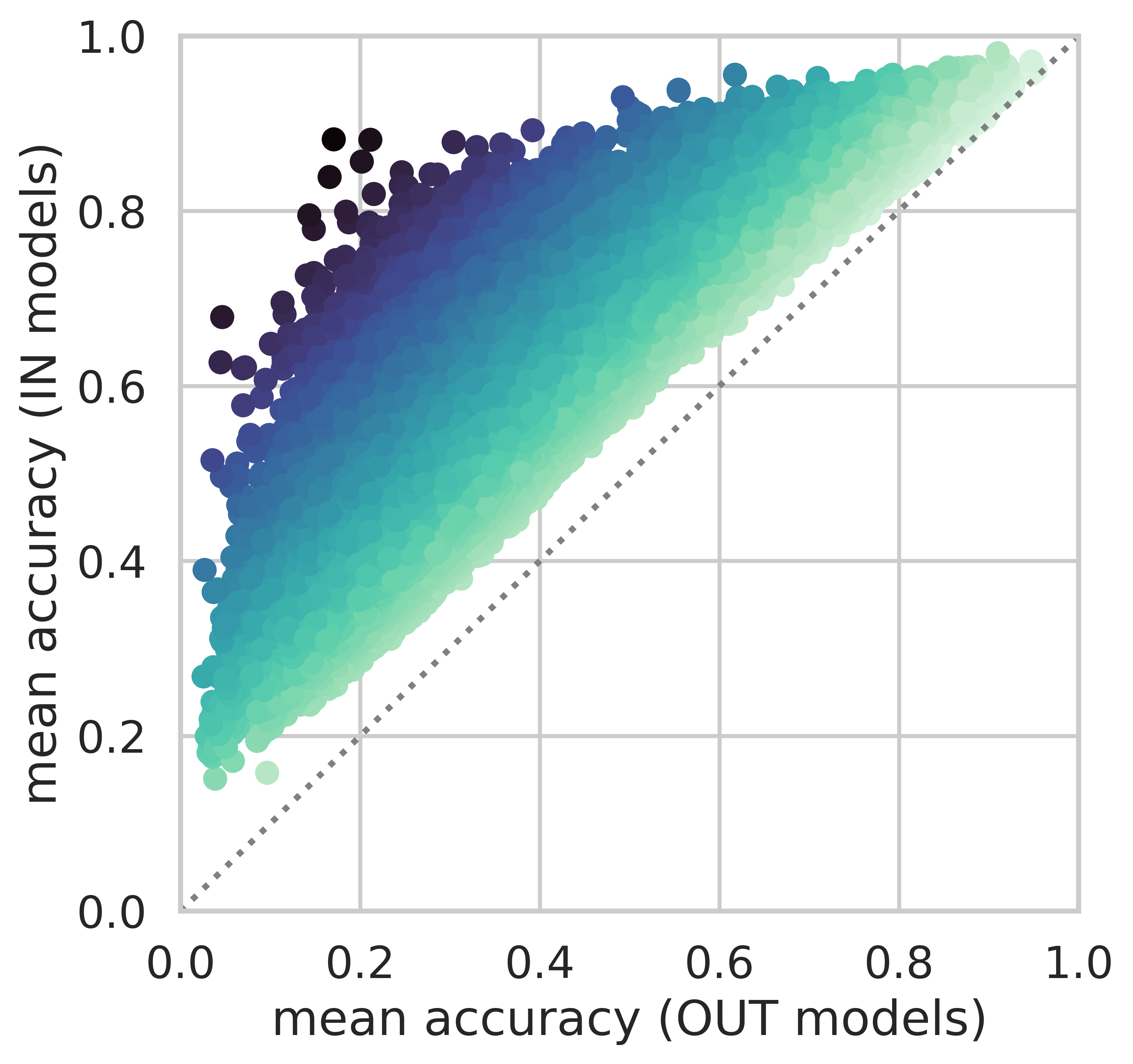}
            \put(45,20){\fbox{\scriptsize\realnews}}
    \end{overpic}
    \begin{overpic}[percent,width=.32\linewidth]{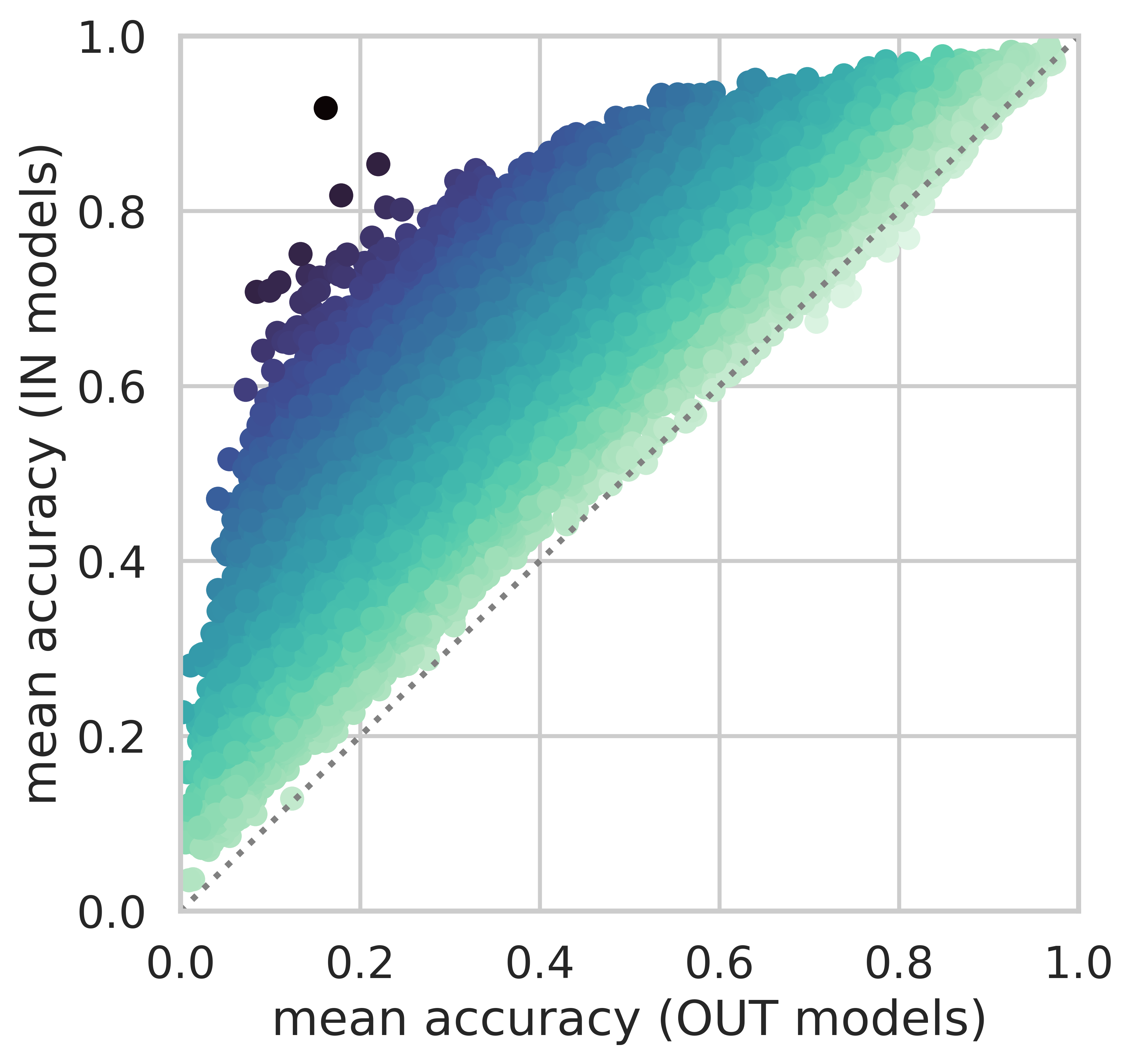}
            \put(69,20){\fbox{\scriptsize\cfour}}
    \end{overpic}
    \begin{overpic}[percent,width=.32\linewidth]{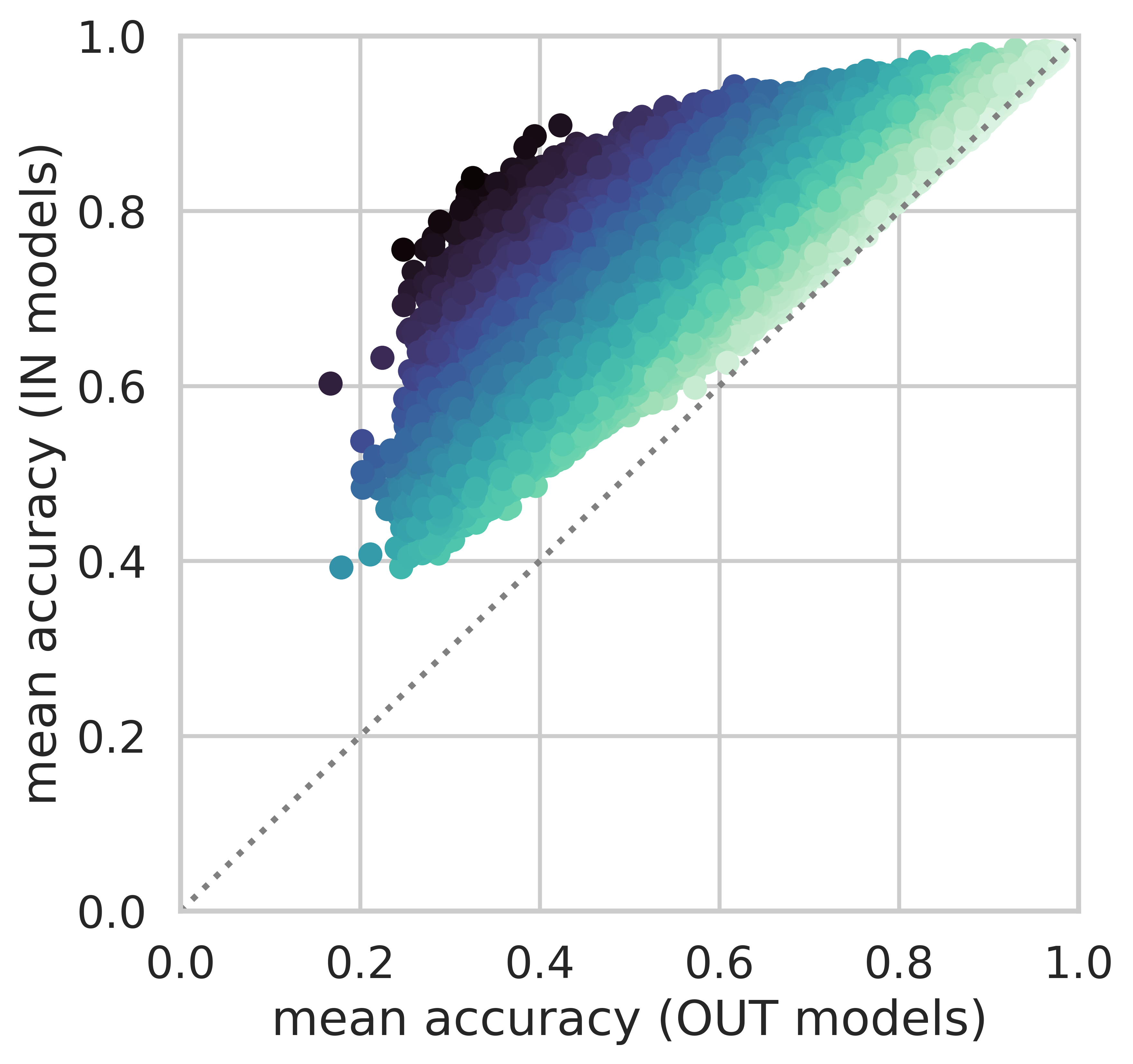}
            \put(40,20){\fbox{\scriptsize\wikien}}
    \end{overpic}
    \vskip-9pt
    \caption{Per-token accuracy of training examples evaluated on \inmodels vs \outmodels.}
    \label{fig:scatter}
\end{figure}
Figure~\ref{fig:scatter} compares the per-token accuracy between the \inmodels and \outmodels for the training examples from three different datasets. Counterfactual memorization is estimated by taking the difference between the average \textsf{IN}-accuracy and the average \textsf{OUT}-accuracy. Thus, the examples closer to the upper left corner are more counterfactually memorized, while the examples near the diagonal are not.

\begin{figure}
    \centering
    \includegraphics[width=.32\linewidth]{./figs/largetext/mem_infl-real_news.pdf}
    \includegraphics[width=.32\linewidth]{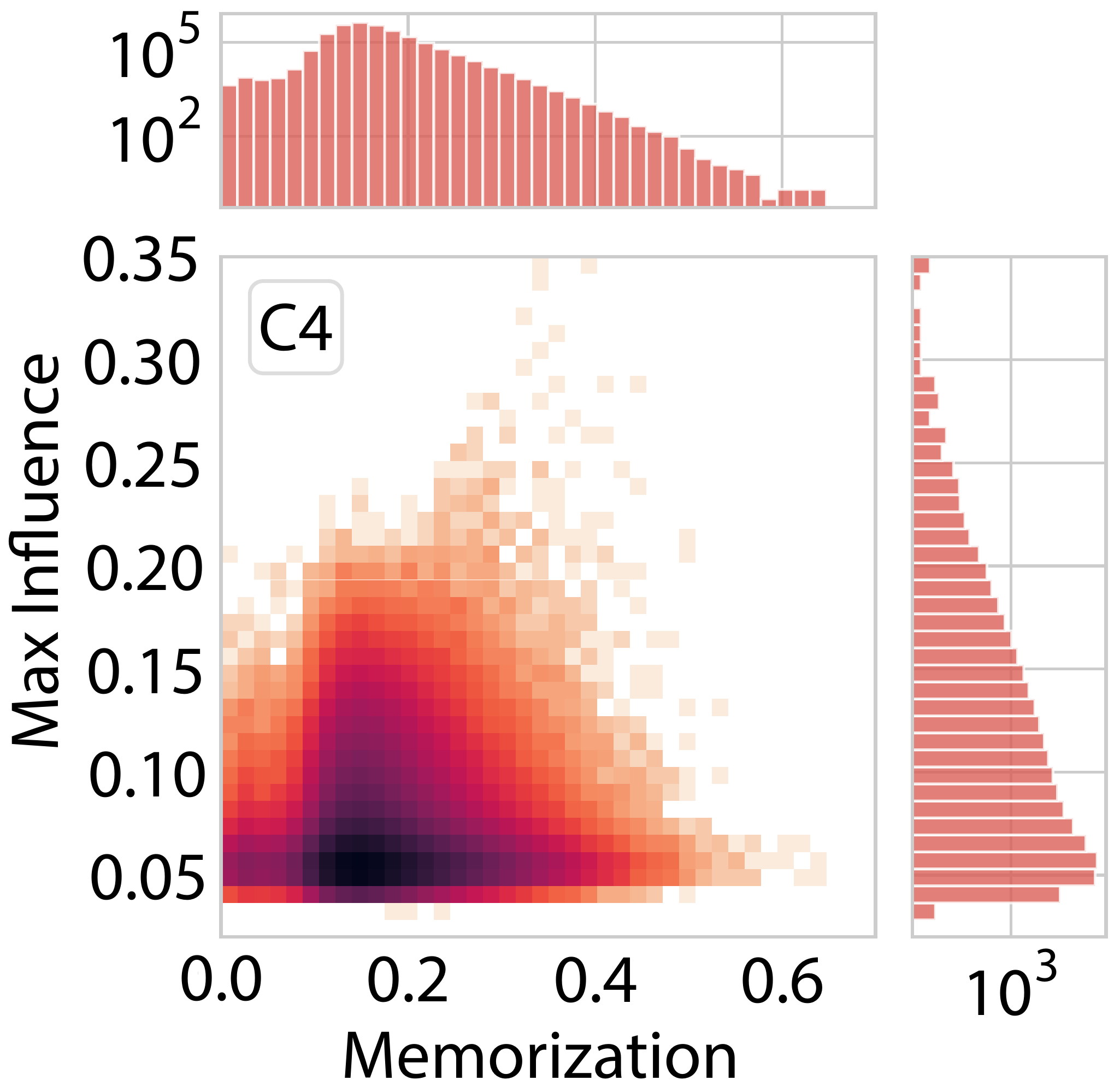}
    \includegraphics[width=.32\linewidth]{./figs/largetext/mem_infl-wiki40b_en.pdf}
    
    \caption{Full version of Figure~\ref{fig:mem-infl}b: The joint distribution of the memorization score of each training example and its maximum influence on any validation set example. The histograms are in log scale to better visualize the tail of the distributions.}
    
    \label{fig:mem-infl-full}
\end{figure}

\section{The Impact of Data Deduplication on Memorization}

\begin{figure}[b]
    \centering
    \begin{overpic}[percent,width=.42\linewidth]{./figs/mem_sim-accuracy-c4.pdf}
            \put(60,20){\fbox{\cfour}}
    \end{overpic}
    \hspace{10pt}
    \begin{overpic}[percent,width=.42\linewidth]{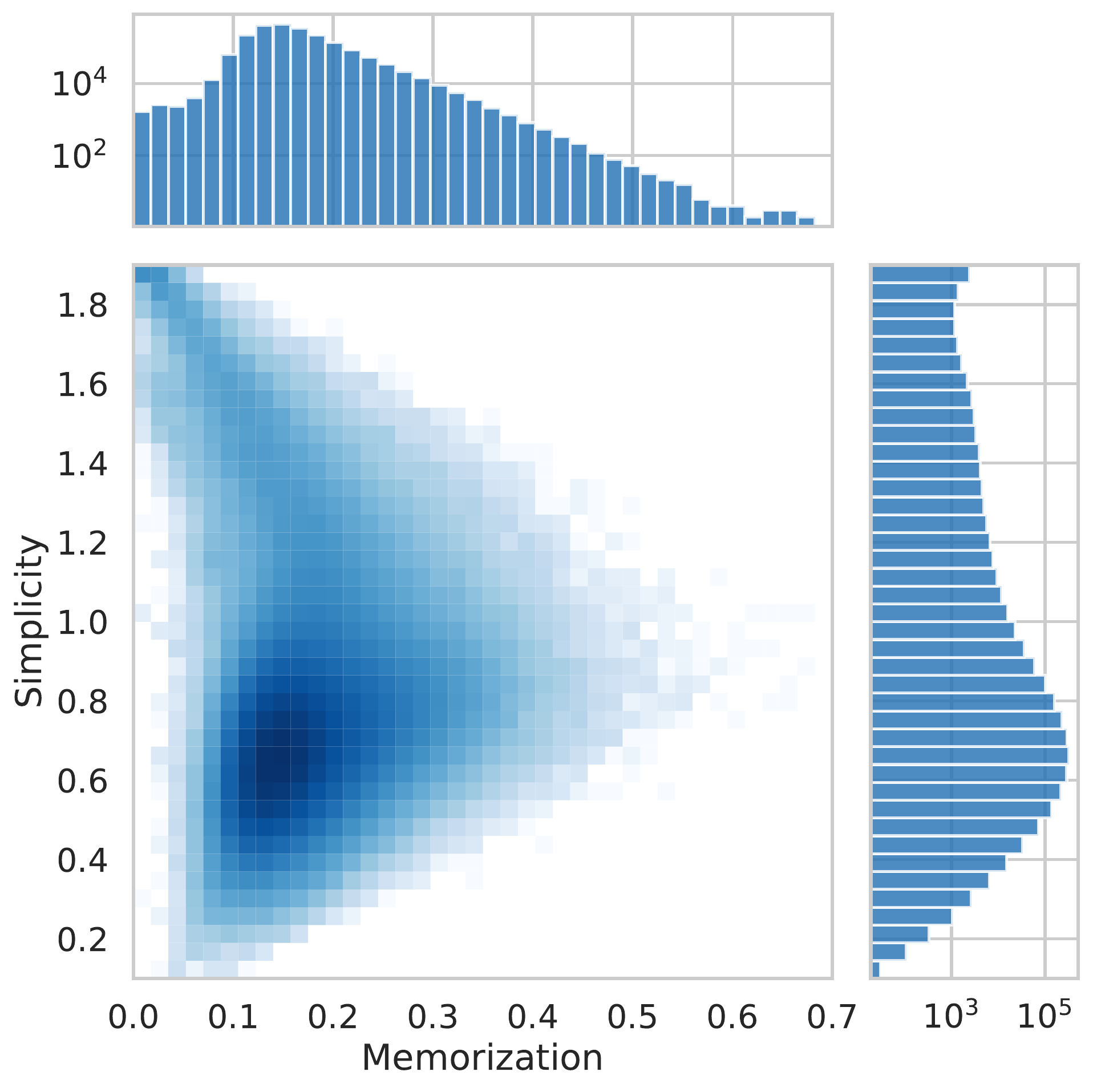}
            \put(35,20){\fbox{\cfour-\textsc{NearDup}}}
    \end{overpic}
    \caption{The joint distribution of memorization and simplicity. The histograms are plotted in log scale to better visualize the tail of the distributions.}
    \label{fig:mem-sim-c4-nd3}
\end{figure}

To investigate the impact of data deduplication on counterfactual memorization, 
we compared \cfour with \cfour-\textsc{NearDup}~\citep{2021dedup}, which is derived from \cfour with deduplication using approximate document matching. Figure~\ref{fig:mem-sim-c4-nd3} compares the distribution of memorization between the original \cfour and the deuplicated dataset. We did not find significant difference between the two datasets. One potential reason is that the deduplication criterion was relatively conservative, which removed only $\sim 3\%$ of the training examples. In fact, we can still easily see near duplicate examples in \cfour-\textsc{NearDup} among examples with low memorization, as shown below:

\begin{description}
    \item[Example 1380925 ($\mem=0.0374$)] \eglink{http://brecksvillebroadviewheightshighschool.com/alumni/3041814/joshua-baldridge.html} This is a placeholder page for Joshua Baldridge, which means this person is not currently on this site. We do suggest using the tools below to find Joshua Baldridge. You are visiting the placeholder page for Joshua Baldridge. This page is here because someone used our placeholder utility to look for Joshua Baldridge. We created this page automatically in hopes Joshua Baldridge would find it. If you are not Joshua Baldridge, but are an alumni of Brecksville Broadview Heights High School, register on this site for free now.
    \item[Example 2048352 ($\mem=0.0320$)] \eglink{http://mainlandhighschool.net/alumni/1073803/laytoya-brannon.html} This is a placeholder page for Laytoya Brannon, which means this person is not currently on this site. We do suggest using the tools below to find Laytoya Brannon. You are visiting the placeholder page for Laytoya Brannon. This page is here because someone used our placeholder utility to look for Laytoya Brannon. We created this page automatically in hopes Laytoya Brannon would find it. If you are not Laytoya Brannon, but are an alumni of Mainland High School, register on this site for free now.
    \item[Example 1314053 ($\mem=0.0278$)] \eglink{http://kankakeevalleyhighschool.org/alumni/2032487/devin-mcguire.html} This is a placeholder page for Devin Mcguire, which means this person is not currently on this site. We do suggest using the tools below to find Devin Mcguire. You are visiting the placeholder page for Devin Mcguire. This page is here because someone used our placeholder utility to look for Devin Mcguire. We created this page automatically in hopes Devin Mcguire would find it. If you are not Devin Mcguire, but are an alumni of Kankakee Valley High School, register on this site for free now.
    \item [Example 1085524 ($\mem=0.0209$)] \eglink{http://oldbridgehighschool.net/alumni/6714550/anthony-christie.html} This is a placeholder page for Anthony Christie, which means this person is not currently on this site. We do suggest using the tools below to find Anthony Christie. You are visiting the placeholder page for Anthony Christie. This page is here because someone used our placeholder utility to look for Anthony Christie. We created this page automatically in hopes Anthony Christie would find it. If you are not Anthony Christie, but are an alumni of Old Bridge High School, register on this site for free now.
\end{description}

Measurements of the edit distances show that they are near the boundary of the deduplication threshold chosen in \citet{2021dedup}. On the other hand, the tail of the distribution --- examples with high counterfactual memorization are mostly unaffected by text deduplication.

\section{Variance of Memorization Scores}
\label{app:variance}
\begin{figure}
    \centering
    \includegraphics[width=.4\linewidth]{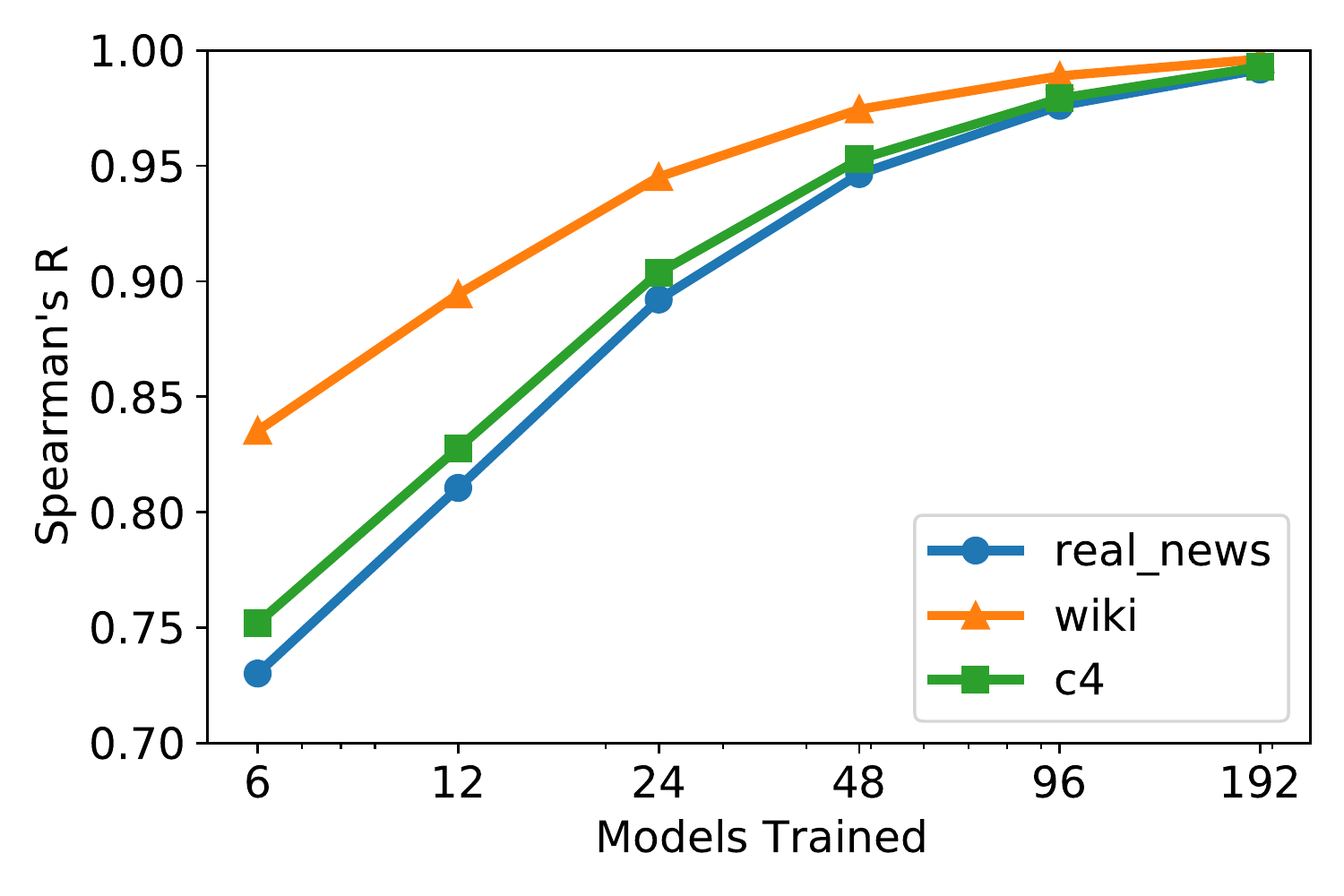}
    \caption{Spearman's R between memorization rankings from a set of $m$ models and our full set of 400 models. As more models are trained, the ranking changes very little, with the ranking at 192 models having a Spearman's R of at least 0.992 on all datasets.}
    \label{fig:spearman_400_variance}
\end{figure}

In Figure~\ref{fig:spearman_400_variance}, we measure the Spearman's R between our total set of 400 models and an $m$ model subset. As expected, as $m$ increases, so does Spearman's R---in particular, at 192 models, the Spearman's R is at least 99.2\% for all datasets, and increasing $m$ already appears to have diminishing returns.

\begin{figure}
    \centering
    \begin{overpic}[width=.32\linewidth]{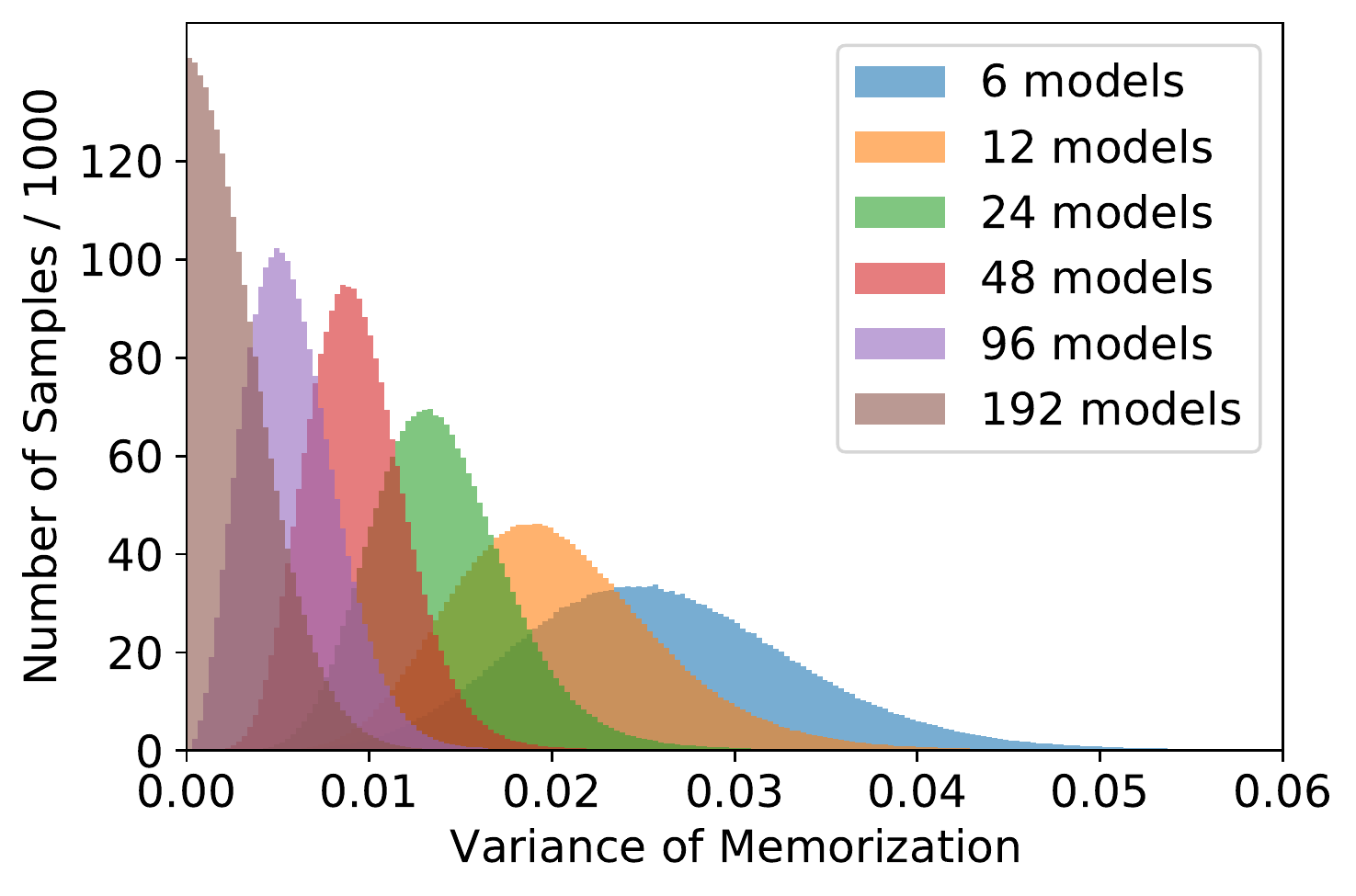}
    \put(35,-5){\scriptsize(a) \realnews}
    \end{overpic}
    \begin{overpic}[width=.32\linewidth]{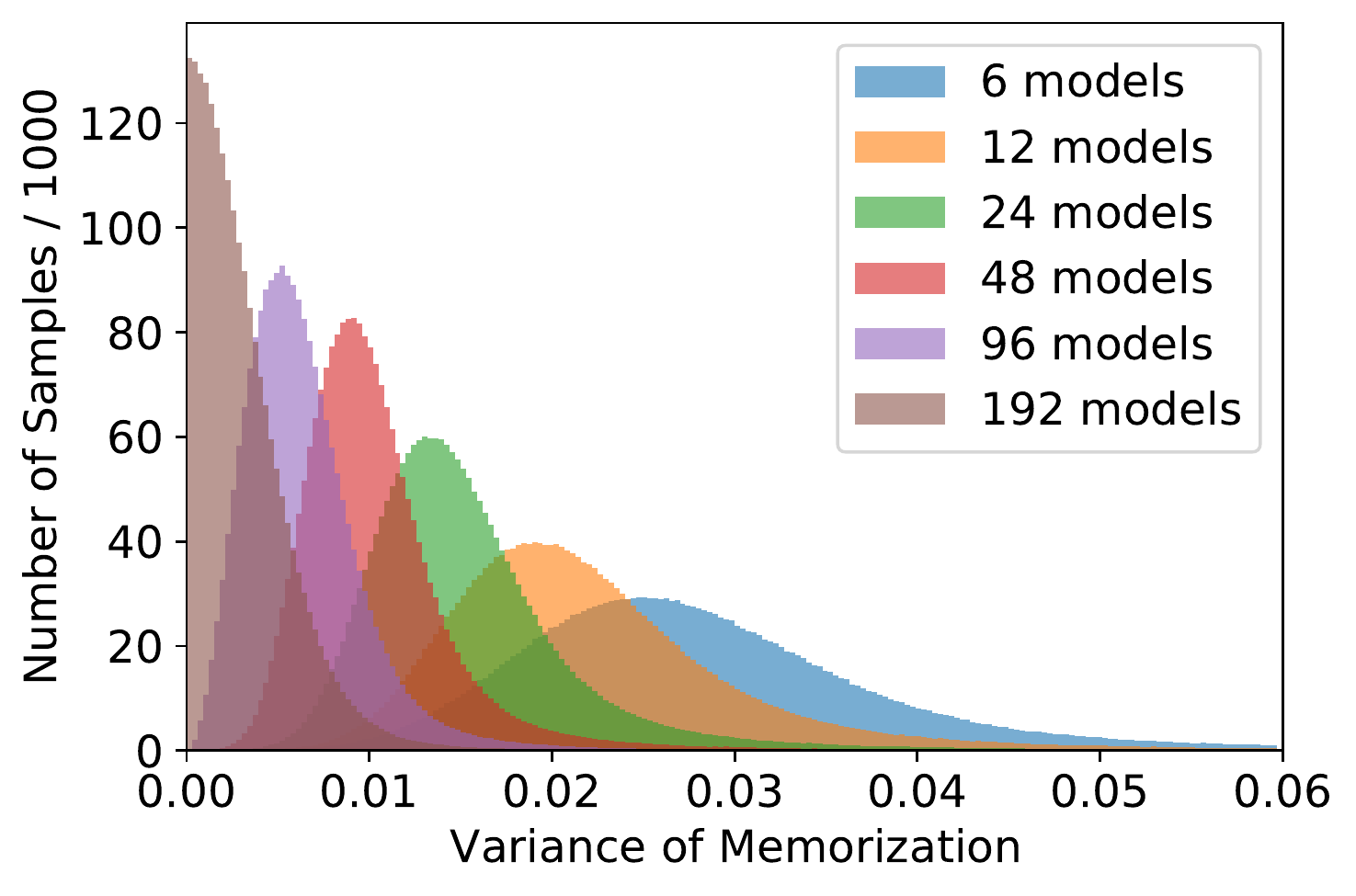}
    \put(45,-5){\scriptsize(b) \cfour}
    \end{overpic}
    \begin{overpic}[width=.32\linewidth]{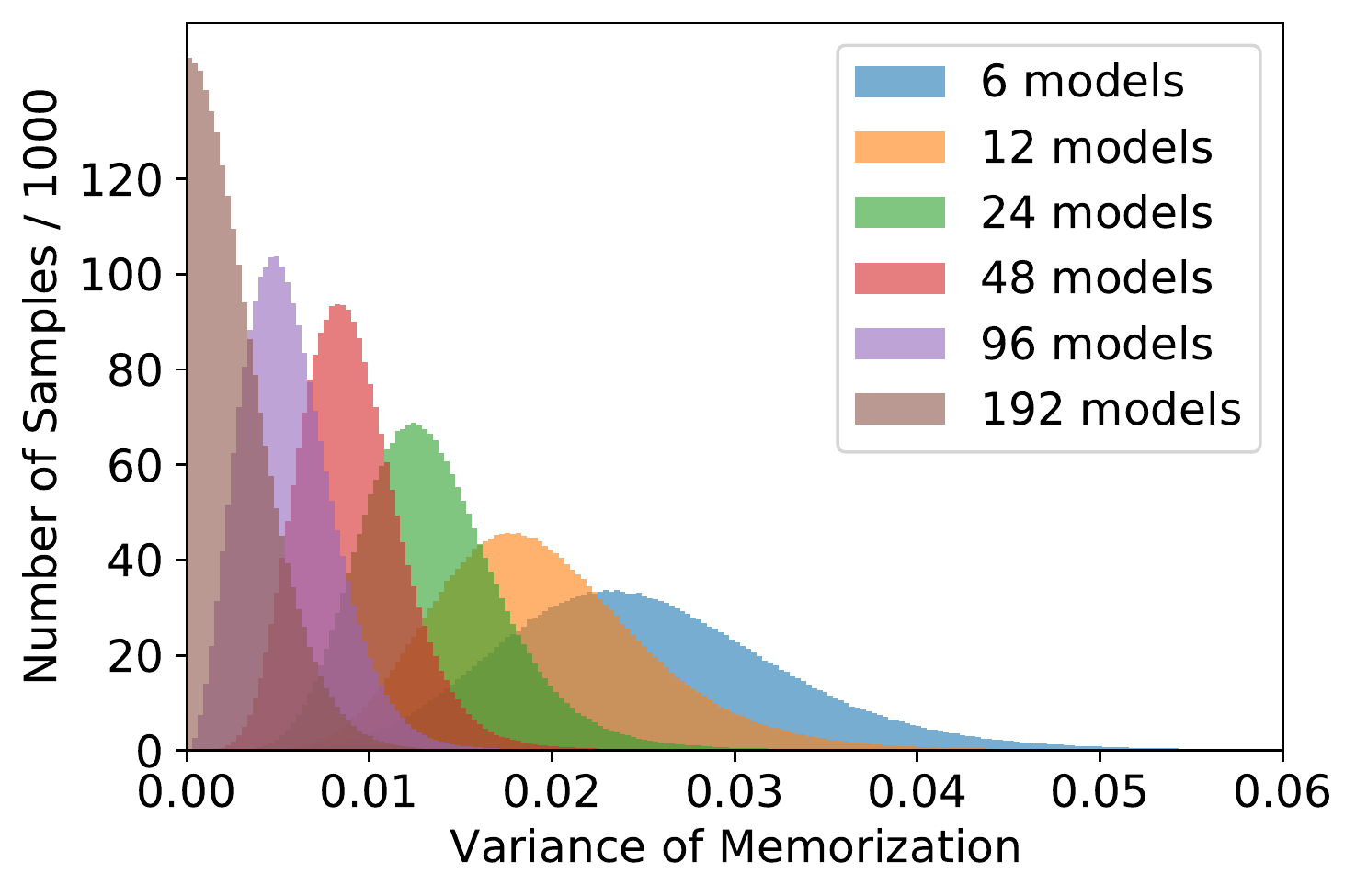}
    \put(35,-5){\scriptsize(c) \wikien}
    \end{overpic}
    \caption{The variance in memorization scores decreases significantly as the number of models increases for all 3 datasets.}
    \label{fig:variance}
\end{figure}

Using the same partitioning into size $m$ sets of models, we analyze the variance of memorization scores assigned to each sample. To do this, within each partition, we compute the memorization score assigned to each sample. We then compute the standard deviation of all partitions' memorization scores \emph{for each sample}. In Figure~\ref{fig:variance}, we plot each sample's standard deviation --- in all, this demonstrates the distribution of the variance of memorization scores. We find that the variance decreases substantially as $m$ grows, and concentrates near 0 already with $m=192$, for all datasets.

\section{Histogram of Max-Influence on Generated Texts}

\begin{figure}
    \centering
    \includegraphics[width=0.4\linewidth]{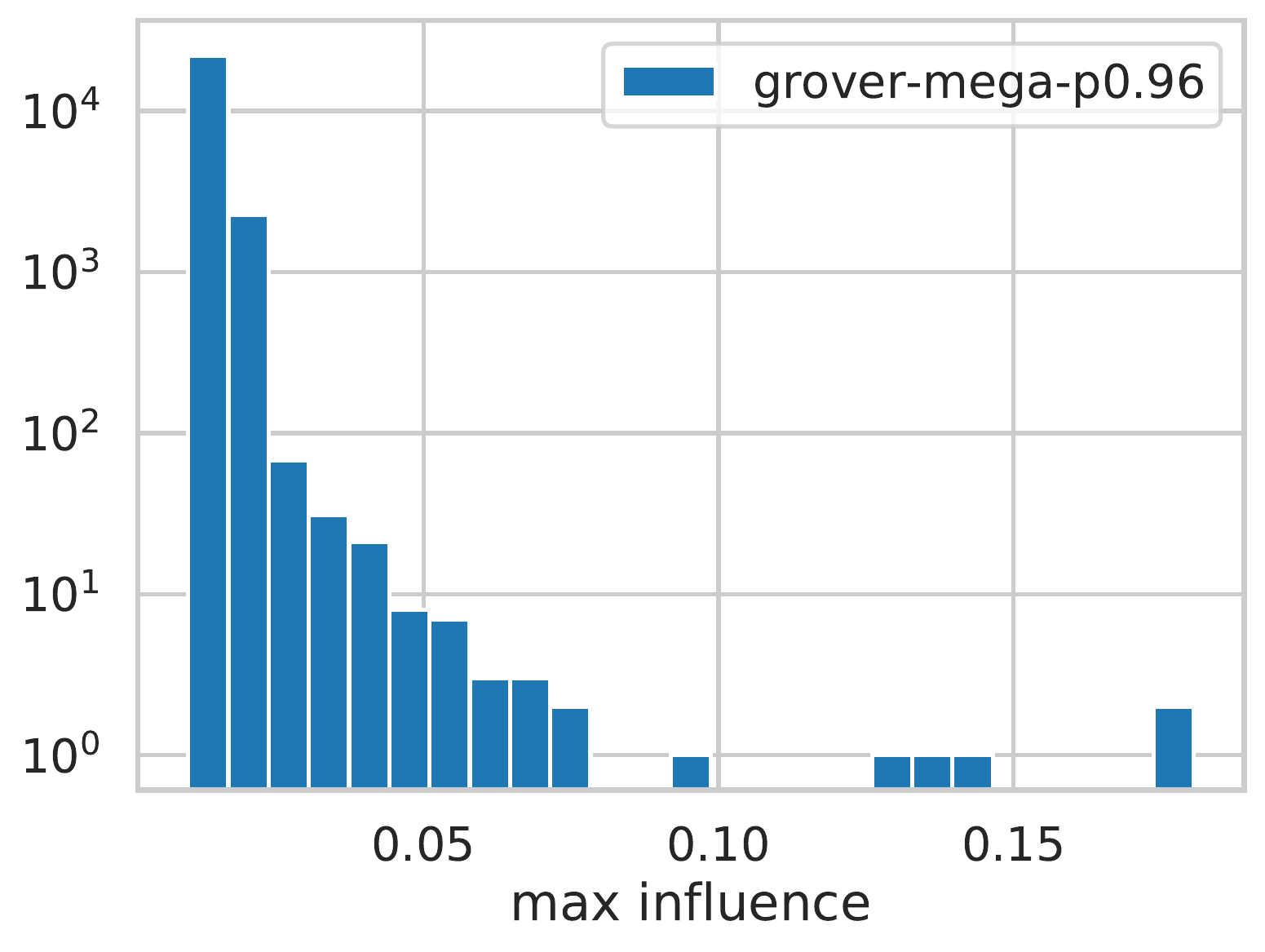}
    \caption{Histogram of max-influence on each generated example by Grover-Mega (p=0.96), from the \realnews training examples.}
    \label{fig:grover-infl}
\end{figure}

Figure~\ref{fig:grover-infl} shows the histogram of max-influence on each generated example by Grover-Mega (p=0.96)~\citep{zellers2019defending}, from the \realnews training examples.
Those generated examples are publicly released at \url{https://github.com/rowanz/grover/tree/master/generation_examples}.

\section{Miscellaneous Experiment Details}
\label{sec:misc-exp-details}

Our experiments are implemented using JAX~\citep{jax2018github} and Flax~\citep{flax2020github}, both open sourced library under the Apache-2.0 license. In the study of influence on generated texts, we use the publicly released generations from the Grover models~\citep{zellers2019defending}, available at \href{https://github.com/rowanz/grover/tree/master/generation_examples}{their open source code repository}, under the  Apache-2.0 license.

We run the experiments using our internal cluster. The majority of the compute is consumed by model training. In this paper, we use standard training setup for transformer based neural language models, which could run on single node machines with one or multiple GPUs. However, to carry out the full analysis, we need to train 400 different models for each of the three datasets analyzed in this paper.

\section{Subsampling Procedure}
\label{sec:subsampling}

\begin{figure}
    \centering
    \includegraphics[width=.95\linewidth]{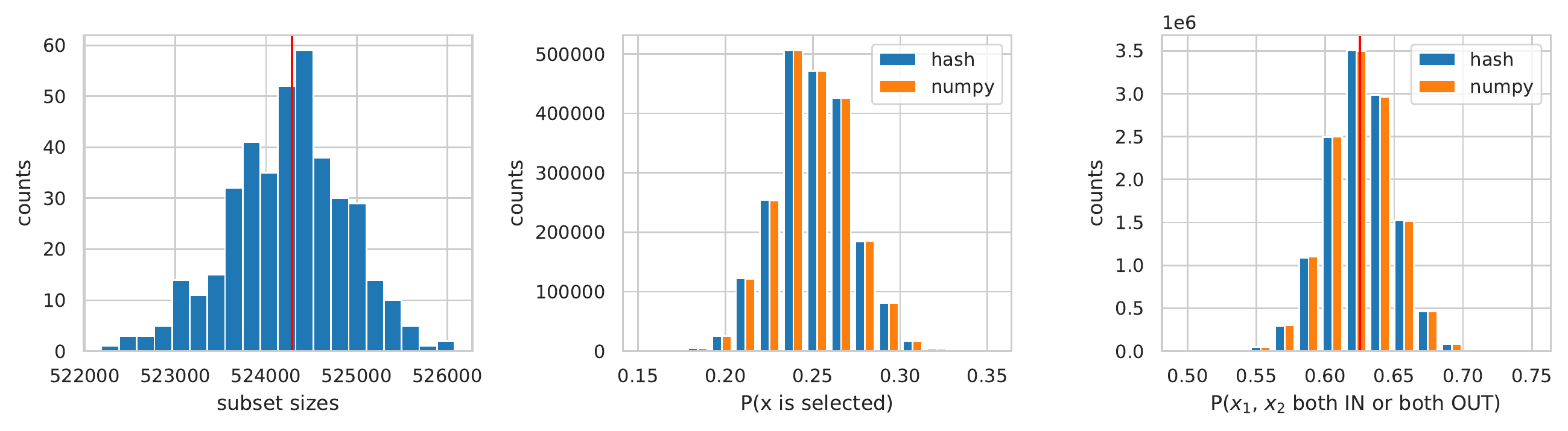}
    \caption{Comparison of hash based subset sampling with \texttt{numpy.random.choice}.}
    \label{fig:hash-sampler}
\end{figure}

In the estimation of memorization and influence, we trained 400 models each on an independent random subset of training examples. We use \emph{Tensorflow Datasets} (TFDS) \footnote{\url{https://www.tensorflow.org/datasets}} to load our training data. TFDS supports loading a \emph{continuous} range of examples, but does not support subset loading from a list of indices of individual examples. The API has a \texttt{filter} function which allows us to provide a Tensorflow predicate to precisely control the subset loading. However, a naive implementation of checking whether the index of the current example is in a given list of subset indices is very slow and scales poorly with the subset size.

To mitigate the issue, we implemented a hash based subset sampling predicate that can be evaluated efficiently for each example, and (approximately) select a random subset of a specified size. Let $N$ be the total number of training examples, $n<N$ be the expected subset size. The idea is to map the index $i$ of each example to $N/n$ hash buckets, and select all the examples that fall into one particular bucket. To make sure each model gets an independent subset sampling, we need to use different hash functions for different models. In our implementation, we compose a known hash function for \texttt{uint64} types with a simple pseudo number based on the index of the current model to achieve this. Note the subset size sampled is close to $n$ but is not guaranteed to be exactly $n$. But this is not a problem in our settings. The specific implementation is shown below:

\begin{verbatim}
def hash_sampler(mod, seed, system):
  """Get hash based subset sampler.

  Args:
    mod: total_n_egs // subset_size
    seed: different seed leads to different subset sample
    system: 'np' or 'tf'.

  Returns:
    A Tensorflow or Numpy subset sampler.
  """
  np_hash = hash_uint64_builder('np')
  mul, offset, remainder = np_hash(seed + 1234 + np.arange(3))
  remainder = remainder %

  if system == 'np':
    def np_sampler(n_total):
      x = np.arange(n_total, dtype=np.uint64)
      return np_hash(x*mul + offset) %
    return np_sampler
  elif system == 'tf':
    tf_hash = hash_uint64_builder('tf')
    def tf_filter(idx, _):
      return tf.equal(tf_hash(idx*mul + offset) %
    return tf_filter
  raise KeyError(f'Unknown system: {system}')


def hash_uint64_builder(system):
  """Build a hash function in tf/np for uint64."""
  if system == 'np':
    uint64_cast = functools.partial(np.array, dtype=np.uint64)
    op_xor = operator.xor
    op_rshift = operator.rshift
  elif system == 'tf':
    uint64_cast = functools.partial(tf.cast, dtype=tf.uint64)
    op_xor = tf.bitwise.bitwise_xor
    op_rshift = tf.bitwise.right_shift
  else:
    raise KeyError(f'Unknown system: {system}')

  # https://stackoverflow.com/questions/664014/
  # what-integer-hash-function-are-good-that-accepts-an-integer-hash-key
  def hash_uint64(x):
    x = uint64_cast(x)
    x = op_xor(x, op_rshift(x, 30)) * uint64_cast(0xbf58476d1ce4e5b9)
    x = op_xor(x, op_rshift(x, 27)) * uint64_cast(0x94d049bb133111eb)
    x = op_xor(x, op_rshift(x, 31))
    return x

  return hash_uint64
\end{verbatim}

In Figure~\ref{fig:hash-sampler}, we compare our hash-based subset sampler with \texttt{numpy.random.choice(N, size=n, replace=False)}. The leftmost section of the figure shows that the sampling procedure always samples close to $n$ points, with a small variance. The middle section plots a histogram of the empirical fraction of total models that each point appears in. Note that, because we use $r=0.25$, this fraction should be 0.25 on average, although, because we only use 400 models, each value will not be identically 0.25. We find that our hash-based sampler produces probabilities which are highly consistent with those produced by \texttt{numpy.random.choice}. We also measure the pairwise independence of the hash-based sampler, measuring the probability that two different training points $x_1, x_2$ appear both IN or OUT of a model's training set. We expect this value to be 0.625 (=$r^2 + (1-r)^2$). We plot this in the right portion of the figure, demonstrating that the independence of our hash-based sampler is very similar to \texttt{numpy.random.choice}.

\section{Alternative Memorization Metrics with Logit Scaling}

\begin{figure}
    \centering
    \begin{overpic}[percent,width=.32\linewidth]{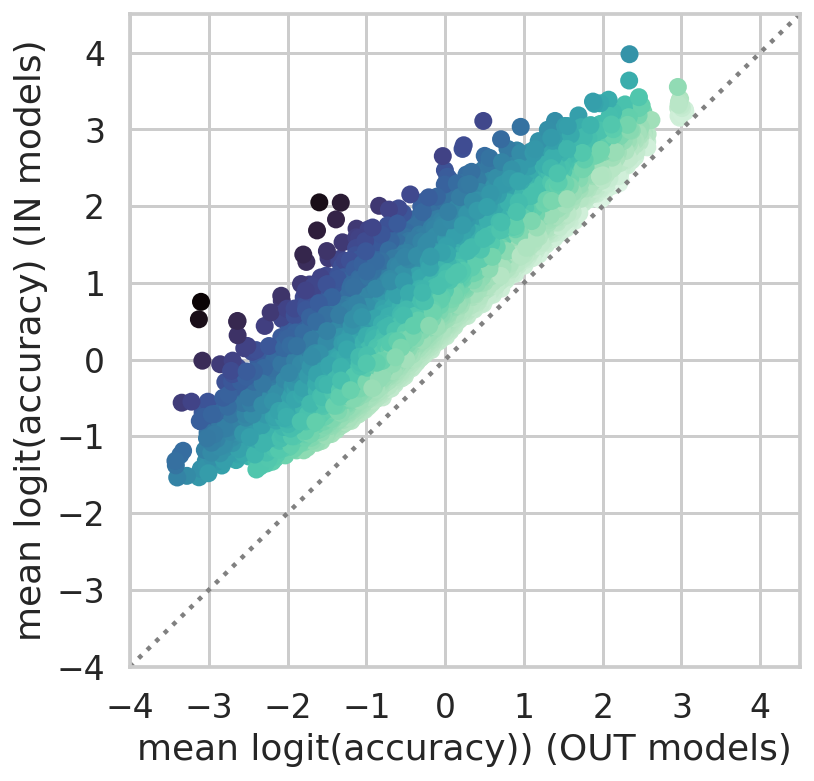}
            \put(45,20){\fbox{\realnews}}
    \end{overpic}
    \begin{overpic}[percent,width=.32\linewidth]{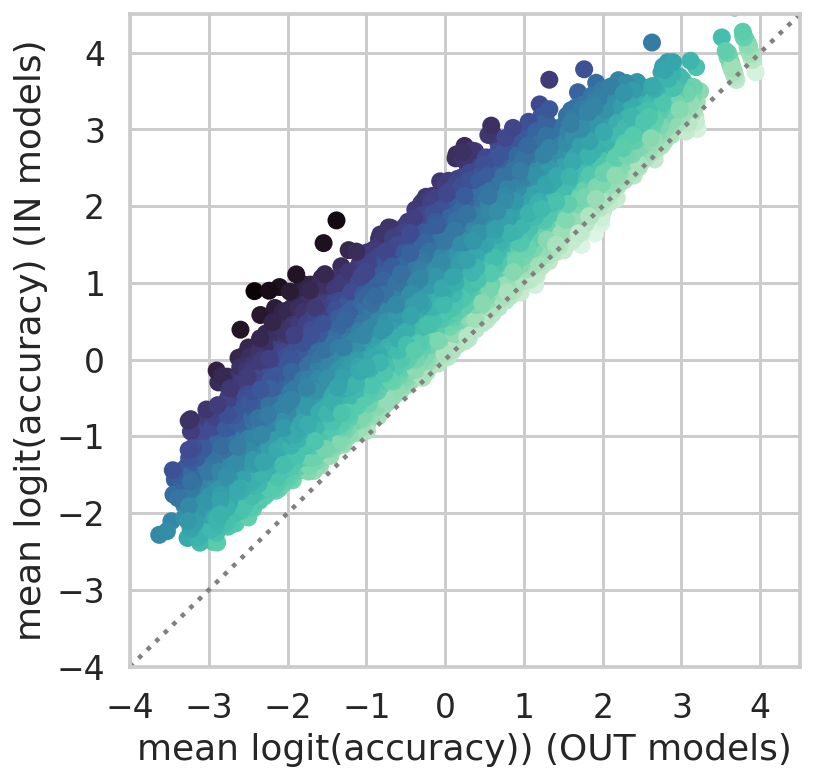}
            \put(69,20){\fbox{\cfour}}
    \end{overpic}
    \begin{overpic}[percent,width=.32\linewidth]{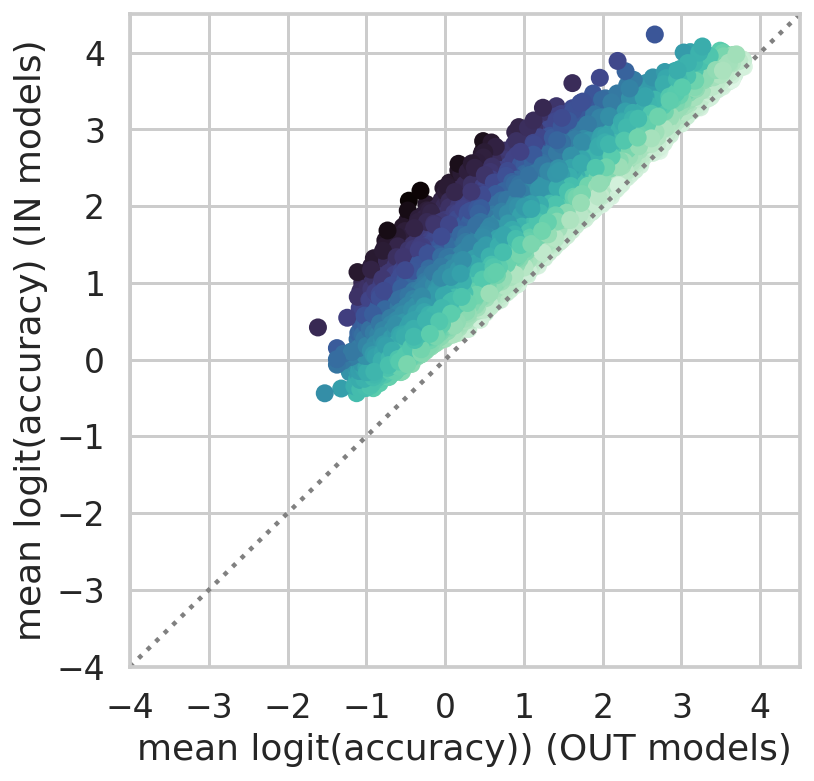}
            \put(40,20){\fbox{\wikien}}
    \end{overpic}
    \begin{overpic}[percent,width=.32\linewidth]{./figs/scatter-accuracy-real_news.png}
            \put(45,20){\fbox{\realnews}}
    \end{overpic}
    \begin{overpic}[percent,width=.32\linewidth]{./figs/scatter-accuracy-c4.png}
            \put(69,20){\fbox{\cfour}}
    \end{overpic}
    \begin{overpic}[percent,width=.32\linewidth]{./figs/scatter-accuracy-wiki40b_en.png}
            \put(40,20){\fbox{\wikien}}
    \end{overpic}
    
    \caption{Comparison of directly using the per-token-accuracy vs. taking the logit of the per-token-accuracy. \textit{Top row}: logit(per-token-accuracy). \textit{Bottom row}: per-token-accuracy. Figures exactly the same as Figure~\ref{fig:scatter}.}
    \label{fig:scatter-logit}
\end{figure}

We defined the counterfactual memorization in \eqref{eq:mem} with a generic performance measure $M$. Throughout the paper, we define $M$ as per-token accuracy--the fraction of the times the model assigns the highest score to the true next token in the sequence. The finite value range could cause unnecessary compression for values near the interval boundary. As a result, the resolution of memorization estimation is lower for models with very high or very low performance. To mitigate this issue, we explore an alternative measure by taking the logit on the per-token accuracy~\citep{carlini2021membership}. The logit function maps to $(-\infty,\infty)$ before aggregating across independently trained models. Figure \ref{fig:scatter-logit} compares the scatter plots of average performance on IN / OUT models measured by the logit scaled per-token accuracy and the raw per-token accuracy. Comparing to the raw per-token accuracy, the scatter plots generated with the logit scaled measure are no longer artificially constrained to be a triangular shape. As a result, the memorization estimation, which is proportional to the distance to the diagonal line, has a higher resolution on the two ends (lower left and upper right) than the unscaled version. 

Note there is no absolutely right or wrong measure. While the scaled version has better resolution on the two ends, the advantage of the unscaled version is that the value range $[0,1]$ makes it straightforward to interpret the numerical values of counterfactual memorization. Since the consistency between the two versions are high (Spearman's $\rho$ correlation between the two versions are 0.947 / 0.903 / 0.944  on \realnews / \cfour / \wikien), we use the unscaled version throughout the paper for easier interpretation. 

\section{Definition of Edit Similarity}\label{sec:edit-sim}

We define the edit similarity between two sequences $x_i$ and $x_j$ as. In our case, we use token-level similarity. 

\begin{equation*}
    \operatorname{EditSim}(x_i, x_j) = 1 - \frac{\operatorname{EditDistance}(x_i, x_j)}{\max(|x_i|, |x_j|)}
\end{equation*}
\section{Examples of Train-Generation Pairs at Different Influence Ranges}

In table~\ref{tab:infl-egs-grover}, we show examples of train-generation pairs sampled from different influence ranges. 
The patterns generally follow the train-validation pairs shown in table~\ref{tab:infl-egs-realnews}, although many of the relations are due to some form of templating. 

\begin{table}[]
    \caption{\small Pairs of \realnews training examples and Grover generations sampled at several influence levels. ``link'' contains the document URL.
    \egomit indicate text omitted for brevity.
    Differences in each pair are \hlc{highlighted}.}
    \label{tab:infl-egs-grover}
    \centering\tiny
    \begin{tabularx}{\linewidth}{@{}ccX@{}}
    \toprule
    Index & Estim. & Text \\\midrule\midrule
    \makecell[tc]{Generation\\1361} & \makecell[tc]{$\infl$\\\textbf{0.1805}} & \eglink{https://en.trend.az/business/finance/3046505.html} Baku, Azerbaijan, \hlc{April} 15 Trend: Official exchange rate of the US dollar and euro against Azerbaijani manat was set at 1.7 and 1.92\hlc{25} manats, respectively, for \hlc{April} 15. Below are the rates of Azerbaijani manat against world currencies, according to the data from the Central Bank of Azerbaijan for \hlc{April} 15. \egomit 100 Japanese yen 100 JPY 1.5\hlc{187} 1 New Zealand dollar 1 NZD 1.1\hlc{513} Follow Trend on Telegram. Only most interesting and important news
    \\\rowcolor{TableHL}
    \makecell[tc]{Train\\2072973} & \makecell[tc]{$\mem$ \\0.3534} & \eglink{https://en.trend.az/business/finance/3033358.html} Baku, Azerbaijan, \hlc{March} 15 Trend: Official exchange rate of the US dollar and euro against Azerbaijani manat was set at 1.7 and 1.92\hlc{41} manats, respectively, for \hlc{March} 15. Below are the rates of Azerbaijani manat against world currencies, according to the data from the Central Bank of Azerbaijan for \hlc{March} 15. \egomit 100 Japanese yen 100 JPY 1.52\hlc{20} 1 New Zealand dollar 1 NZD 1.16\hlc{36} Follow Trend on Telegram. Only most interesting and important news
    \\\midrule
    \makecell[tc]{Generation\\21998} & \makecell[tc]{$\infl$\\\textbf{0.0218}} & \eglink{http://www.arabnews.com/node/1483111/world} NEW DELHI\hlc{:} India is likely to \hlc{see} average monsoon rains \hlc{this year}, the \hlc{state-run }weather office said\hlc{ on Monday, which should support agricultural production} and economic growth in Asia's third-biggest economy, where half of the farmland lacks irrigation. Monsoon rain\hlc{fall is }expected to be \hlc{96} percent of the long-term average, \hlc{M. Rajeevan, secretary at the Ministry of Earth Sciences}, told a news conference. \hlc{The India Meteorological Department (IMD)} defines average, or normal, rainfall as between 96 percent and 104 percent of a 50-year average of 89 \hlc{centimeters} for the entire four-month season beginning June. \egomit \hlc{India's weather office will update its forecast in the first week of June. However, on average, the IMD has forecast accurately only once every five years over the past two decades, even after taking into account an error band of plus or minus 5 percentage points}.
    \\\rowcolor{TableHL}
    \makecell[tc]{Train\\326212} & \makecell[tc]{$\mem$ \\0.1555} & \eglink{https://www.reuters.com/article/us-india-monsoon/india-set-to-receive-average-monsoon-rains-in-boost-for-poll-bound-modi-idUSKBN1HN1AQ} NEW DELHI\hlc{ (Reuters) -} India is likely to \hlc{receive} average monsoon rains \hlc{in 2018}, the weather office said\hlc{, raising the possibility of higher farm and} economic growth in Asia's third-biggest economy, where half of the farmland lacks irrigation. Monsoon rain\hlc{s, the lifeblood of the country's \$2 trillion economy, are} expected to be \hlc{97} percent of \hlc{a} long-term average, \hlc{K.J. Ramesh, director general of the state-run India Meteorological Department (IMD)}, told a news conference. \hlc{``We see very less probability of a deficit monsoon,'' Ramesh said on Monday. Other than lifting farm and wider economic growth, a spell of good rains will keep a lid on inflation, potentially tempting Prime Minister Narendra Modi to bring forward general elections due in May 2019. India's weather office} defines average, or normal, rainfall as between 96 percent and 104 percent of a 50-year average of 89 \hlc{cms} for the entire four-month season beginning June. \egomit \hlc{said a Mumbai-based dealer with a global trading firm. Average monsoon rainfall will help India retain its position as the world's top rice exporter}.
    \\\bottomrule
    \end{tabularx}
\end{table}
\section{Examples Sampled at Different Level of Memorization}

\begin{figure}
    \centering
    \includegraphics[width=\linewidth]{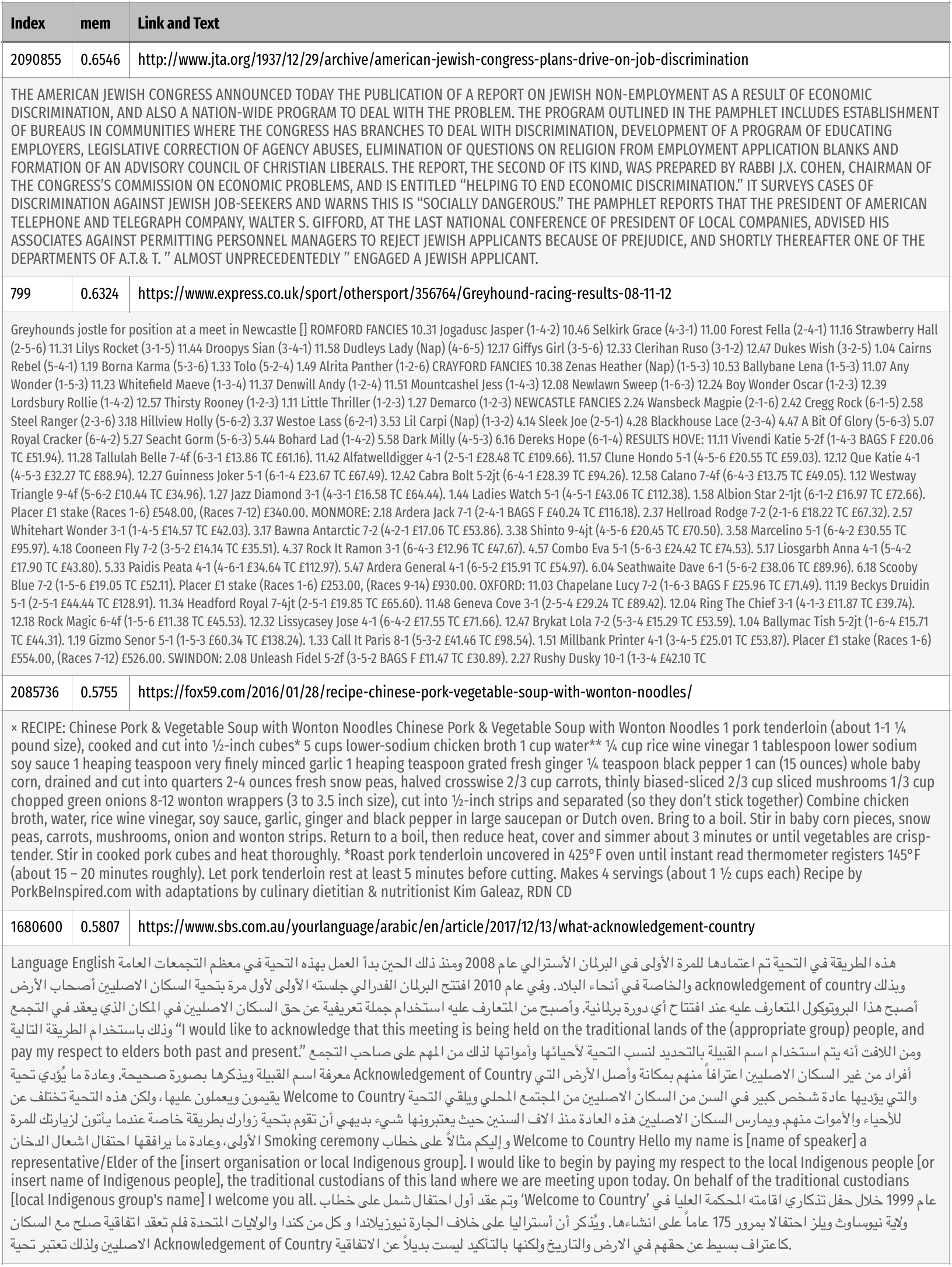}
    \caption{Text examples from \realnews with high memorization.}
    \label{fig:egs-realnews-high-mem}
\end{figure}

\begin{figure}
    \centering
    \includegraphics[width=\linewidth]{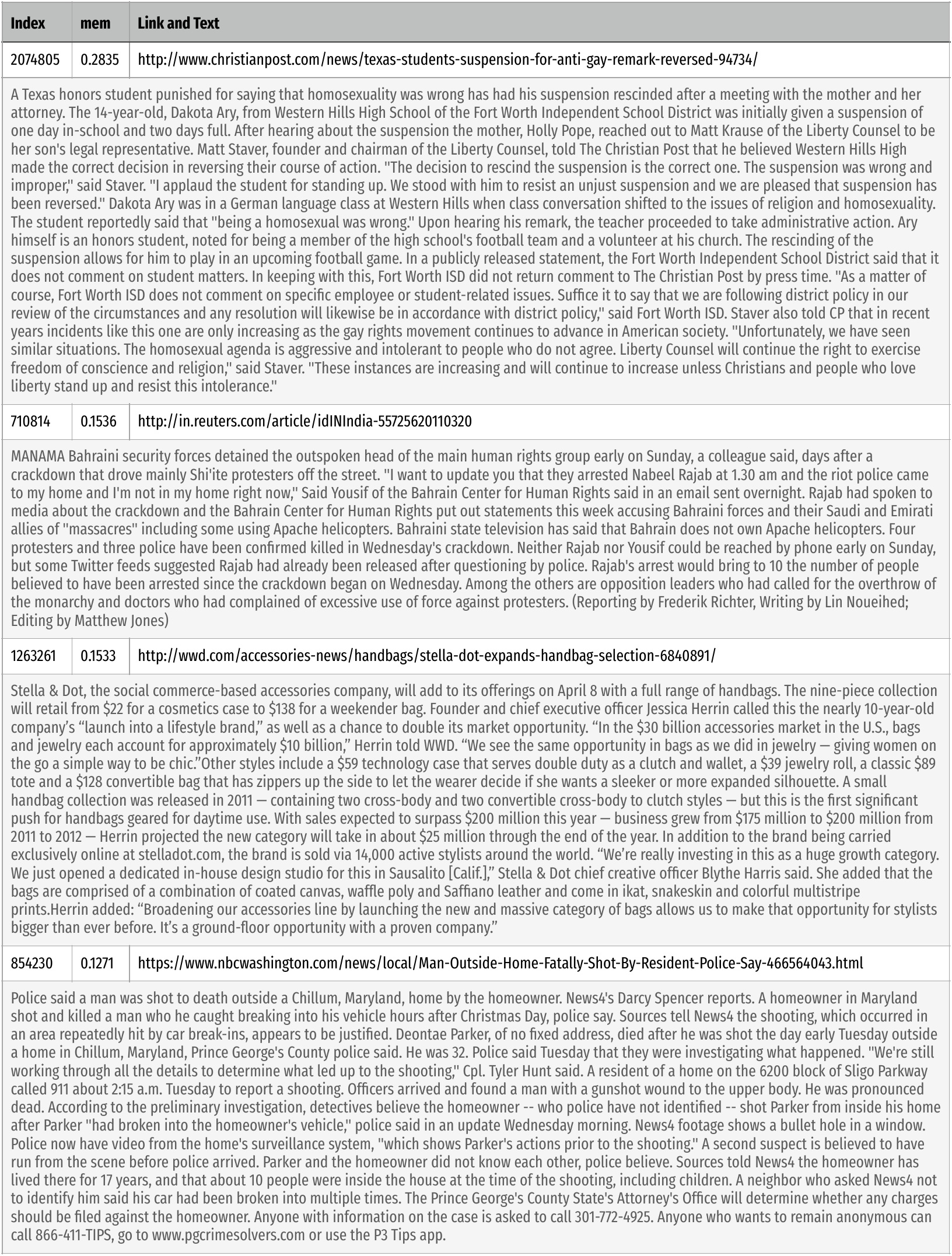}
    \caption{Text examples from \realnews with intermediate memorization.}
    \label{fig:egs-realnews-mid-mem}
\end{figure}

\begin{figure}
    \centering
    \includegraphics[width=\linewidth]{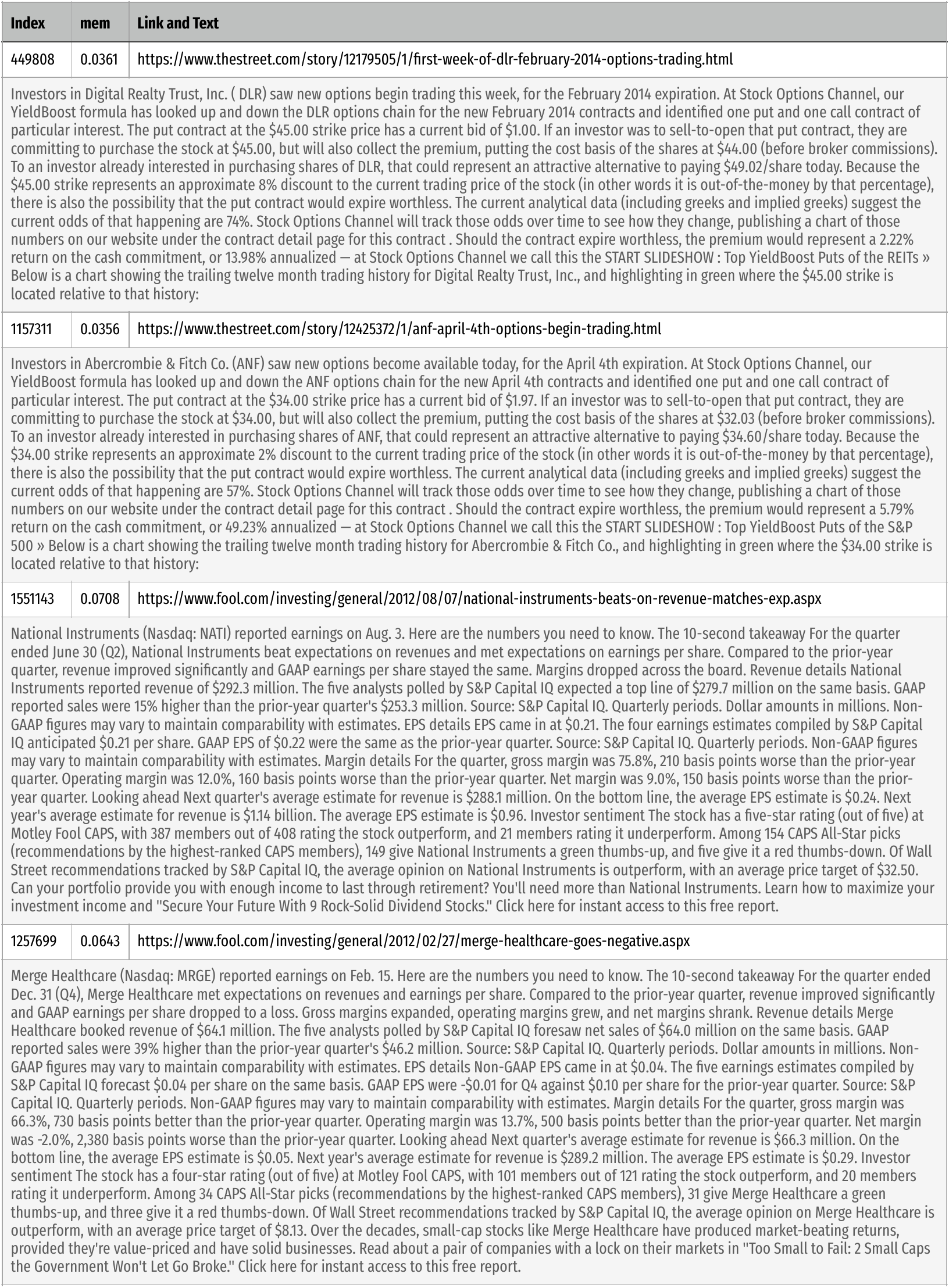}
    \caption{Text examples from \realnews with low memorization.}
    \label{fig:egs-realnews-low-mem}
\end{figure}

\begin{figure}
    \centering
    \includegraphics[width=\linewidth]{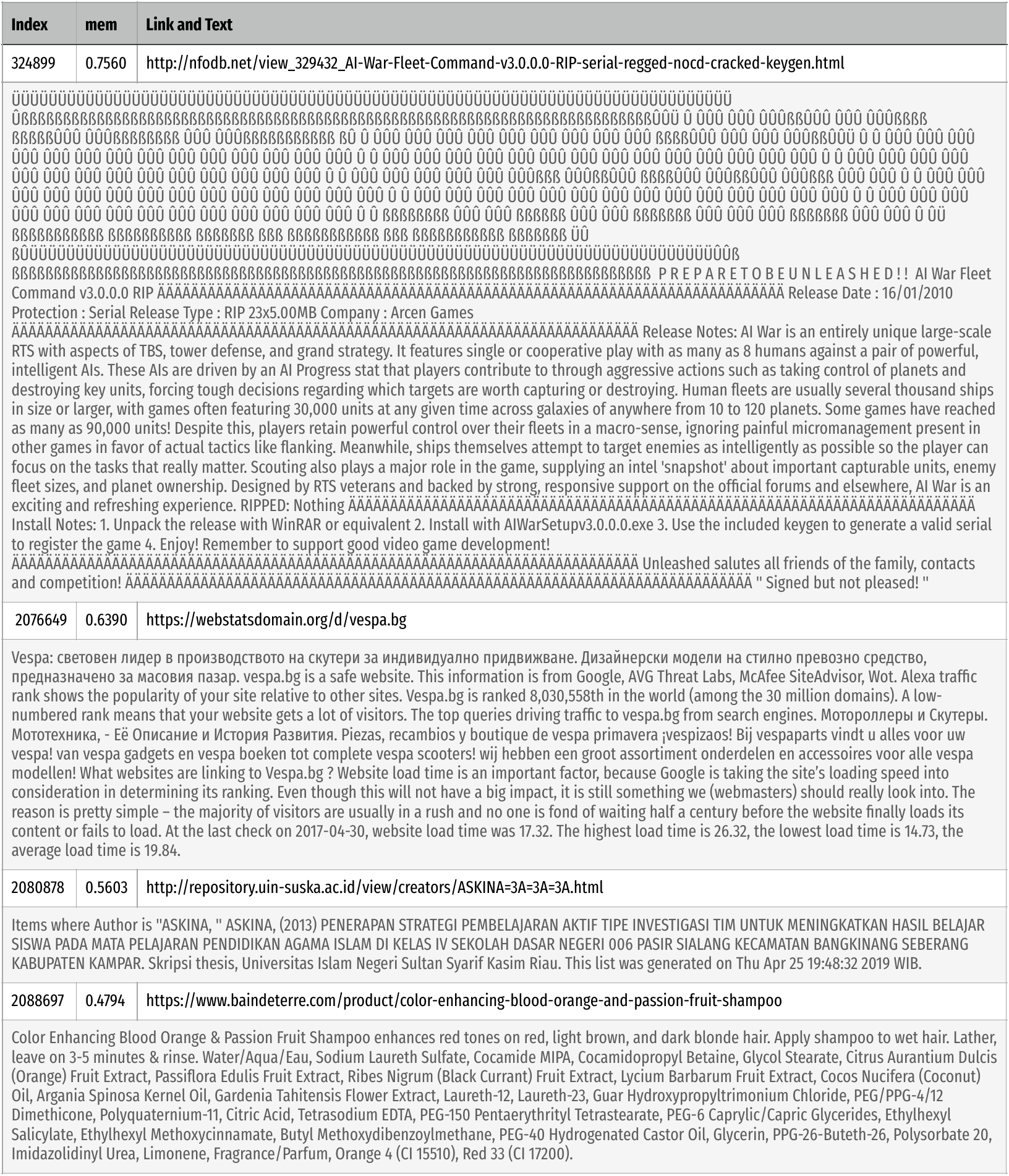}
    \caption{Text examples from \cfour with high memorization.}
    \label{fig:egs-c4-high-mem}
\end{figure}

\begin{figure}
    \centering
    \includegraphics[width=\linewidth]{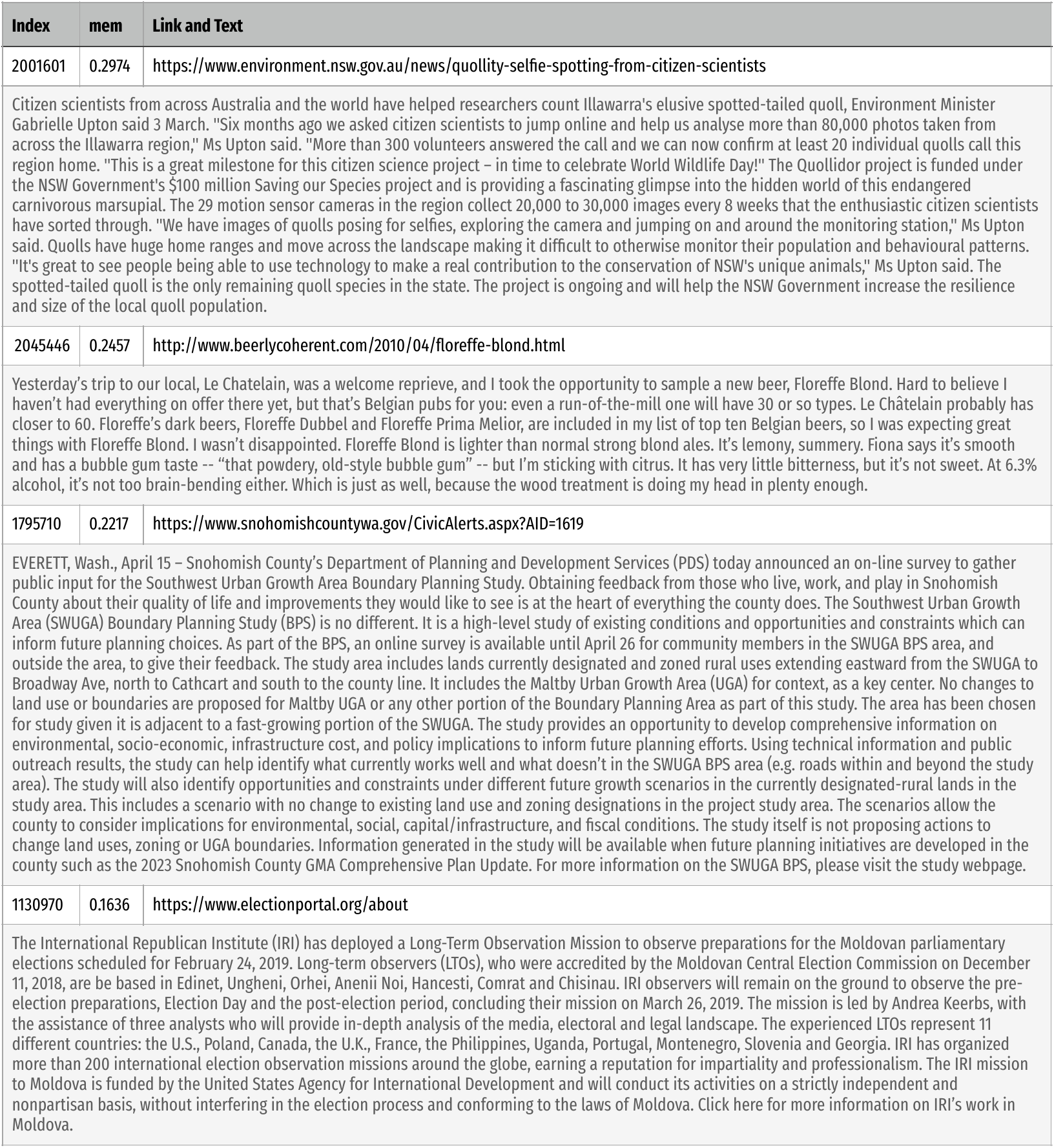}
    \caption{Text examples from \cfour with intermediate memorization.}
    \label{fig:egs-c4-mid-mem}
\end{figure}

\begin{figure}
    \centering
    \includegraphics[width=\linewidth]{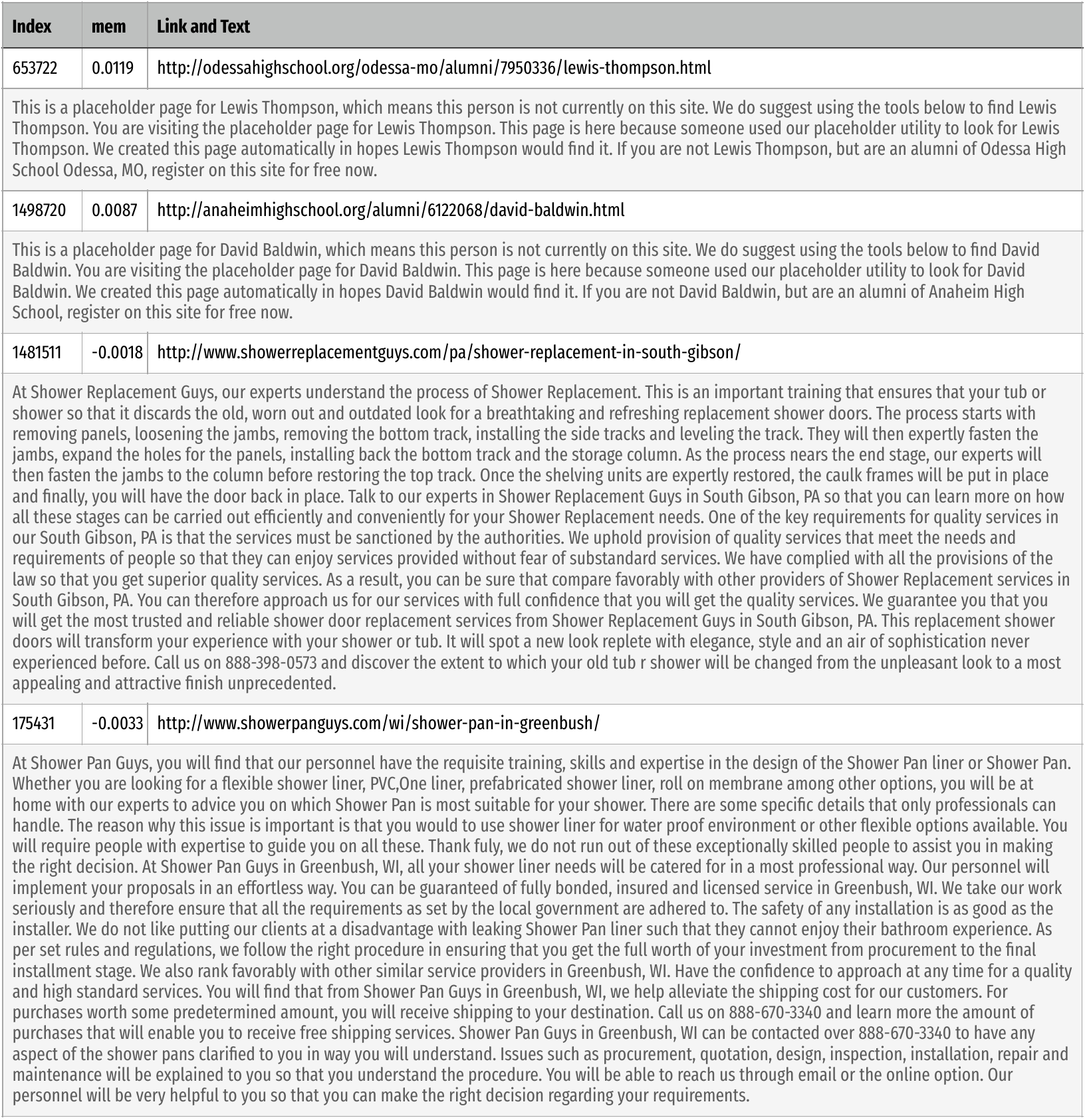}
    \caption{Text examples from \cfour with low memorization.}
    \label{fig:egs-c4-low-mem}
\end{figure}

\begin{figure}
    \centering
    \includegraphics[width=\linewidth]{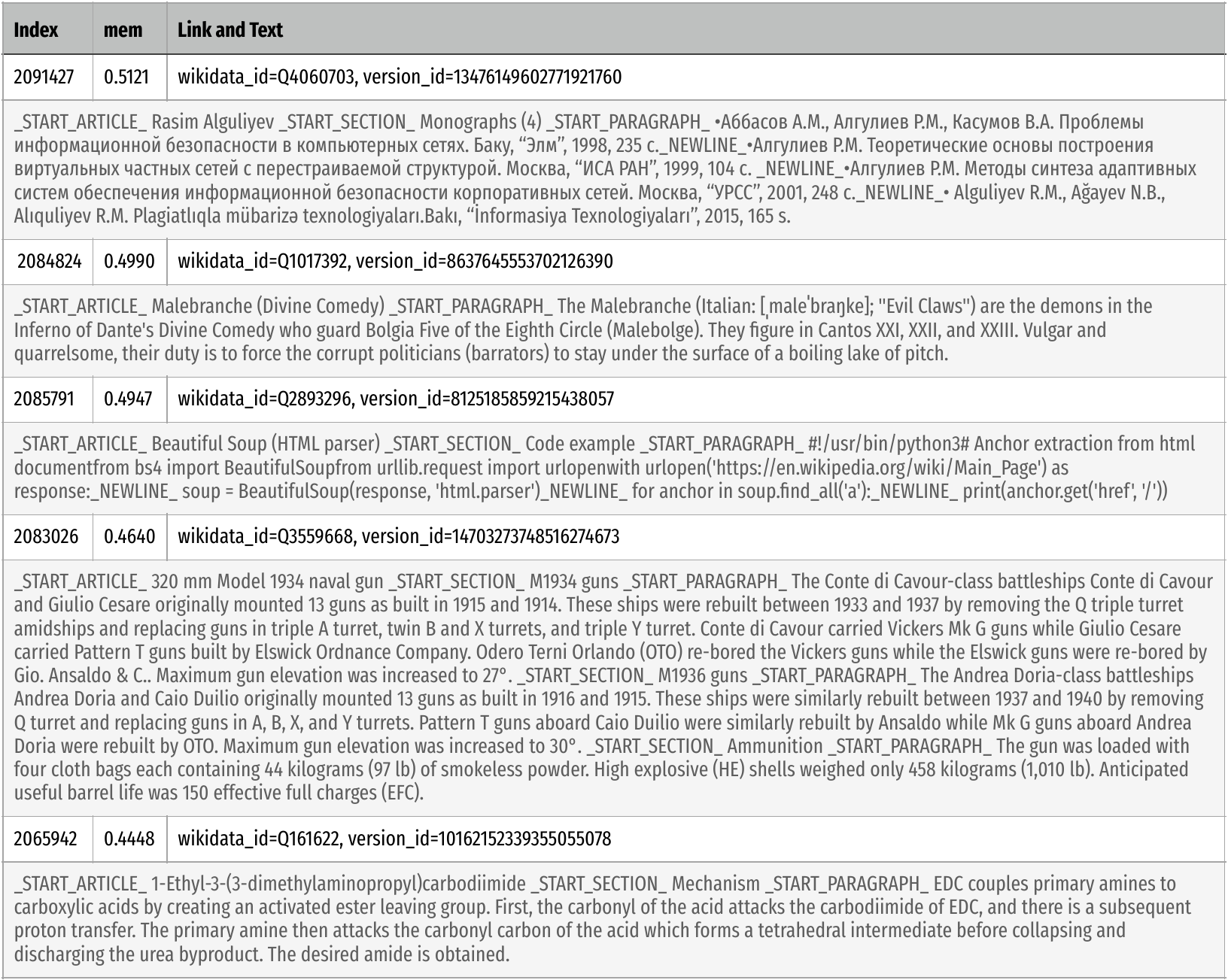}
    \caption{Text examples from \wikien with high memorization.}
    \label{fig:egs-wikitext-high-mem}
\end{figure}

\begin{figure}
    \centering
    \includegraphics[width=\linewidth]{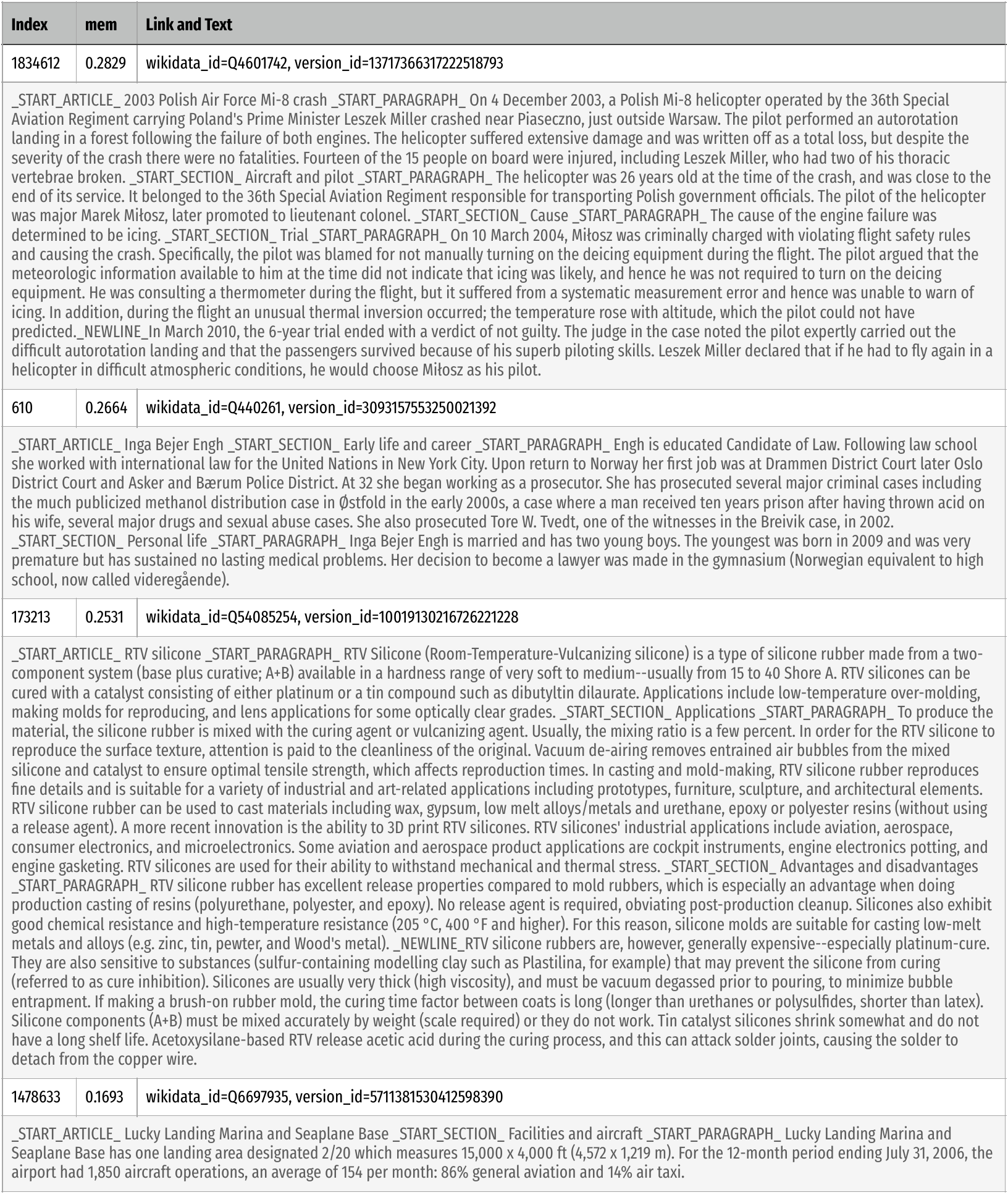}
    \caption{Text examples from \wikien with intermediate memorization.}
    \label{fig:egs-wikitext-mid-mem}
\end{figure}

\begin{figure}
    \centering
    \includegraphics[width=\linewidth]{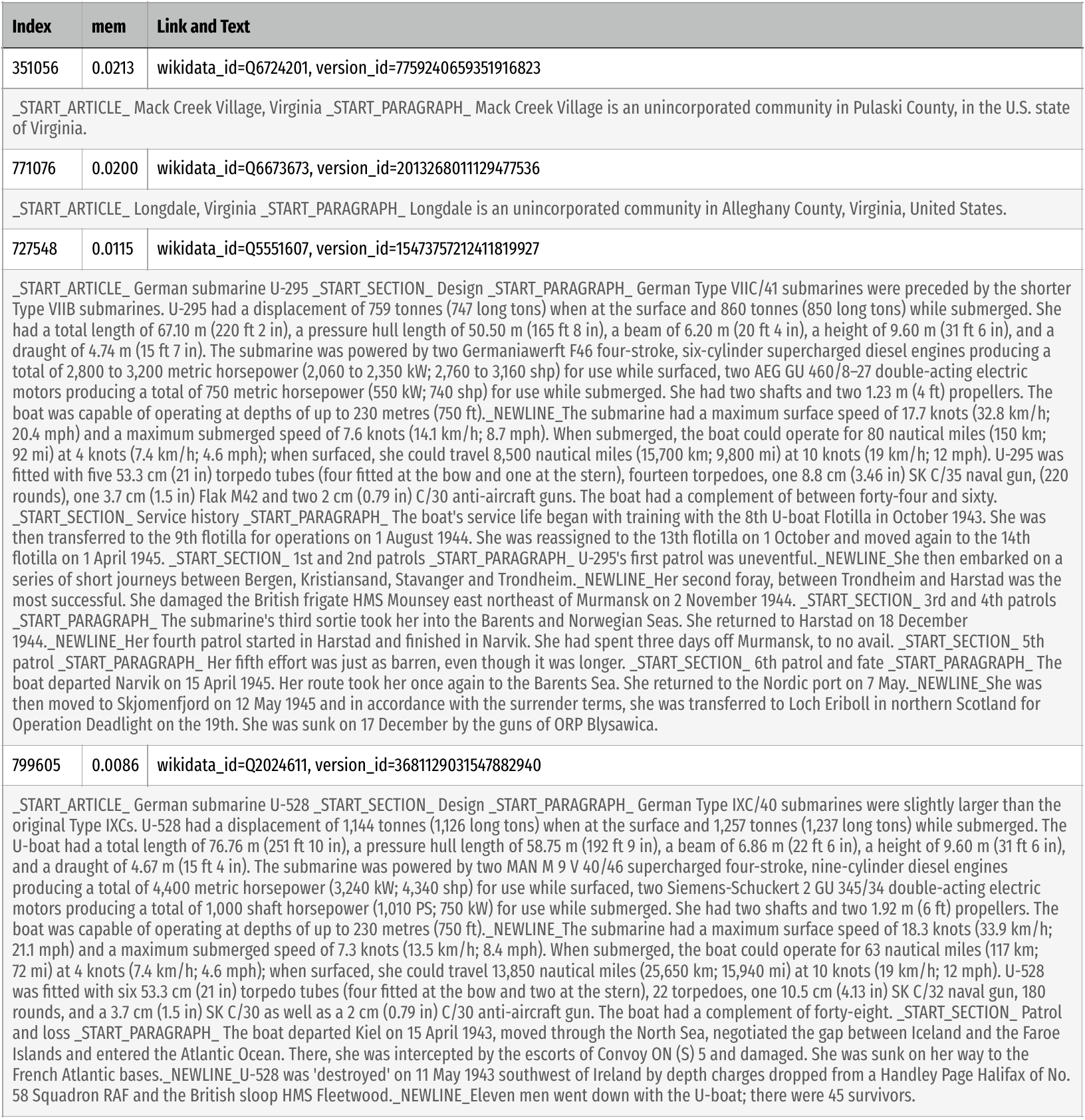}
    \caption{Text examples from \wikien with low memorization.}
    \label{fig:egs-wikitext-low-mem}
\end{figure}

Figure~\ref{fig:egs-realnews-high-mem}, Figure~\ref{fig:egs-realnews-mid-mem}, and Figure~\ref{fig:egs-realnews-low-mem} show full examples from \realnews sampled at high, middle and low memorization value ranges, respectively. Similarly, Figure~\ref{fig:egs-c4-high-mem}, Figure~\ref{fig:egs-c4-mid-mem}, and Figure~\ref{fig:egs-c4-low-mem} show examples from \cfour sampled at high, middle and low memorization value ranges, respectively. Figure~\ref{fig:egs-wikitext-high-mem}, Figure~\ref{fig:egs-wikitext-mid-mem}, and Figure~\ref{fig:egs-wikitext-low-mem} show examples from \wikien sampled at high, middle and low memorization value ranges, respectively.




\section{Example Pairs Sampled at Different Level of Influence}

\begin{figure}
    \centering
    \includegraphics[width=\linewidth]{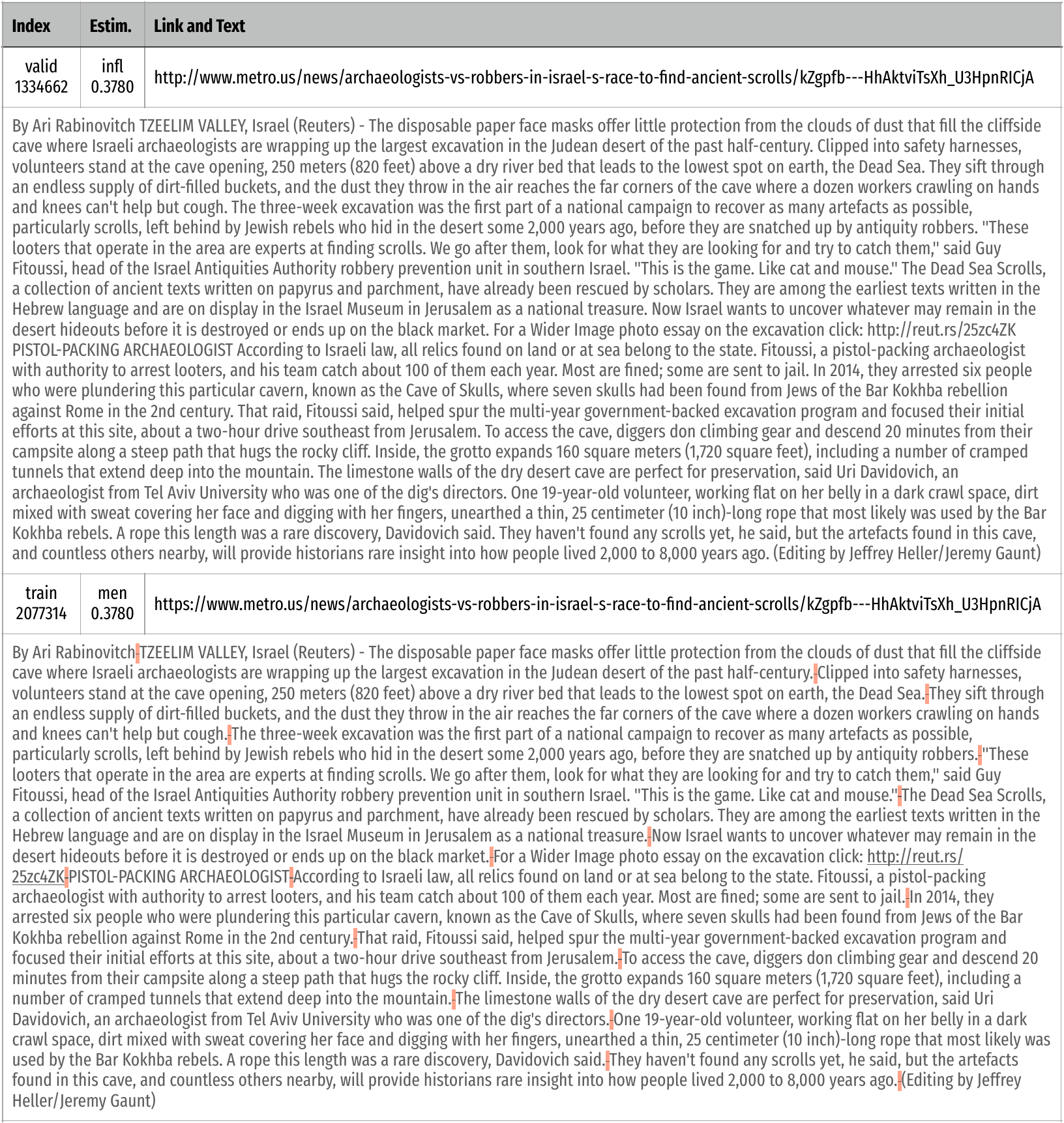}
    \caption{Validation / training example pair from \realnews with high influence. Red / green highlighted text indicate deleted / added text in the training example comparing to the corresponding validation example, generated using Python \texttt{difflib}.}
    \label{fig:egs-realnews-infl-1}
\end{figure}

\begin{figure}
    \centering
    \includegraphics[width=\linewidth]{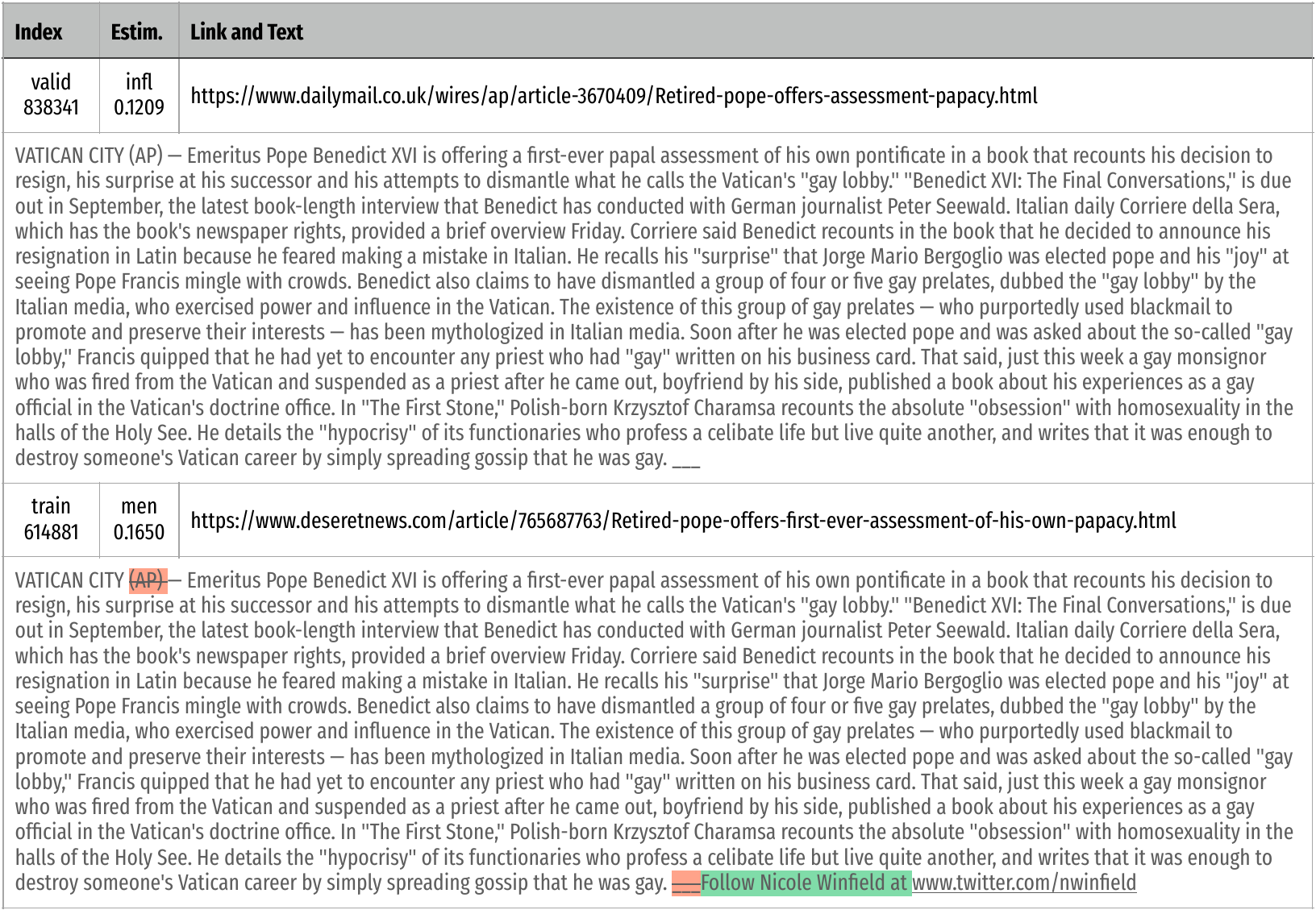}
    \caption{Validation / training example pair from \realnews with relatively high influence. Red / green highlighted text indicate deleted / added text in the training example comparing to the corresponding validation example, generated using Python \texttt{difflib}.}
    \label{fig:egs-realnews-infl-2}
\end{figure}

\begin{figure}
    \centering
    \includegraphics[width=\linewidth]{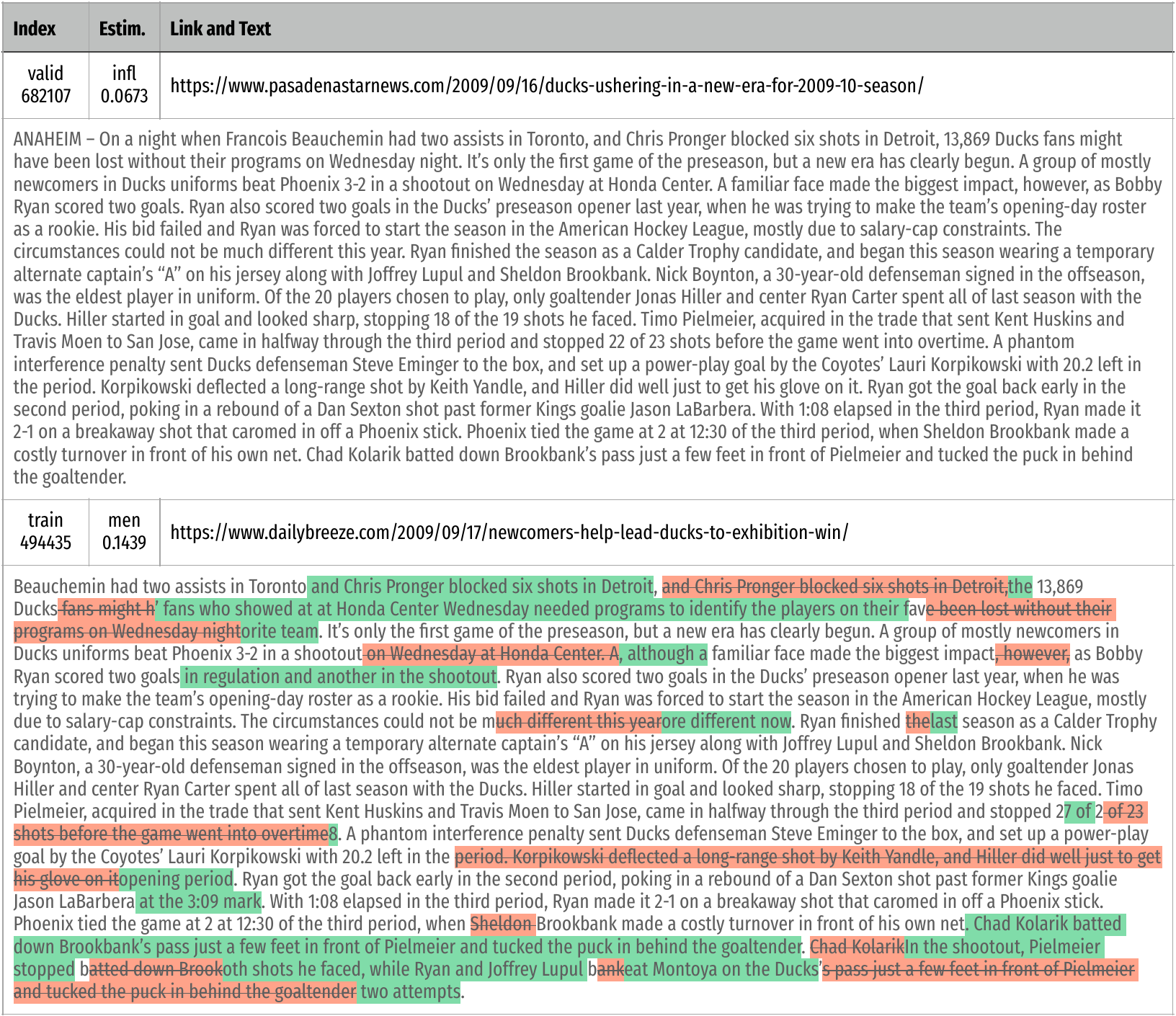}
    \caption{Validation / training example pair from \realnews with intermediate influence. Red / green highlighted text indicate deleted / added text in the training example comparing to the corresponding validation example, generated using Python \texttt{difflib}.}
    \label{fig:egs-realnews-infl-3}
\end{figure}

\begin{figure}
    \centering
    \includegraphics[width=\linewidth]{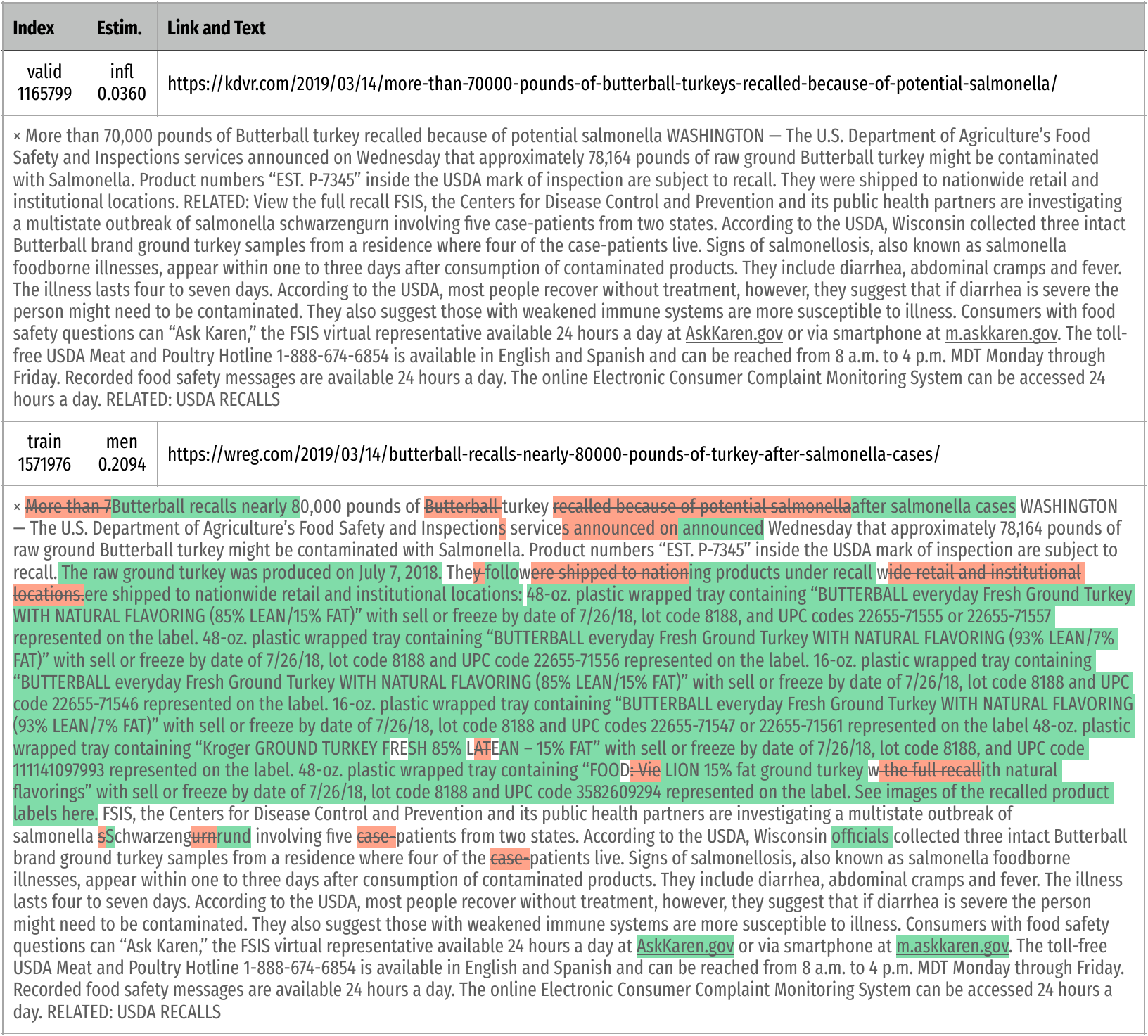}
    \caption{Validation / training example pair from \realnews with relatively low influence. Red / green highlighted text indicate deleted / added text in the training example comparing to the corresponding validation example, generated using Python \texttt{difflib}.}
    \label{fig:egs-realnews-infl-4}
\end{figure}

\begin{figure}
    \centering
    \includegraphics[width=\linewidth]{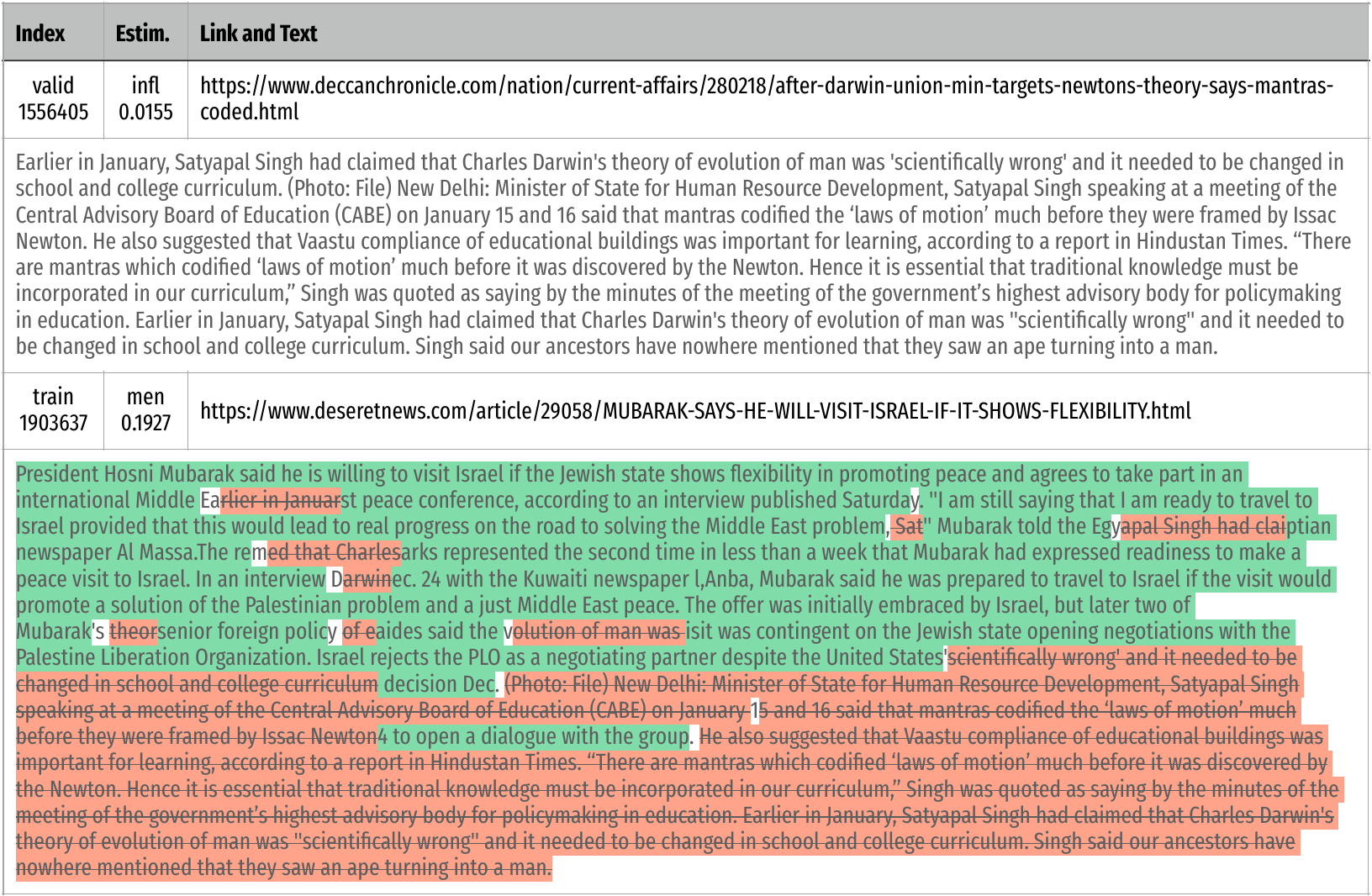}
    \caption{Validation / training example pair from \realnews with low influence. Red / green highlighted text indicate deleted / added text in the training example comparing to the corresponding validation example, generated using Python \texttt{difflib}.}
    \label{fig:egs-realnews-infl-5}
\end{figure}

Figure~\ref{fig:egs-realnews-infl-1}, Figure~\ref{fig:egs-realnews-infl-2}, Figure~\ref{fig:egs-realnews-infl-3}, Figure~\ref{fig:egs-realnews-infl-4}, and Figure~\ref{fig:egs-realnews-infl-5} show train-validation example pairs from \realnews sampled from high to low influence ranges. For each pair, we show the validation set example first, and then show the corresponding training example with a \texttt{difflib} generated visualization of textual difference with the training example.

\begin{figure}
    \centering
    \includegraphics[width=\linewidth]{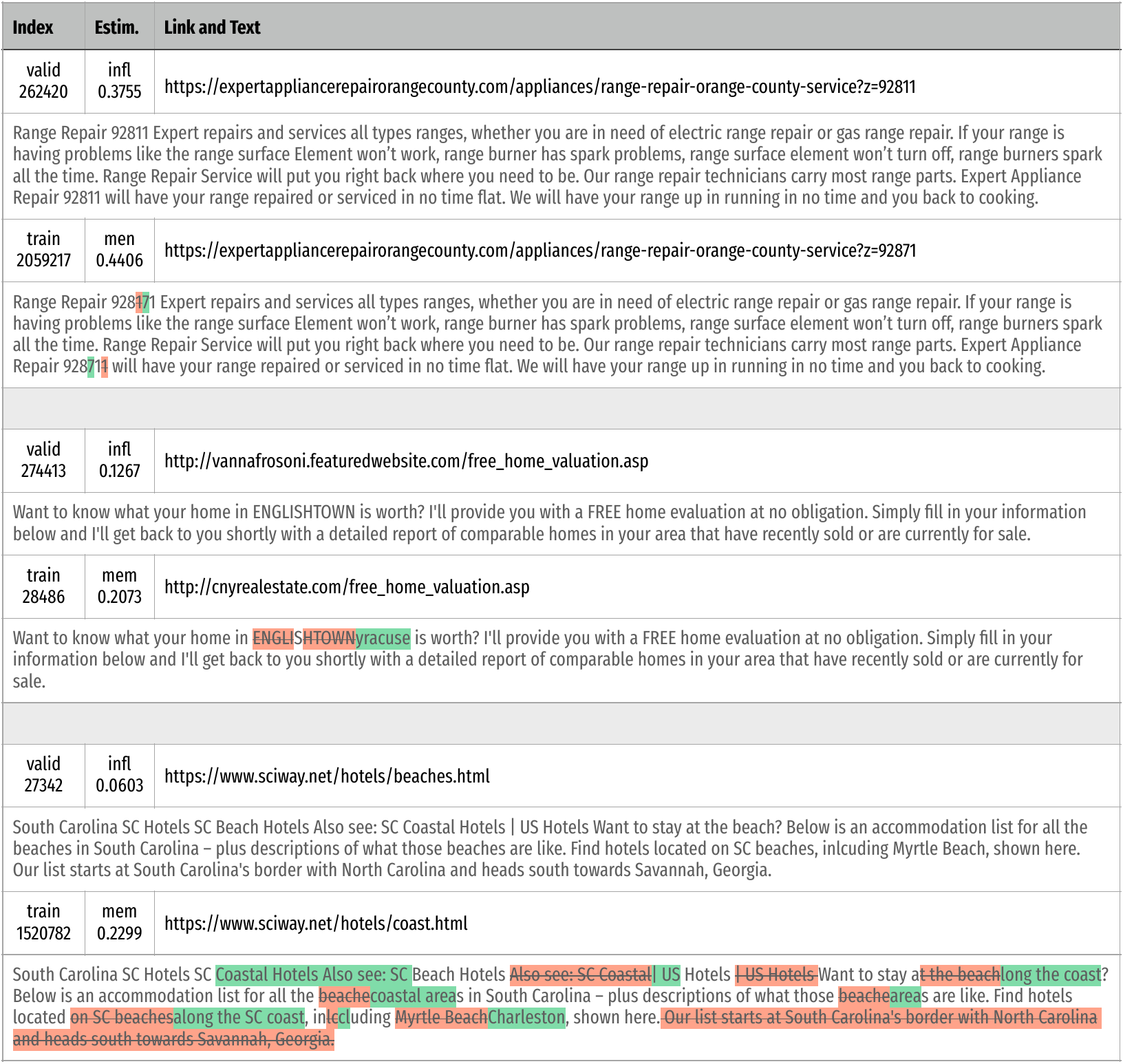}
    \caption{Validation / training example pair from \cfour with high to intermediate influence. Red / green highlighted text indicate deleted / added text in the training example comparing to the corresponding validation example, generated using Python \texttt{difflib}.}
    \label{fig:egs-c4-infl-1}
\end{figure}

\begin{figure}
    \centering
    \includegraphics[width=\linewidth]{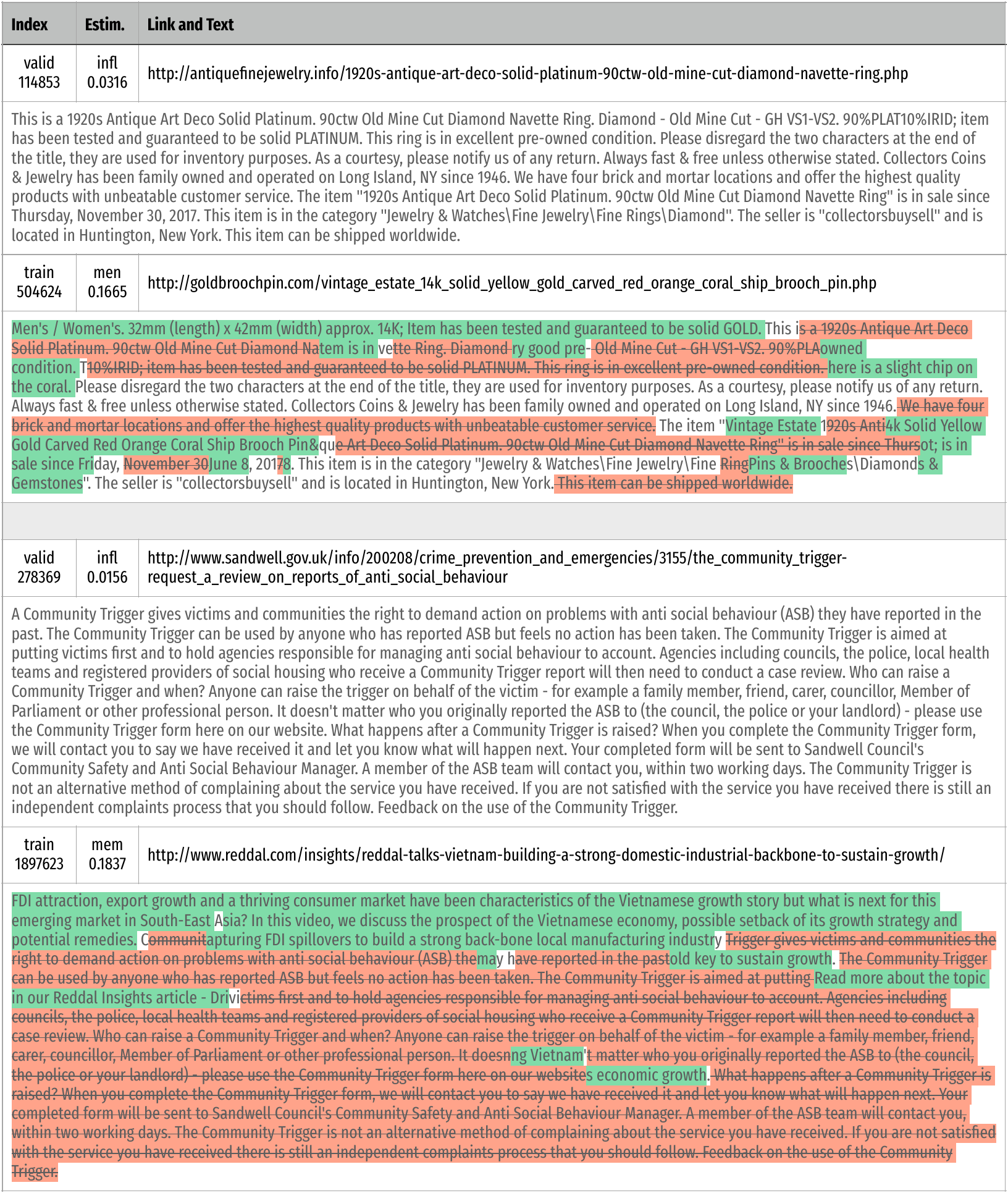}
    \caption{Validation / training example pair from \cfour with intermediate to low influence. Red / green highlighted text indicate deleted / added text in the training example comparing to the corresponding validation example, generated using Python \texttt{difflib}.}
    \label{fig:egs-c4-infl-2}
\end{figure}

\begin{figure}
    \centering
    \includegraphics[width=\linewidth]{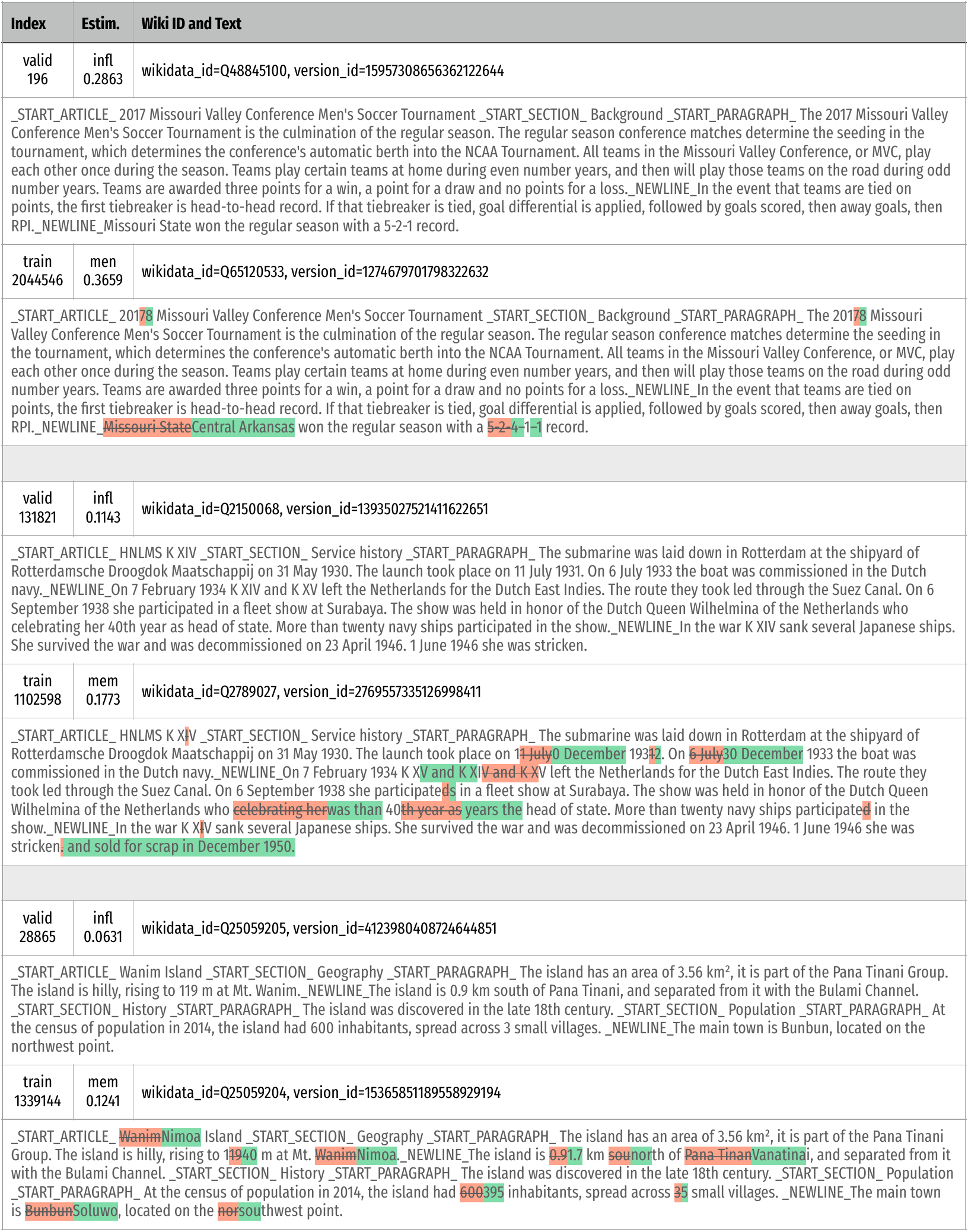}
    \caption{Validation / training example pair from \wikien with high to intermediate influence. Red / green highlighted text indicate deleted / added text in the training example comparing to the corresponding validation example, generated using Python \texttt{difflib}.}
    \label{fig:egs-wikitext-infl-1}
\end{figure}

\begin{figure}
    \centering
    \includegraphics[width=\linewidth]{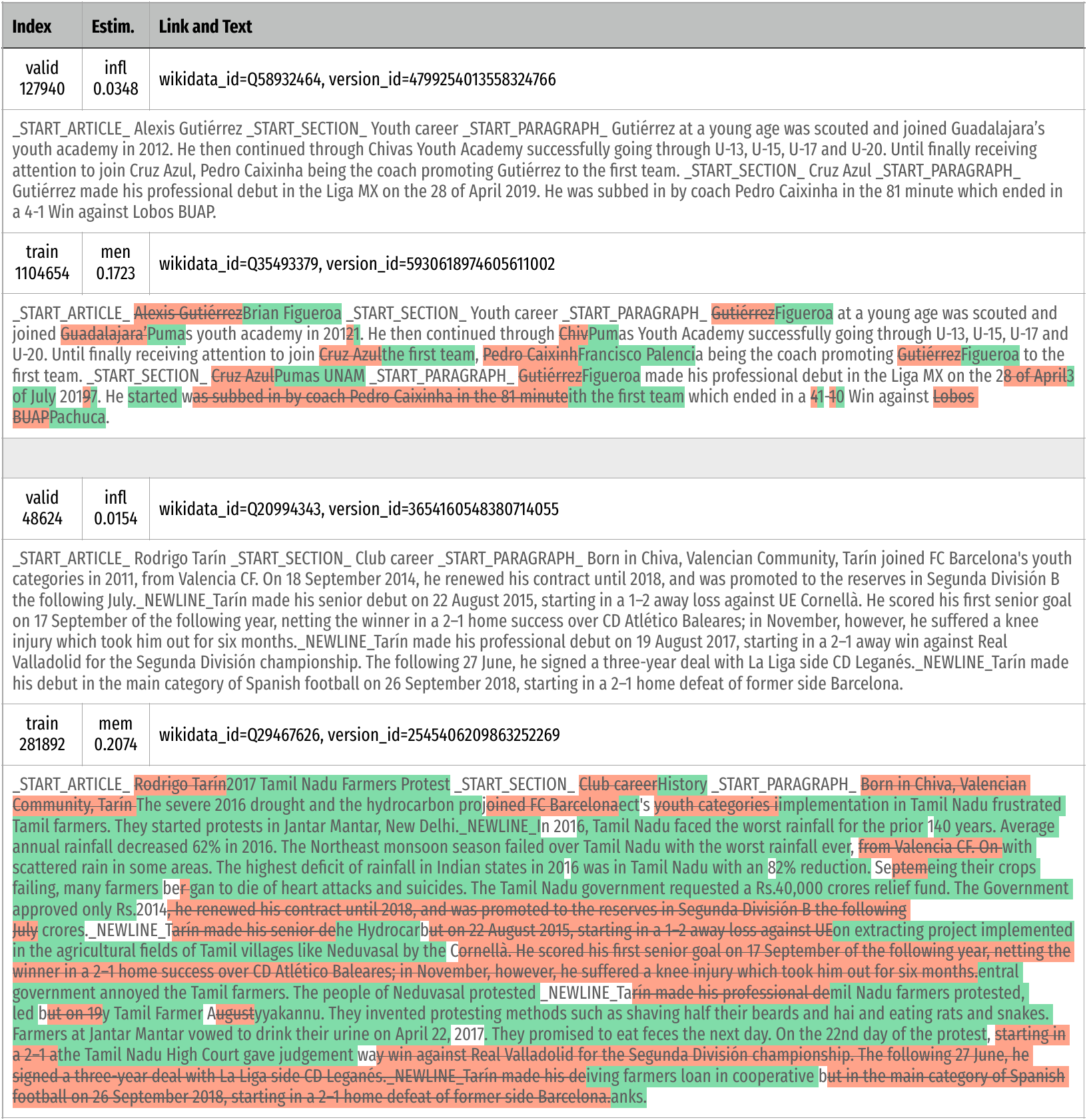}
    \caption{Validation / training example pair from \wikien with intermediate to low influence. Red / green highlighted text indicate deleted / added text in the training example comparing to the corresponding validation example, generated using Python \texttt{difflib}.}
    \label{fig:egs-wikitext-infl-2}
\end{figure}

Similarly, Figure~\ref{fig:egs-c4-infl-1} and Figure~\ref{fig:egs-c4-infl-2} show train-validation example pairs from \cfour, and Figure~\ref{fig:egs-wikitext-infl-1} and Figure~\ref{fig:egs-wikitext-infl-2} from \wikien.

\begin{figure}
    \centering
    \includegraphics[width=\linewidth]{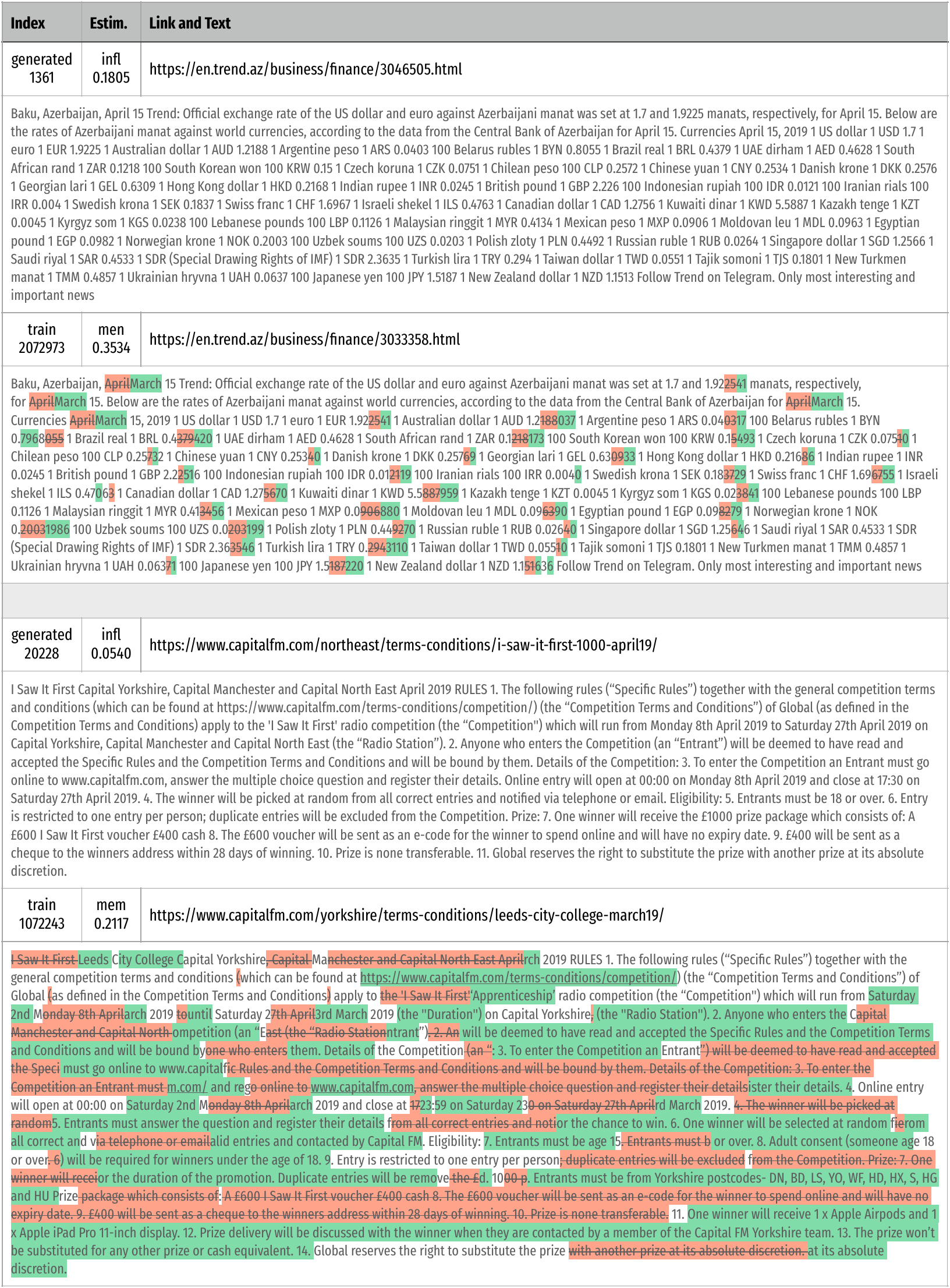}
    \caption{Generated / training example pair from \realnews with high to intermediate influence. The generated examples are directly taken from publicly released generations of the  Grover-Mega (p=0.96) model~\citep{zellers2019defending}. Red / green highlighted text indicate deleted / added text in the training example comparing to the corresponding validation example, generated using Python \texttt{difflib}.}
    \label{fig:egs-grover-infl-1}
\end{figure}

\begin{figure}
    \centering
    \includegraphics[width=\linewidth]{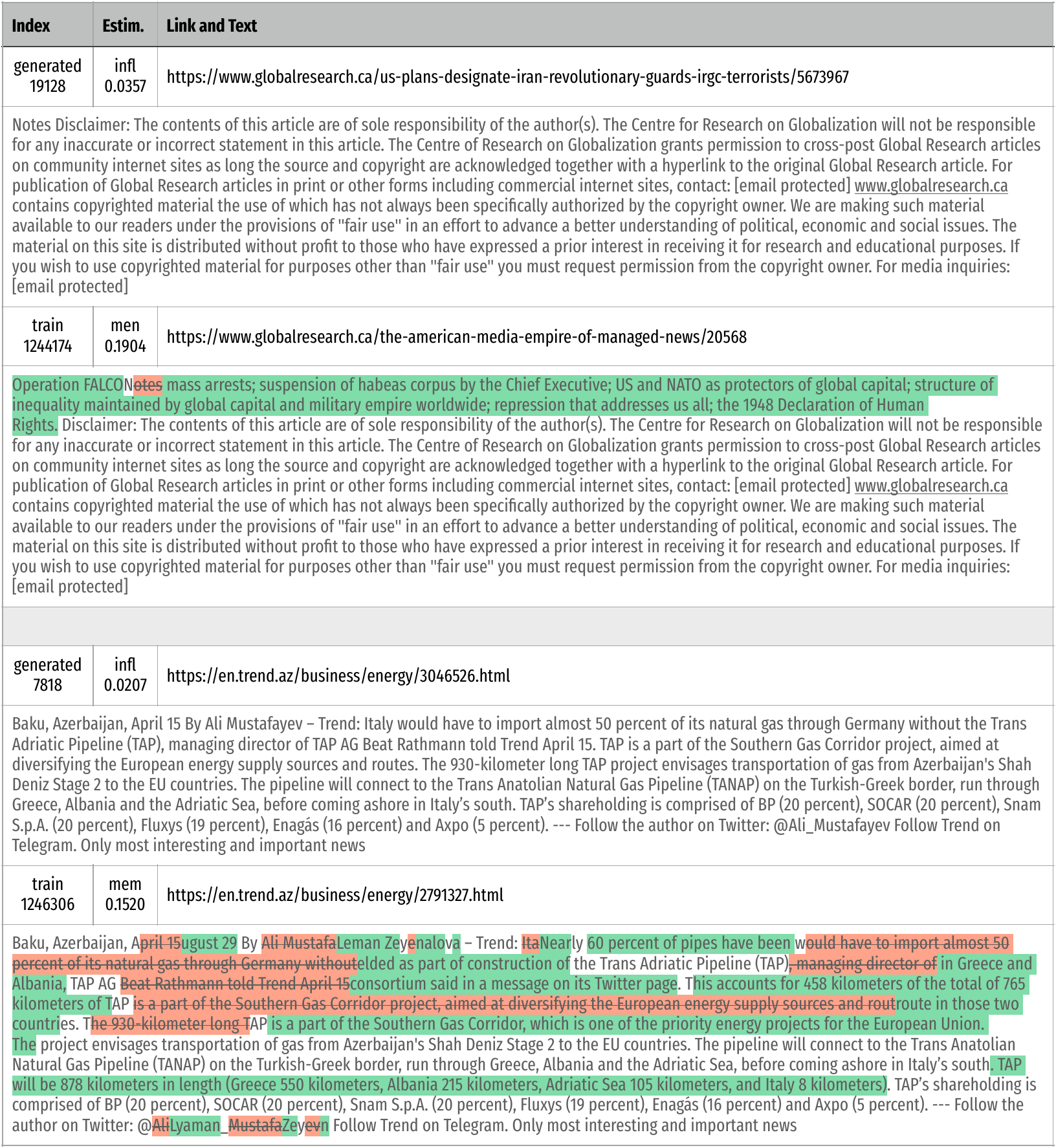}
    \caption{Generated / training example pair from \realnews with intermediate to low influence. The generated examples are directly taken from publicly released generations of the  Grover-Mega (p=0.96) model~\citep{zellers2019defending}.  Red / green highlighted text indicate deleted / added text in the training example comparing to the corresponding validation example, generated using Python \texttt{difflib}.}
    \label{fig:egs-grover-infl-2}
\end{figure}

\begin{figure}
    \centering
    \includegraphics[width=\linewidth]{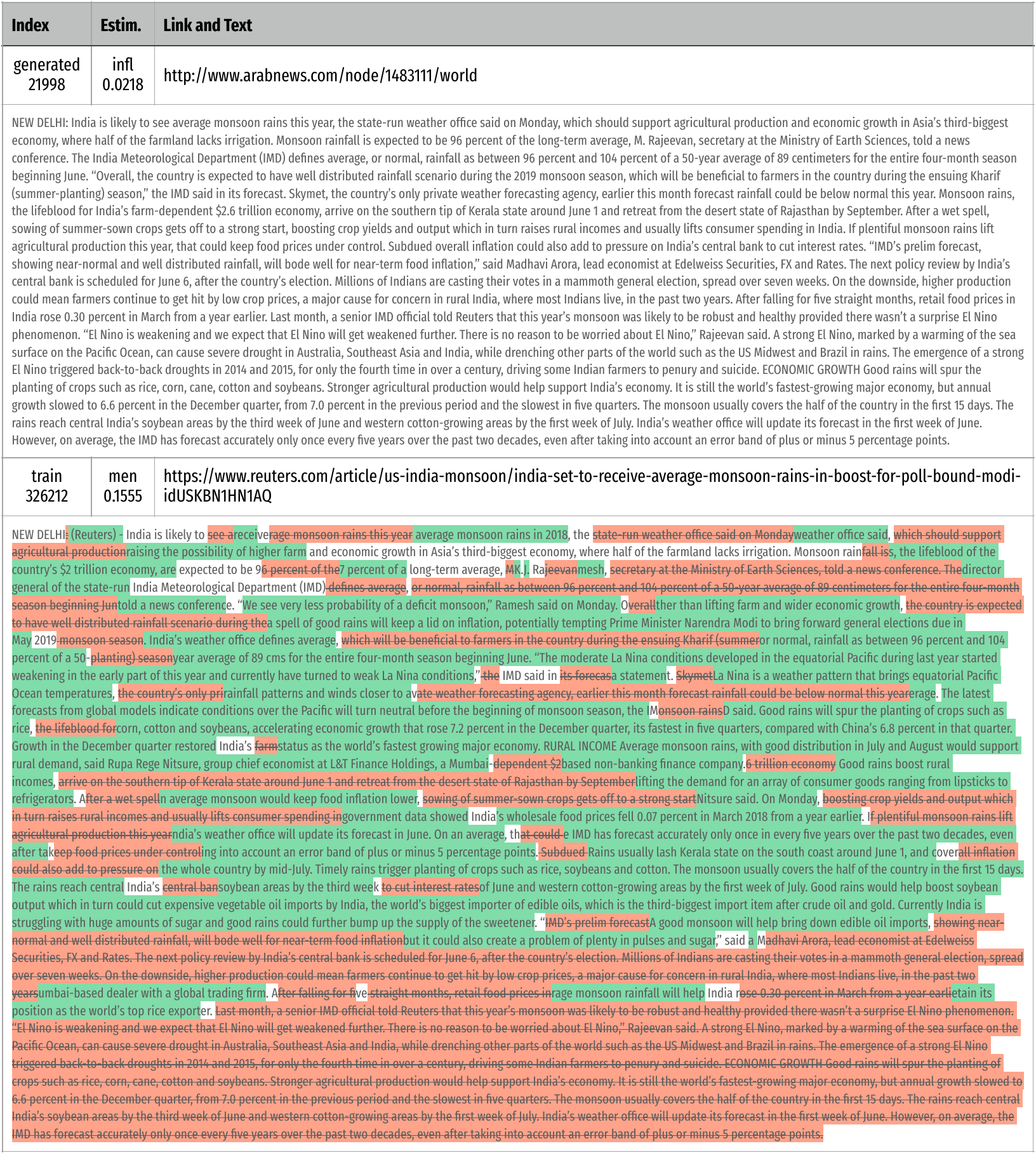}
    \caption{Generated / training example pair from \realnews with low influence. The generated examples are directly taken from publicly released generations of the  Grover-Mega (p=0.96) model~\citep{zellers2019defending}.  Red / green highlighted text indicate deleted / added text in the training example comparing to the corresponding validation example, generated using Python \texttt{difflib}.}
    \label{fig:egs-grover-infl-3}
\end{figure}

We also show train-generation influence pairs between \realnews training set and Grover~\citep{zellers2019defending} model generation in Figure~\ref{fig:egs-grover-infl-1}, Figure~\ref{fig:egs-grover-infl-2}, and Figure~\ref{fig:egs-grover-infl-3}.


\end{document}